%% file: main.tex
\documentclass[10pt,twocolumn,letterpaper]{article}

\usepackage[pagenumbers]{cvpr} 

\input{preamble}

\usepackage{tabularx}
\usepackage{multirow}
\usepackage{booktabs}
\usepackage{algorithm, amsmath, algpseudocode}
\usepackage{float}
\usepackage{caption}
\usepackage{tcolorbox}

\usepackage{etoc}

\newcolumntype{Y}{>{\centering\arraybackslash}p{0.75cm}} 
\newcolumntype{L}{>{\raggedright\arraybackslash}p{0.1cm}} 

\definecolor{cvprblue}{rgb}{0.21,0.49,0.74}
\usepackage[pagebackref,breaklinks,colorlinks,citecolor=cvprblue]{hyperref}
\def\paperID{5569} 
\def\confName{CVPR}
\def\confYear{2024}

\title{
InCoRo: In-Context Learning for Robotics Control
with Feedback Loops

}

\author{Jiaqiang Ye Zhu\\
THEKER ROBOTICS\\
{\tt\small jq@theker.eu}
\and
Carla Gómez Cano\\
THEKER ROBOTICS\\
{\tt\small carla@theker.eu}
\and
David Vazquez\\
ServiceNow Research\\
{\tt\small david.v@servicenow.com}
\\
\and
Michał Drożdżal\\
Meta AI\\
{\tt\small mdrozdzal@meta.com}
}

\begin{document}

\twocolumn[{%
\renewcommand\twocolumn[1][]{#1}%
\maketitle

\begin{center}
    \centering
    \captionsetup{type=figure}
    \includegraphics[width=0.96\linewidth]{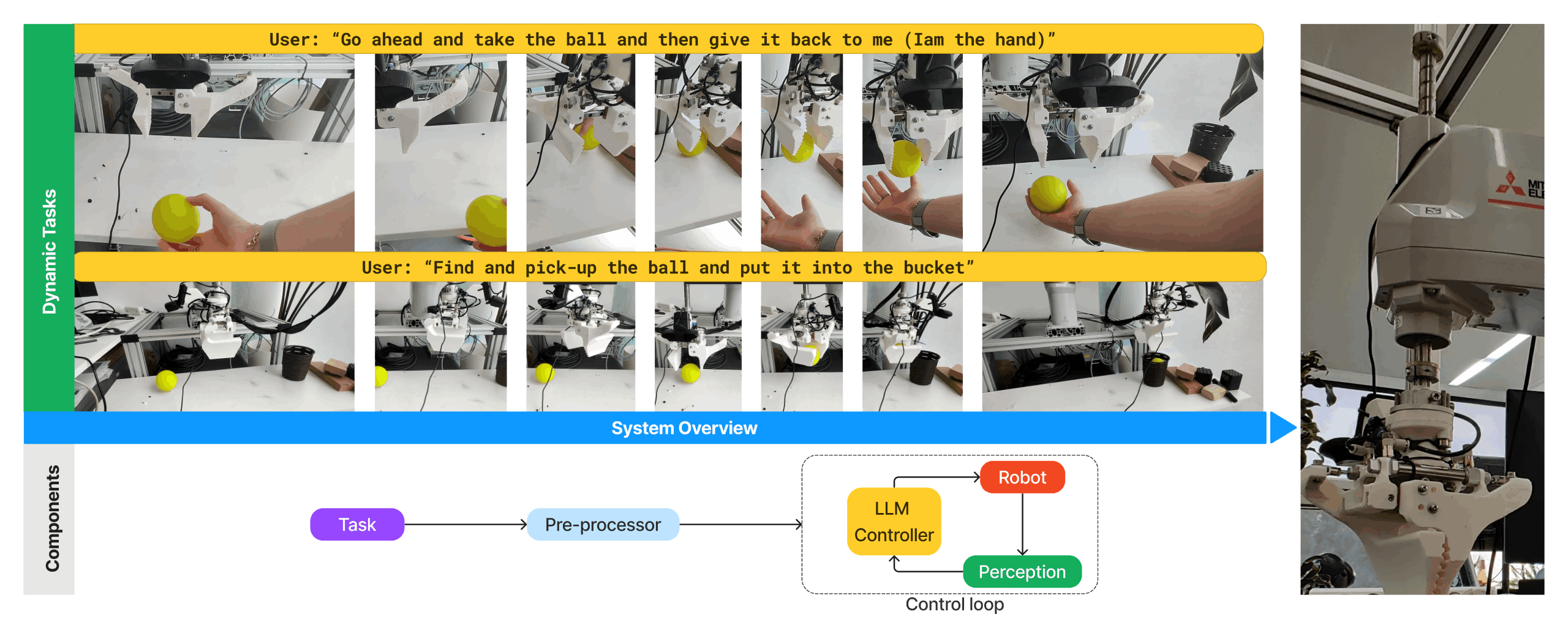}
    \captionof{figure}{\textbf{Our system in action.} The first two lines depict sequences of frames collected when the robot is executing two actions in dynamic conditions. The top row, includes moving user hand, while in the second row the bucket is outside of the camera's initial field of view. Our system achieves high success rate for these challenging tasks. The bottom row, depicts the main components of our system: user-provided input text, pre-processor responsible for decomposition of user prompt into atomic actions and objects, and a control loop equipped with a large language model, perception unit and a robot. The robot is displayed in the right-most image.}
    \label{fig:front}
\end{center}%
}]


\input{sec/0_abstract}

\input{sec/1_intro}

\input{sec/2_method}
\input{sec/3_experiments}

\input{sec/IV_Related_work}
\input{sec/5_Conclusions}

{
    \small
    \bibliographystyle{ieeenat_fullname}
    \bibliography{main}
}
\onecolumn
\input{sec/X_suppl}
\end{document}

%% file: preamble.tex
\usepackage[dvipsnames]{xcolor}


%% file: sec/0_abstract.tex
\begin{abstract}
One of the challenges in robotics is to enable robotic units with the reasoning capability that would be robust enough to execute complex tasks in dynamic environments. Recent advances in Large Language Models (LLMs) have positioned them as go-to tools for simple reasoning tasks, motivating the pioneering work of \citet{liang2023code} that uses an LLM to translate natural language commands into low-level \emph{static} execution plans for robotic units. Using LLMs inside robotics systems brings their generalization to a new level, enabling zero-shot generalization to new tasks. This paper extends this prior work to \emph{dynamic} environments. We propose InCoRo, a system that uses a classical robotic feedback loop composed of an LLM controller, a scene understanding unit, and a robot. Our system continuously analyzes the state of the environment and provides adapted execution commands, enabling the robot to adjust to changing environmental conditions and correcting for controller errors. Our system does not require \emph{any} iterative optimization to learn to accomplish a task as it leverages in-context learning with an off-the-shelf LLM model. Through an extensive validation process involving two standardized industrial robotic units -- SCARA and DELTA types -- we contribute knowledge about these robots, not popular in the community, thereby enriching it. We highlight the generalization capabilities of our system and show that (1) in-context learning in combination with the current state-of-the-art LLMs is an effective way to implement a robotic controller; (2) in static environments, InCoRo surpasses the prior art in terms of the success rate by 72.6\% and 62.3\% for the SCARA and DELTA units, respectively; (3) in dynamic environments, we establish new state-of-the-art obtaining 83.2\% and 65.6\% success rates for the SCARA and DELTA units, respectively. This research paves the way towards building reliable, efficient, intelligent autonomous systems that adapt to dynamic environments.
\end{abstract}

%% file: sec/1_intro.tex
\section{Introduction}
\label{sec:intro}

Robotics is a rapidly growing field with the potential to revolutionize our lives~\citep{SOORI202354}. However, there is an urgent need to bridge the gap between the potential of robots and real-world challenges~\citep{Chaka_2023}. Key ingredients missing from today's robotics include robotic systems that can understand their surroundings in real-time, make instantaneous decisions, and adapt to unforeseen changes. This would mark a monumental stride in this field, enabling robots to navigate and operate effectively in dynamic and unstructured environments. Such a system would not only be able to tackle the intrinsic challenges posed by dynamic environments but also enhance the efficiency, reliability, and safety of robotic operations.

State-of-the-art techniques can convert human-issued natural language directives into machine-executable code~\citep{liang2023code}, enabling easy prototyping of robotic instructions. However, their applicability is still limited to static environments as they lack a feedback loop. 
As a result, they can't adapt well if something unexpected happens, like if the robot's controller makes an error or if something in the environment changes suddenly. Moreover, these methods can't quickly adjust to new situations, making them less functional in complex or unpredictable settings.

This paper proposes a new system, InCoRo (\textbf{In}-\textbf{Co}ntext Learning for \textbf{Ro}botics with Feedback Loops), that extends~\citet{liang2023code} to work with dynamic environments by integrating a classical control feedback loop mechanism~\citep{BENNETT200143, 68075, annurev, 4047983, 760351, 200137}. As a result, InCoRo addresses the perception, control, and complexity limitations of~\citet{liang2023code} 
Our system uses a combination of visual scene understanding and tracking, robot sensor information, and a Large Language Model (LLM) controller to enable robots to understand their environment, plan actions, and execute those actions safely and efficiently with zero-shot generalization to new tasks. 
To the best of our knowledge, our work is the first to integrate an LLM controller with a classical robotic feedback loop. Our system is designed to require only a single input, a text describing a task the robot should accomplish, \emph{e.g.}, \textit{build a tower from the elements present on the table}. Figure~\ref{fig:front} shows examples of the initial and final robot states within an instruction. 

It is important to note that contrary to many prior works that included a control feedback loop, our system does not require costly iterative optimization-based learning approaches such as back-propagation to become proficient in task execution. We leverage recent advances in LLM literature~\citep{dong2023survey, ye2023incontext, zhou2023efficient, li2023otter, Khattak_2023_CVPR, nguyen2023incontext, ICL1, brown2020language} and propose to use in-context learning that provides learning examples as a part of the LLM input prompt. Thus, we can use a pre-trained, off-the-shelf LLM model as our controller. Through our in-depth experiments with real robots, we show that in-context learning and a feedback loop effectively control robots in dynamic environments. The overview of our system is presented in Figure~\ref{fig:front}. Our experiments with two robotic arms -- DELTA and SCARA -- show that we outperform~\citet{liang2023code}. In static environments, InCoRo surpasses the prior art in terms of the success rate by 72.6\% and 62.3\% for the SCARA and DELTA units, respectively; in dynamic environments, we establish new state-of-the-art obtaining 83.2\% and 65.6\% success rates for the SCARA and DELTA units, respectively.

The contributions of this work can be summarized as follows:
\begin{itemize}
    \item We extend~\citet{liang2023code} by adding a classical feedback loop and introduce InCoRo (\textbf{In}-\textbf{Co}ntext Learning for \textbf{Ro}botics with Feedback Loops) control system. Our system can mitigate controller errors and adapt to dynamic environmental conditions.
    \item The use of in-context learning makes our approach very versatile as it does not require any iterative optimization to learn to execute on new tasks enabling zero-shot generalization. As a result one can use an off-the-shelf LLM as robotic controller.
    \item Using an extensive validation, we outperform~\citet{liang2023code} by up to 72.6\% and 62.3\% in static environments with SCARA and DELTA robots, respectively; and reach a success rate of up to 83.2\% and 65.6\% on complex tasks in dynamic environments with SCARA and DELTA robots, respectively.
    \item We deploy our system into two real-world robotic arms acting in static and dynamic environments to demonstrate that the system can be generalised. It is important to note that changing the robotic platform only requires minimal modifications in our system.
\end{itemize}

We believe that the contributions of this paper make a significant advancement in the field of robotics. By addressing the challenges of perception, control, and complexity, our system can help to make robots more reliable, efficient, and intelligent. This will pave the way for the development of autonomous systems that can operate in various environments and perform a wide range of tasks, not only with articulated robots but can be used in any robotic or autonomous system. 
%

%% file: sec/2_method.tex
\section{Method}
\label{method}

Our system takes as input a user-provided task description and executes it on a robotic unit. On a high level, our system comprises two main components: a \emph{pre-processor} and a \emph{control loop}. 

\begin{figure}
  \centering
  \includegraphics[width=1\linewidth]{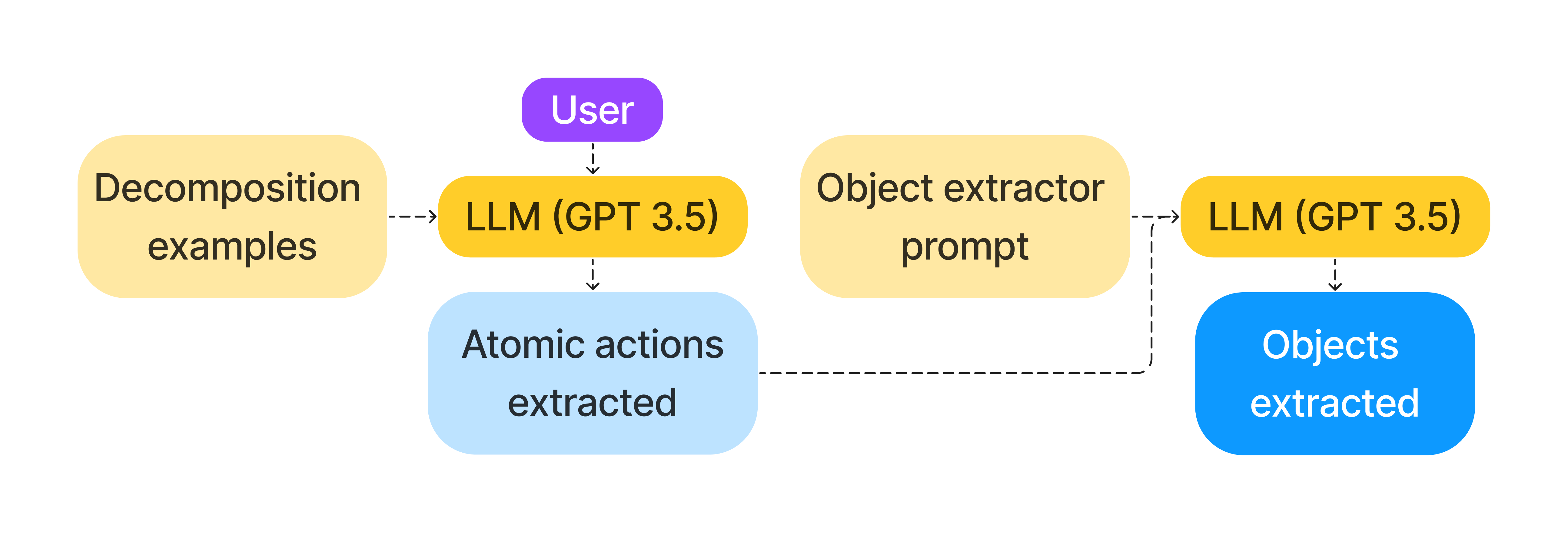}
  \caption{\textbf{Pre-processing diagram.} Our pre-processing unit leverages in-context learning to decompose user-provided text into a sequence of atomic actions and a list of objects.}
  \label{fig:prepro}
\end{figure}

\paragraph{Pre-processor.}
Similarly to prior works~\cite{saycan2022arxiv, liang2023code}, we assume that the user-provided complex instruction could be broken down into a sequence of simple atomic actions that are easier to understand and execute by the robot. In our system we have concluded that it is not necessary to use weight gauge as the results are similar and it lengthens the execution time.
We note that the order in which the atomic actions appear is crucial to accomplish the task -- \eg, \emph{draw a line with a pen} could be decomposed into the following two sequences of atomic actions (1) \emph{take a pen}, \emph{remove the cap}, and \emph{draw a line} and (2) \emph{take a pen}, \emph{draw a line}, and \emph{remove the cap}; however, only the first one allows to draw a line. Thus, our pre-processor leverages in-context learning~\citep{ICL1, wei2022emergent, dong2023survey, zhang2023language} to provide a sequence of atomic actions from the user-provided task description. In addition, the pre-processor extracts objects from the user-provided task description, which we use to guide our perception model. In both cases, we give examples on how to accomplish the tasks in the LLM's context. Note that the pre-processor requires two forward passes through the LLM and is executed only once per user-provided prompt (see Figure~\ref{fig:prepro}). The exemplary prompts to the LLM for both atomic action and object extractions are provided in in section~\ref{ap-sec:ex-prompts} of the appendix.

 \begin{figure*}
   \centering
   \includegraphics[width=1\linewidth]{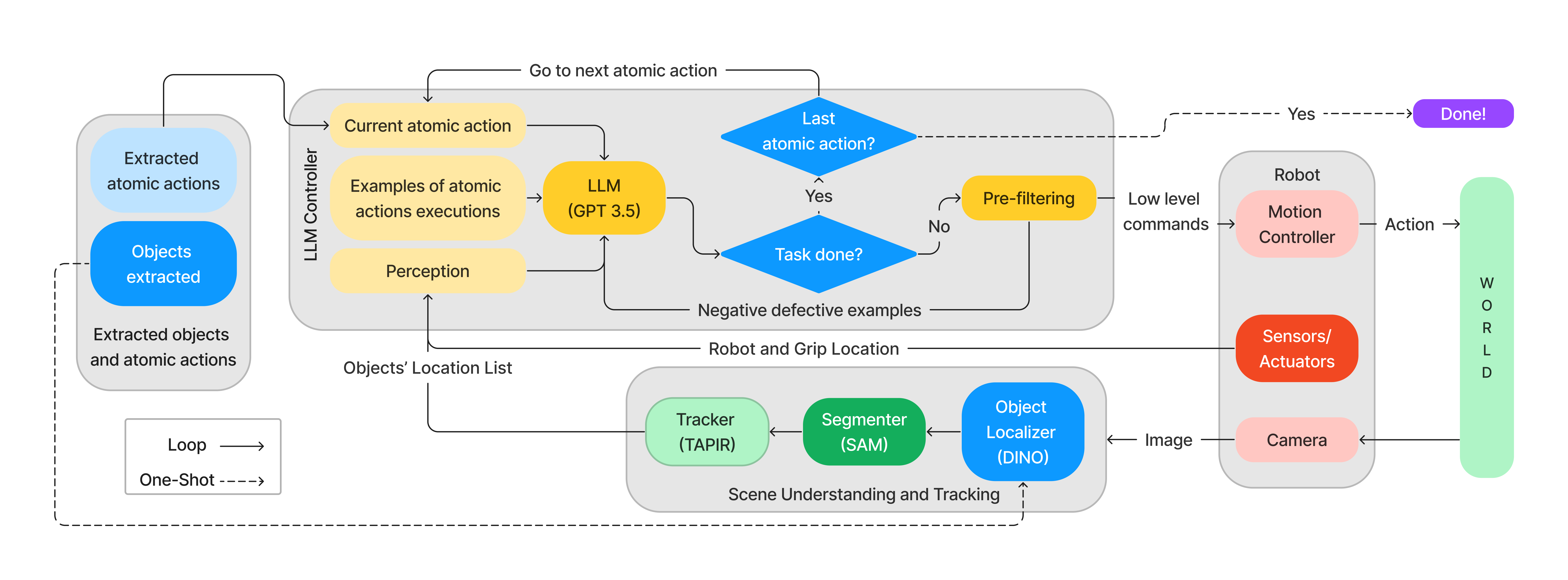}
   \caption{\textbf{Our control loop.} The control loop inputs the list of atomic actions and a set of objects extracted from the user-defined textual descriptions by the pre-processor. The loop consists of three elements: (1) a Large Language Model (LLM) controller that takes as an input the atomic task together with the robot's states and scene description and outputs low-level robot control commands, (2) a robot that acts in the world, and (3) a scene understanding module that continuously process the images to provide the locations of the objects in the scene. The controller can process multiple feedback operations per second when solving the user-defined task.}
   \label{fig:genmethod}
 \end{figure*}



\subsection{Control loop}
The overview of our control loop is provided in Figure~\ref{fig:genmethod}. Our control loop is composed of a \emph{LLM-based controller}, a \emph{robot}, and a \emph{scene understanding and tracking unit}. The LLM controller produces low-level instructions in robot-understandable form from atomic actions while leveraging the robot's state information and current visual perception. The robot is responsible for executing the low-level instructions in the environment. Parallel to this, the robot's perception unit actively processes visual stimuli of the environment. The control loop is a core system component; it continuously updates and provides information about the robot's state and perception to the LLM controller. The loop can provide multiple updates during the action, ensuring that both the controller and, thus, the robot are aware of the current state of the environment. Should any inconsistencies emerge, such as an unforeseen displacement of an object from its anticipated position, the feedback loop instantly recognizes this. The system then recalculates and dispatches adjusted instructions to the robot, ensuring that the robot can adapt on the fly and maintain task accuracy.



\subsubsection{LLM controller}
Our LLM controller is a crucial component of our system as it guides the robot to execute the actions properly. To this end, our LLM controller comprises a pre-trained LLM and pre-execution filter. The LLM controller takes as an input an \emph{atomic action}, \emph{examples of atomic action executions with the corresponding low-level instructions} leveraging in context learning, and the \emph{robot's perception}. The LLM outputs low-level instructions and a halting signal via a frame structure whereby the robot reads the instruction and comprehends the position indicating task accomplishment. This frame is displayed as in the following example:
\begin{equation}
  \left[ F,X,Y,Z,T_r,Tc_0,Tc_1 \right], \left[ H \right],
  \label{eq:important}
\end{equation}
where $F$ is a frame type, $X, Y, Z$ represents the coordinates in 3D space, $T_r$ is the gripper (or end-effector) rotation, $Tc_0$ and $Tc_1$ are gripper (or end-effector) controls (usually open-close) and $H$ is a binary halting signal. 
Before sending the instructions to the robot, we pre-filter them by removing inappropriate instructions or incomprehensible format. If the LLM produces a halt signal, we proceed with the next atomic action and continue the control loop until no atomic action is left.
Examples of low level actions can be found on Appendix \ref{ap-sec:low-level}.

\textbf{In-context learning examples.} The examples component includes 11 tasks that allow the language model to learn about the space of available robotic actions as it contains patterns and strategies that proved effective in similar past scenarios. It includes five basic robot movement tasks -- \emph{e.g.}, \textit{move into a rectangle shape} --, three picking up and moving object tasks -- \emph{e.g.}, \textit{pick up the red ball} --, and three examples of objects and environment interaction -- \emph{e.g.}, \textit{open a nut with a stone} --. Samples of our in-context learning examples are depicted in Appendix \ref{ap-sec:examples-learning}. 

\textbf{Robot's perception information.} The feedback comprises information from the \emph{robot's actuators}, such as gripper status, motor positions, etc., and \emph{object locations} obtained via computer vision-based scene understanding and tracking. 
This perception feedback refines the LLM controller by providing real-time data about the robot's position and surrounding objects. This not only narrows down the possible actions that the LLM controller could take but also mitigates the risk of generating hallucinations~\citep{zhang2023language} -- \eg, actions that are either unrealistic or inconsistent with the physical and operational constraints of the robotic arm --. By doing so, our system enhances its reliability, reduces the likelihood of erroneous or impractical commands, and reduces dependency on its in-context learning examples corroborating the work of \citet{hallucinations}.

\textbf{Pre-execution filter.}
The filter's primary function is to scrutinize the low-level commands generated by the LLM controller for potential issues; these include structural inconsistencies, overly ambitious movements that exceed hardware constraints, and actions of an inappropriate type for the given task. Commands that fail any of these checks are flagged and discarded. These rejected commands are not merely eliminated; they are recycled back into the LLM controller as negative defective examples, facilitating an in-context learning process that aids the LLM controller in refining its future decision-making. This filter thus acts as a vital safeguard, enhancing the system's reliability and adaptability by ensuring that every command passed to the robotic arm is technically feasible and contextually appropriate. The basic structure of the pre-execution filter command can be found in the Appendix \ref{ap-sec:pre-filter}.

\begin{figure}
  \centering
  \includegraphics[width=1\linewidth]{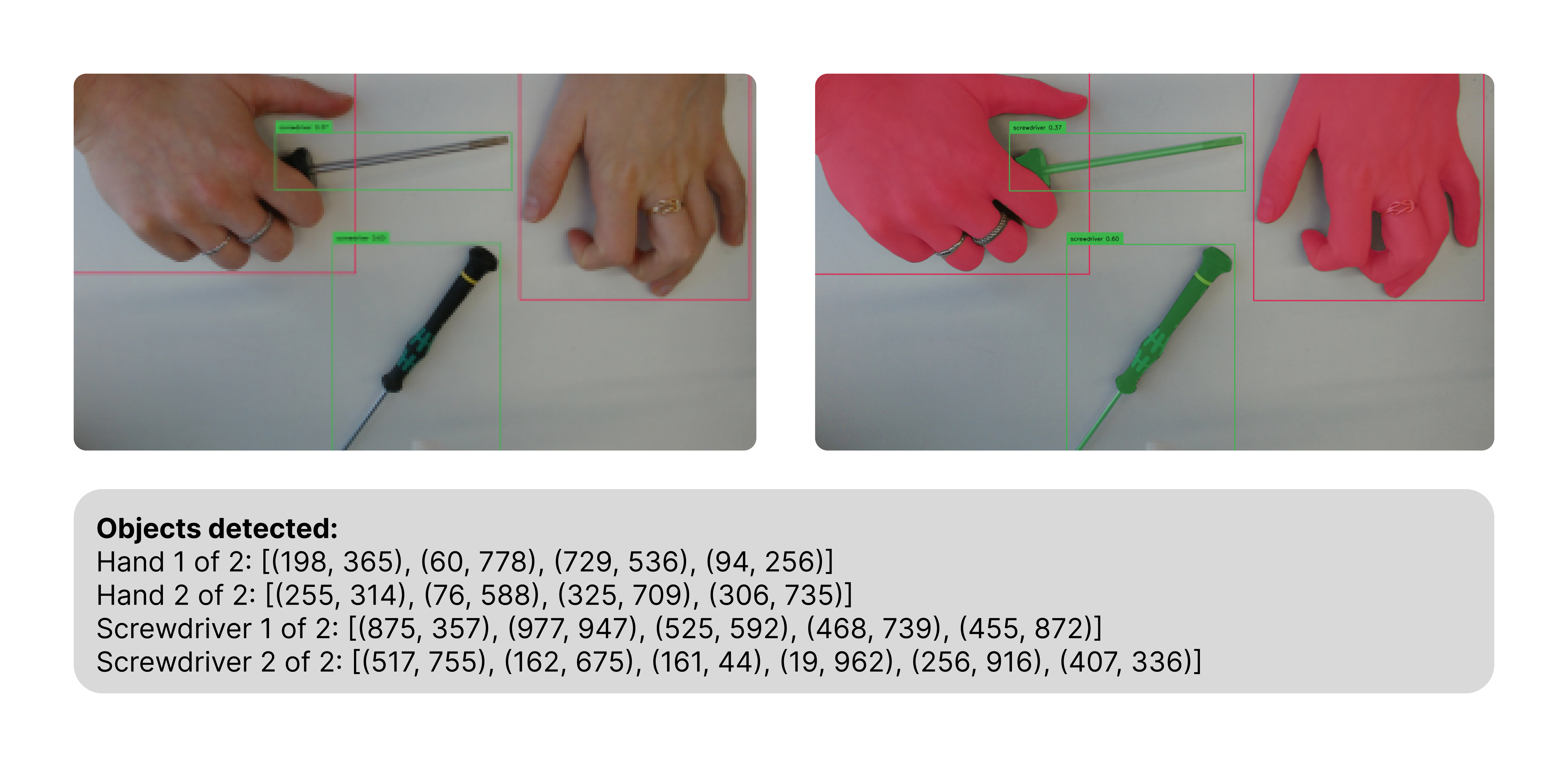}
  \caption{\textbf{Example of scene understanding input-output.} The two pictures on the left show the detection of the objects in the image, and on the right, the segmentation of these objects. Next, the coordinates of the different objects are passed in the form of a bounding box and a simplified mask. In this example, the four edges of a polygon are shown for reasons of space and understanding.}
  \label{fig:see-send}
\end{figure}

\textbf{Computer Vision Perception}
\label{computer-vision-perception}
The robot continuously scans the world with a camera system. It processes each image with the scene understanding and tracking module that localizes, classifies, segments, tracks, and projects to the robot coordinates each object in the scene. Figure~\ref{fig:see-send} shows an example of how an image is processed by this module and described as a text to send to the LLM controller.

\textbf{Object Detection.} Our object detector leverages Grounding DINO~\citep{liu2023grounding} that processes each image frame and produces scene decomposition into the object's bounding boxes. The objects the detector focuses on are determined from the user-provided task description by the pre-processing module. For example, the pre-processor would dissect this instruction into atomic actions if the user issues a command such as \emph{put the blocks in the bowl}. In this specific scenario, the relevant objects to be focused on are \emph{blocks} and \emph{bowls}. The set of objects is provided as an input to the detector, which outputs bounding boxes of the objects. 


\textbf{Object segmentation.} Precise object manipulation -- \eg, picking up a tool without knocking over nearby items -- requires fine-grained descriptions of the object's boundaries. 
Thus, we equip our system with a segmenter derived from Fast Segment Anything Model (FastSAM)~\citep{zhao2023fast} that takes as an input the bounding boxes and outputs segmentation polygons. 

\textbf{Object tracking.} To enable robust recognition of scene elements in dynamic environments characterized by object's occlusions and displacement, we equip our system with object's tracking capabilities. 
We use Tracking Any Point with per-frame Initialization and temporal Refinement (TAPIR) method~\citep{doersch2023tapir} as it maintains the continuity and stability of tracking, even when the objects undergo transformations or occlusions. Our tracking takes as an input points inside bounding boxes and segmentation polygons for four consecutive frames -- the current frame and the three past frames -- and outputs tracklets. 



\textbf{Projection to robot coordinates.} 
\label{Projection to robot coordinate}
The origin of the robot coordinate system (0,0) is situated at the robot's end effector. For the camera coordinate system, the origin (0, 0) is located at the bottom-left of the image. Both coordinate systems are right-handed, ensuring compatibility and straightforward transformation between the two. Given that the displacement between the camera and the robot is known, projecting coordinates from the camera system to the robot system involves a simple vector addition. This allows for accurate and consistent coordinates mapping from the camera's perspective to the robot's operational context.

\subsubsection{Robotic Unit}
\label{robotic-unit}
The robot is the physical embodiment of the system, carrying out the operations upon receiving low-level commands from the LLM controller. 
To facilitate seamless communication and interoperability between different robotic units, our system employs ROS2 (Robot Operating System 2)~\citet{ros2} and ~\citet{ros2-composition} as the primary interface. The use of ROS2 not only enhances the flexibility of the system but also significantly contributes to our generalization capabilities by allowing easy integration and exchange of various robotic units. In our experiments, we use two robots: SCARA and DELTA robots, and image of them can be seen on the Appendix \ref{ap-sec:robot}. \textbf{DELTA robots~\citep{delta_pierrot_reynaud_fournier_1990}} excel in high-speed, precise pick-and-place operations due to their parallel kinematic structure and compact footprint. However, they have a limited range of motion and are primarily suited for vertical movements, sometimes making their kinematics complex. 
\textbf{SCARA robots~\citep{scara-9936755}} are adept at tasks demanding high repeatability and meticulous accuracy. Their compact and horizontal design makes them especially suitable for environments where space is at a premium, allowing for efficient operations without occupying excessive vertical room, unlike articulated robots. This specific design orientation means they naturally excel in horizontal movements, aligning with everyday tasks like pick-and-place, assembly, and inspection. We installed a gripper terminal on the tip of both robots to facilitate the object's manipulation. The detailed specification of both robots is presented in the Appendix \ref{ap-sec:robot}.





%% file: sec/3_experiments.tex
\section{Experiments}

\subsection{Robotic units setup}
In our experiments, we use the two types of robots mentioned above, a SCARA~\citep{scara-9936755} and a DELTA~\citep{delta_pierrot_reynaud_fournier_1990}. More information about these robots can be found in the Appendix \ref{ap-sec:robot}.
The same terminal end-effector, controller, and cameras have been used for both robots. Details can be found in the Appendix \ref{ap-sec:robot}.
Camera calibration is performed on both used cameras, the Logitech HD Pro C920 and the Razer Kiyo Pro, to understand the intrinsic and extrinsic parameters of the camera. The intrinsic parameters include focal length, sensor size, and pixel dimensions, while the extrinsic parameters involve the camera's pose relative to the robot coordinate system. Calibration is performed with the checkerboard calibration method~\citep{checkerboard} and ArUco~\citep{ROMERORAMIREZ201838} marker method that computes these parameters based on multiple captured images. Combined with these two methods, the calibration can be done with ChArUco~\citep{charuco}, also used for pose estimation. Additional details about the camera calibration can be found in the Appendix \ref{ap-sec:calibration}.

\subsection{Experimental environment}
All experiments are conducted in a controlled environment designed to mimic real-world conditions. The robotic system is tested on many scenarios ranging from simple, predictable tasks to complex, dynamic challenges. Unless otherwise stated, all examples and experiments in this paper use OpenAI gpt-3.5-turbo-0613~\citep{brown2020language}. Our experiments are bifurcated into two setups: static and dynamic. 

\textbf{Static setup.} These tasks are set in an environment where elements remain stationary and unchanged throughout the experiment. In Table~\ref{tab:st-exp} we include the list of tasks, these tasks aimed to gauge the robotic system's performance in a predictable setting, where all the scene is fully observable by the camera. Thus, the objective of this setup is to assess the robot's precision, adherence to the prompt, and its adaptability to changes in the prompt structure. The list of the static tasks is mostly taken from ~\citet{liang2023code}.

\textbf{Dynamic setup.} Dynamic tasks are characterized by a perpetually changing environment where the robot is compelled to react to moving or evolving elements in real time. This setup also contains tasks where not all the objects are present in the camera's field of view and as such require the robot to explore the environment prior to executing the task. Overall, the dynamic experiments, described in Table~\ref{tab:dyn-experiments}, were designed to challenge the robot's real-time processing, adaptability, and responsiveness. Both static and dynamic tasks are described in depth in the section \ref{ap-sec:task-description} of the Appendix. 


\textbf{Metrics.} We use the following metric to evaluate the static and dynamic setups:
\begin{itemize}
    \item \textbf{Success Rate (SR)}. SR measures the robot's task completion rate, expressed as a percentage; a score of 100\% is the maximum value.
    \item \textbf{Average Completion Time (ACT)}. Denotes the duration required for the robot to finalize a task after command reception, serving as a key indicator of operational efficiency.
\end{itemize}

\textbf{System hyperparameters.} The detailed hyperparameters can be found in the Appendix \ref{ap-sec:hyperparameters}.

\textbf{Baselines.} We use~\citet{liang2023code} as our baseline as similarly to our approach it leverages an LLM and enables zero-shot generalization to novel robotic tasks.

\subsection{Experimental results}
\label{experimental-results}

\begin{table*}
\centering
    \caption{Comparison between the results of the static experiments obtained by CaP and InCoRo (Ours) with both Mitsubishi Electric SCARA Robot and DELTA robot, including descriptions of the tasks used.}
    \label{tab:st-exp}
    \footnotesize 
\begin{tabular}{p{1cm}p{7cm}cc|cc|cc|cc}
\toprule
    \textbf{Task id} & \textbf{Task description} & \multicolumn{4}{c}{SCARA robot} & \multicolumn{4}{c}{DELTA robot} \\
    \cmidrule(r){3-6} \cmidrule(l){7-10}
    & & \multicolumn{2}{c}{CaP \citep{liang2023code}} & \multicolumn{2}{c}{InCoRo (Ours)} & \multicolumn{2}{c}{CaP \citep{liang2023code}} & \multicolumn{2}{c}{InCoRo (Ours)} \\
    \cmidrule(r){3-4} \cmidrule(l){5-6} \cmidrule(r){7-8} \cmidrule(l){9-10}
    & & SR $\uparrow$ & ACT $\downarrow$ & SR $\uparrow$ & ACT $\downarrow$ & SR $\uparrow$ & ACT $\downarrow$ & SR $\uparrow$ & ACT $\downarrow$ \\
    \midrule
    1 & Stack all the blocks & 0/25 & - & 23/25 & 43.8 s & 0/25 & - & 18/25 & 61.3 s \\
    2 & Put all the blocks on the $\langle corner/side \rangle$ & 6/25 & 12.2 s & 25/25 & 26.3 s & 7/25 & 12.8 s & 22/25 & 49.3 s \\
    3 & Put the blocks in the $\langle receptacle-bowl \rangle$ & 5/25 & 15.2 s & 22/25 & 29.1 s & 4/25 & 14.3 s & 18/25 & 55.2 s \\
    4 & Put all the blocks in the bowls with matching colors & 6/25 & 23.7 s & 24/25 & 51.5 s & 3/25 & 23.8 s & 19/25 & 76.2 s \\
    5 & Pick up the block to the $\langle direction \rangle$ of the $\langle receptacle-bowl \rangle$ and place it on the $\langle corner/side \rangle$ & 3/25 & 18.3 s & 23/25 & 24.2 s & 3/25 & 19.9 s & 20/25 & 49.1 s \\
    6 & Pick up the block $\langle distance \rangle$ to the $\langle receptacle-bowl \rangle$ and place it on the $\langle corner/side \rangle$ & 8/25 & 8.6 s & 24/25 & 14.1 s & 5/25 & 9.1 s & 21/25 & 48.8 s \\
    7 & Pick up the $\langle nth \rangle$ block from the $\langle direction \rangle$ and place it on the $\langle corner/side \rangle$ & 7/25 & 11.0 s & 21/25 & 19.8 s & 6/25 & 10.3 s & 19/25 & 49.2 s \\
    \midrule
    & \textbf{Overall} & 20\% & - & 92.6\% & - & 16\% & - & 78.3\% & - \\
    \bottomrule
    \end{tabular}
\end{table*} 

\begin{table*}
    \centering
    \footnotesize
    \caption{Comparison between the results of the dynamic experiments obtained by InCoRo (Ours) with both Mitsubishi Electric SCARA Robot and DELTA robot, including descriptions of the tasks used.}
    \label{tab:dyn-experiments}
    \begin{tabular}{p{1cm}p{10.4cm}p{1cm}p{1cm}|p{1cm}p{1cm}}
    \toprule
    \textbf{Task id} & \textbf{Task description} & \multicolumn{2}{c}{InCoRo (SCARA)} & \multicolumn{2}{c}{InCoRo (DELTA)} \\
    \cmidrule(r){3-4} \cmidrule(l){4-6}
    & & SR $\uparrow$ & ACT $\downarrow$ & SR $\uparrow$ & ACT $\downarrow$ \\
    \midrule
    8 & Follow and pick-up the ball & 24/25 & 13.3 s & 19/25 & 69.3 s \\
    9 & Go ahead and take the ball and then give it back to me (I am the hand) & 20/25 & 14.8 s & 15/25 & 53.9 s \\
    10 & Find and pick-up the $\langle round/any \rangle$ object and put it into the $\langle bucket/any location \rangle$ & 19/25 & 22.1 s & 16/25 & 68.6 s \\
    11 & Order everything in a logical way (Moving the objects during the experiment) & 20/25 & 28.6 s & 15/25 & 92.8 s \\
    12 & Give me the $\langle screwdriver/any object \rangle$ when you see my hand & 21/25 & 15.6 s & 17/25 & 63.2 s \\
    \midrule
    & \textbf{Overall} & 83.2\% & - & 65.6\% & - \\
    \bottomrule
    \end{tabular}
\end{table*}

\begin{table*}[t]
    \centering
    \caption{Ablation experiments. The experiments are performed on dynamic tasks with SCARA robot. The description of the tasks is presented in Table~\ref{tab:dyn-experiments}.}
    \label{tab:ablations}
    \footnotesize 
    \begin{tabularx}{0.9\textwidth}{p{1cm}cc|cc|cc|cc|cc|cc}
        \toprule
        \multirow{2}{*}{\textbf{Task id}} & \multicolumn{2}{c}{InCoRo} & \multicolumn{2}{c}{w/o TAPIR} & \multicolumn{2}{c}{w/o SAM} & \multicolumn{2}{c}{\parbox{2cm}{\centering LLM context ablation A}} & \multicolumn{2}{c}{\parbox{2cm}{\centering LLM context ablation B}} & \multicolumn{2}{c}{\parbox{2cm}{\centering LLM context ablation C}}\\
        \cmidrule(lr){2-3} \cmidrule(lr){4-5} \cmidrule(lr){6-7} \cmidrule(lr){8-9} \cmidrule(lr){10-11} \cmidrule(lr){12-13}
        & SR $\uparrow$ & ACT $\downarrow$ & SR $\uparrow$ & ACT $\downarrow$ & SR $\uparrow$ & ACT $\downarrow$ & SR $\uparrow$ & ACT $\downarrow$ & SR $\uparrow$ & ACT $\downarrow$ & SR $\uparrow$ & ACT $\downarrow$ \\
        \midrule
        8 & 24/25 & 13.3 s & 22/25 & 15.8 s & 23/25 & 16.2 s & 12/25 & 19.2 s & 13/25 & 18.1 s & 3/25 & 48.3 s\\
        9 & 20/25 & 14.8 s & 21/25 & 18.1 s & 15/25 & 15.3 s & 11/25 & 18.6 s & 13/25 & 19.9 s & 8/25 & 35.5 s\\
        10 & 19/25 & 22.1 s & 20/25 & 35.3 s & 11/25 & 26.5 s & 9/25 & 39.1 s & 5/25 & 36.5 s & 7/25 & 49.8 s\\
        11 & 20/25 & 28.6 s & 18/25 & 32.9 s & 5/25 & 62.3 s & 8/25 & 58.4 s & 6/25 & 76.6 s & 6/25 & 82.3 s\\
        12 & 21/25 & 15.6 s & 22/25 & 19.4 s & 9/25 & 59.7 s & 19/25 & 46.5 s & 8/25 & 58.6 s & 10/25 & 32.7 s\\
        \midrule
        \textbf{Overall} & 83.2\% & & 82.4\% & & 50.4\% & & 47.2\% & & 36.0\% & & 27.2\% & \\
        \bottomrule
    \end{tabularx}
\end{table*}

\begin{figure*}[t!]
    \centering
    \begin{subfigure}{0.23\textwidth}
        \includegraphics[width=\linewidth]{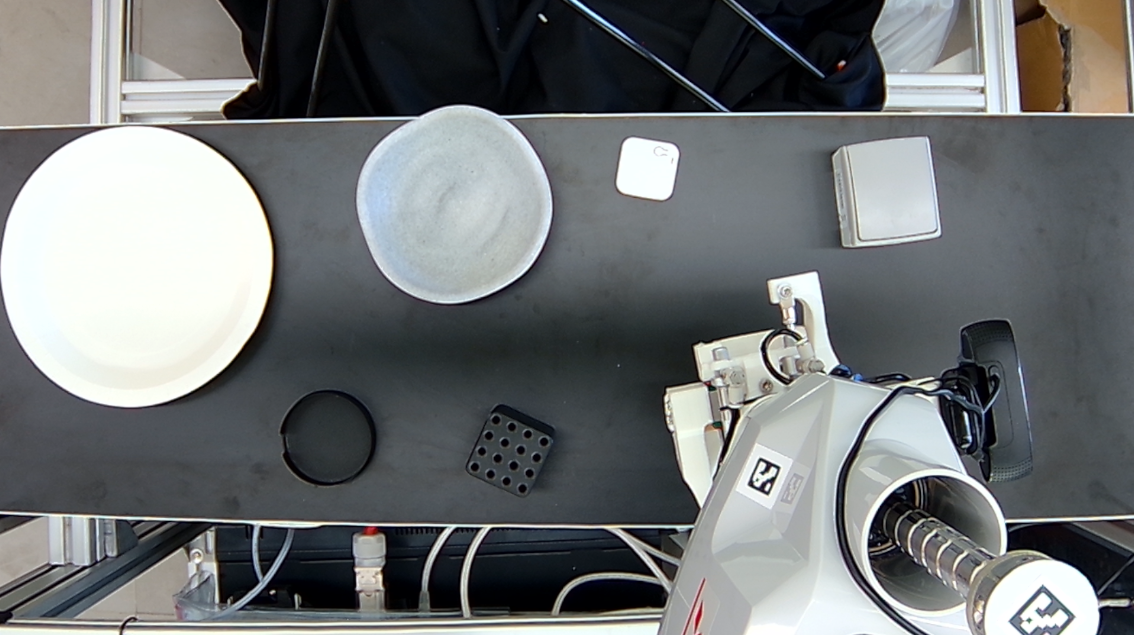}
        \caption{Start: Put all the blocks in the bowls with matching colors}
    \end{subfigure}
    \hfill 
    \begin{subfigure}{0.23\textwidth}
        \includegraphics[width=\linewidth]{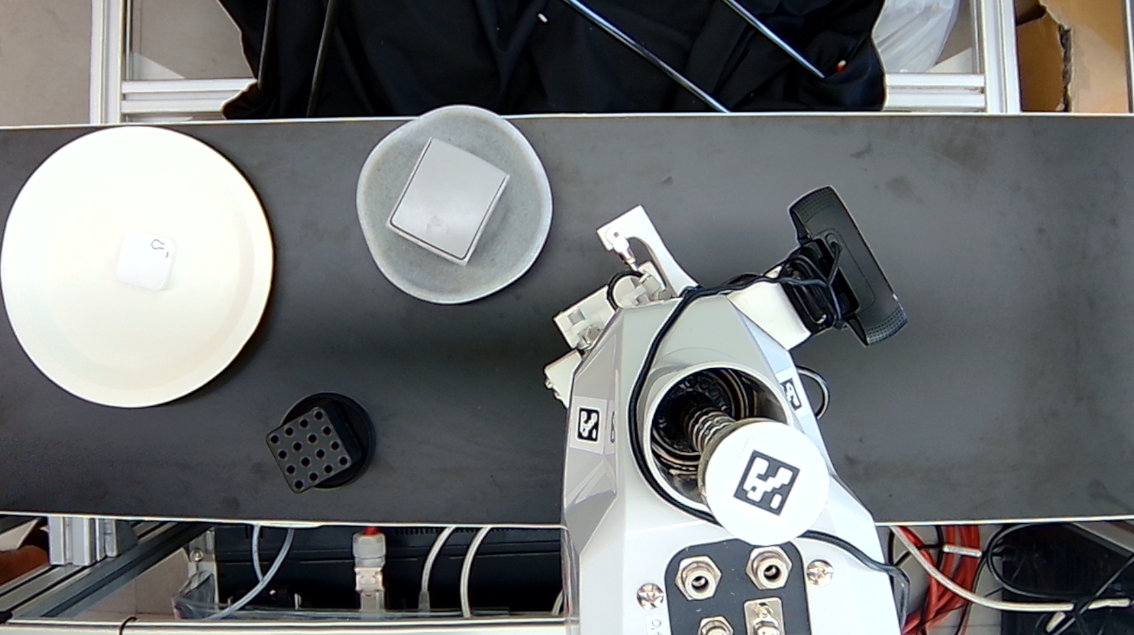}
        \caption{Stop: Put all the blocks in the bowls with matching colors}
    \end{subfigure}
    \hfill
    \vspace{0.5cm}
    \begin{subfigure}{0.23\textwidth}
        \includegraphics[width=\linewidth]{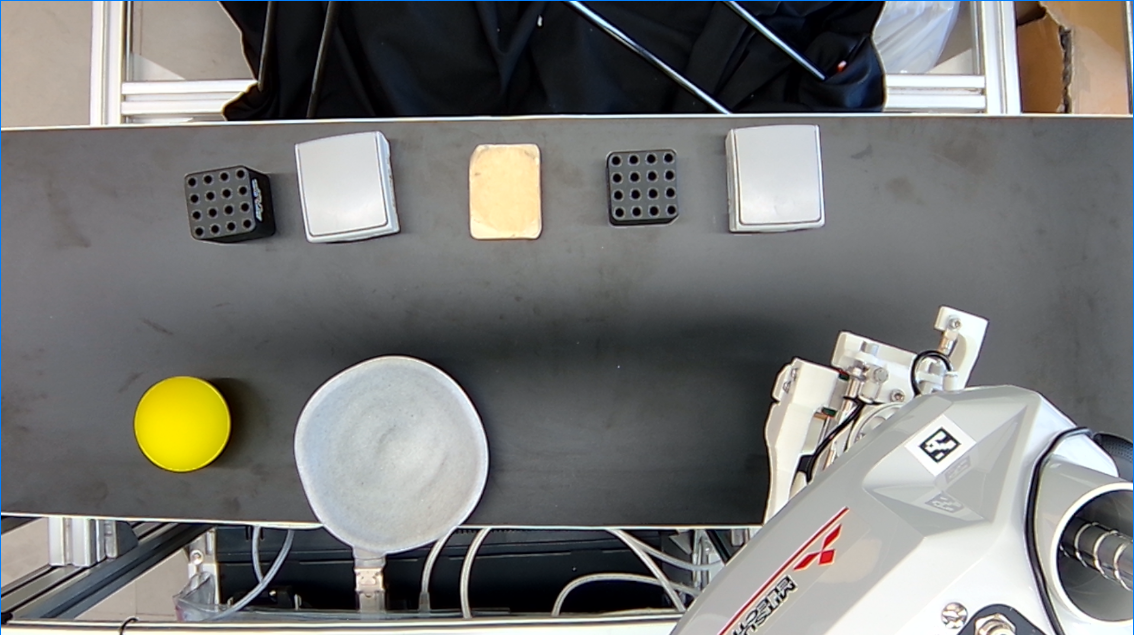}
        \caption{Start: Pick up the $3^{rd}$ closest block and place it on the corner}
    \end{subfigure}
    \hfill
    \begin{subfigure}{0.23\textwidth}
        \includegraphics[width=\linewidth]{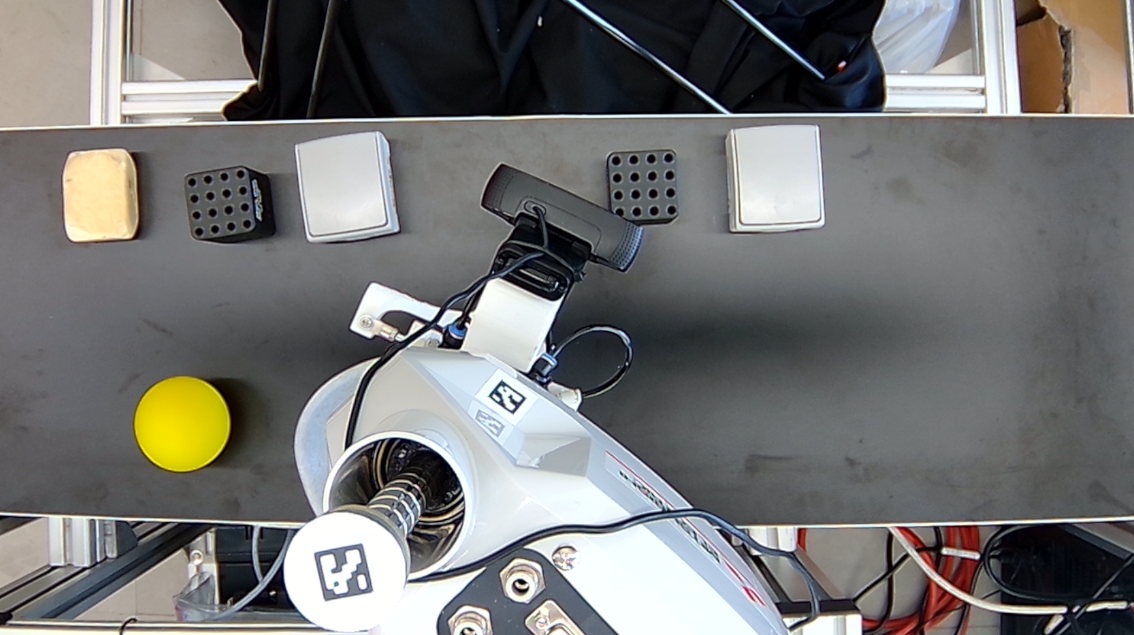}
        \caption{Stop: Pick up the $3^{rd}$ closest block and place it on the corner}
    \end{subfigure}

    \caption{\textbf{Static setup.} Visualization of the initial (Start) and the final (Stop) states for the DELTA robot and two tasks.}
    \label{fig:st-exp-delta}
\end{figure*}

\textbf{Static experiments.} Table~\ref{tab:st-exp} presents the results for static experiments. Figure \ref{fig:st-exp-delta} show some examples of static experiments for the SCARA robot.  
Employing the Mitsubishi Electric SCARA Robot, InCoRo significantly outperforms the CaP system with a 92.6\% success rate in task completion, a nearly five-fold increase over CaP's 20\%. InCoRo also maintains time efficiency despite its higher success rate, justifying slight increases in task duration due to its dynamic adjustments and feedback mechanisms. In an overarching perspective, employing a DELTA Robot InCoRo's performance is superior, with an aggregate success rate of 78.3\% compared to the 16\% success rate demonstrated by CaP. As with the SCARA, time metrics reveal that InCoRo typically requires a longer span to complete tasks than CaP, indicating a more cautious and effective error-mitigation strategy. Specifically, 0\% CaP's success rate Task 1 suggests that the task requires real-time adaptation and improvisation, which CaP lacks. 
In the Appendix \ref{ap-sec:static-experiments-fig} there are some qualitative figures and some videos of static experiments.

\begin{figure*}
    \centering
    \begin{subfigure}{0.28\textwidth}
        \includegraphics[width=\linewidth]{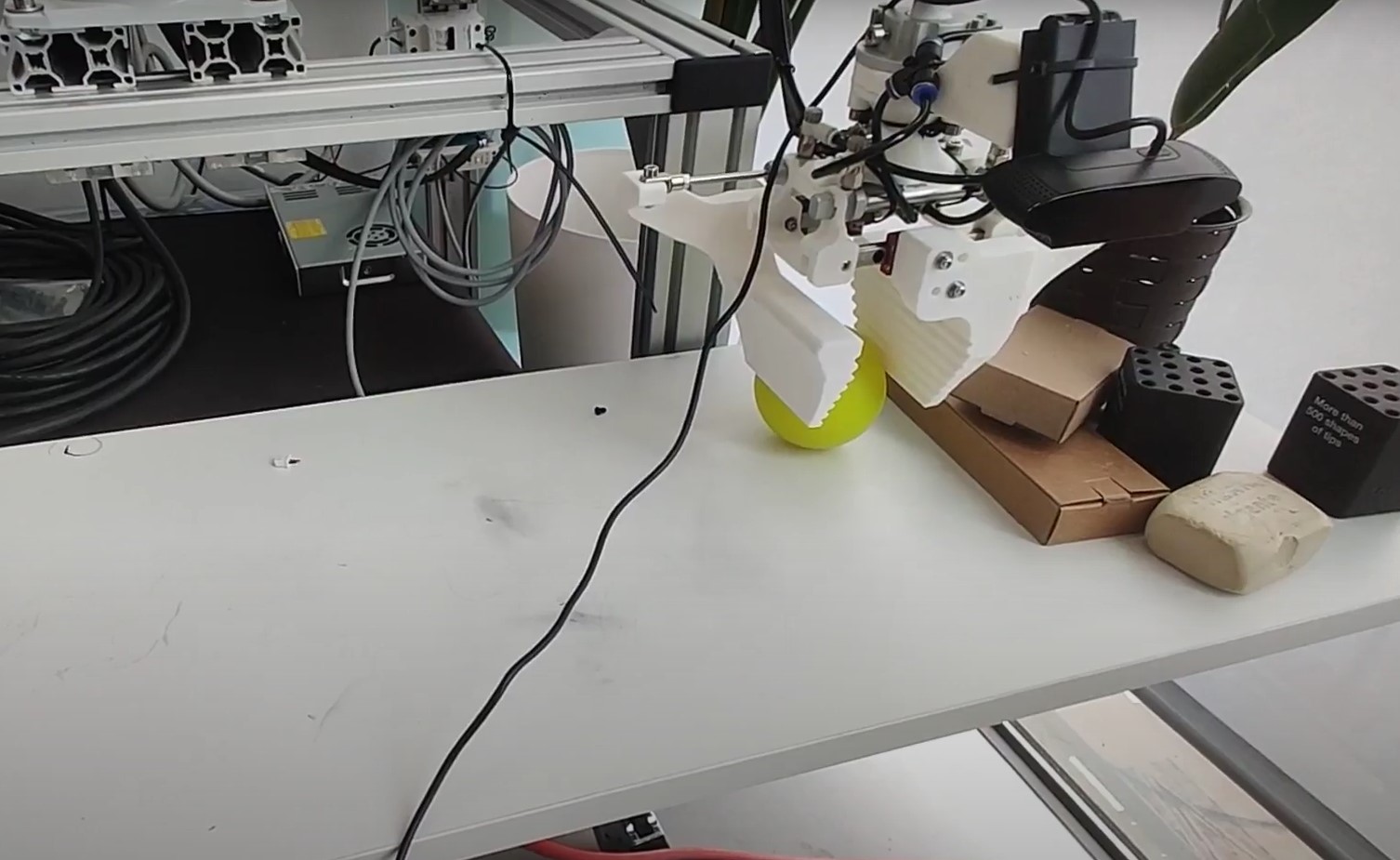}
        \caption{"Take the ball"}
    \end{subfigure}
    \hfill 
    \begin{subfigure}{0.28\textwidth}
        \includegraphics[width=\linewidth]{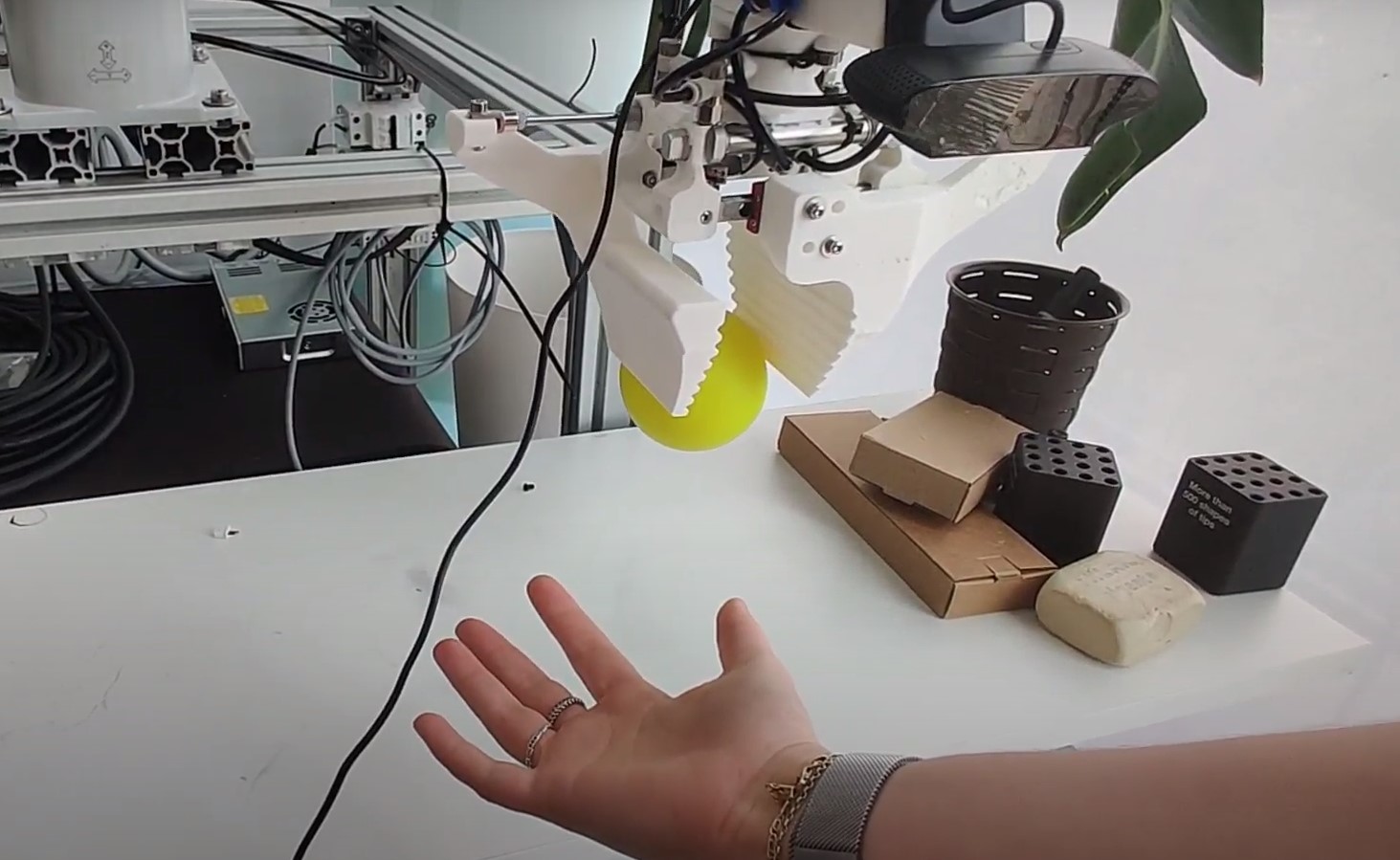}
        \caption{"Follow my hand"}
    \end{subfigure}
    \hfill
    \begin{subfigure}{0.28\textwidth}
        \includegraphics[width=\linewidth]{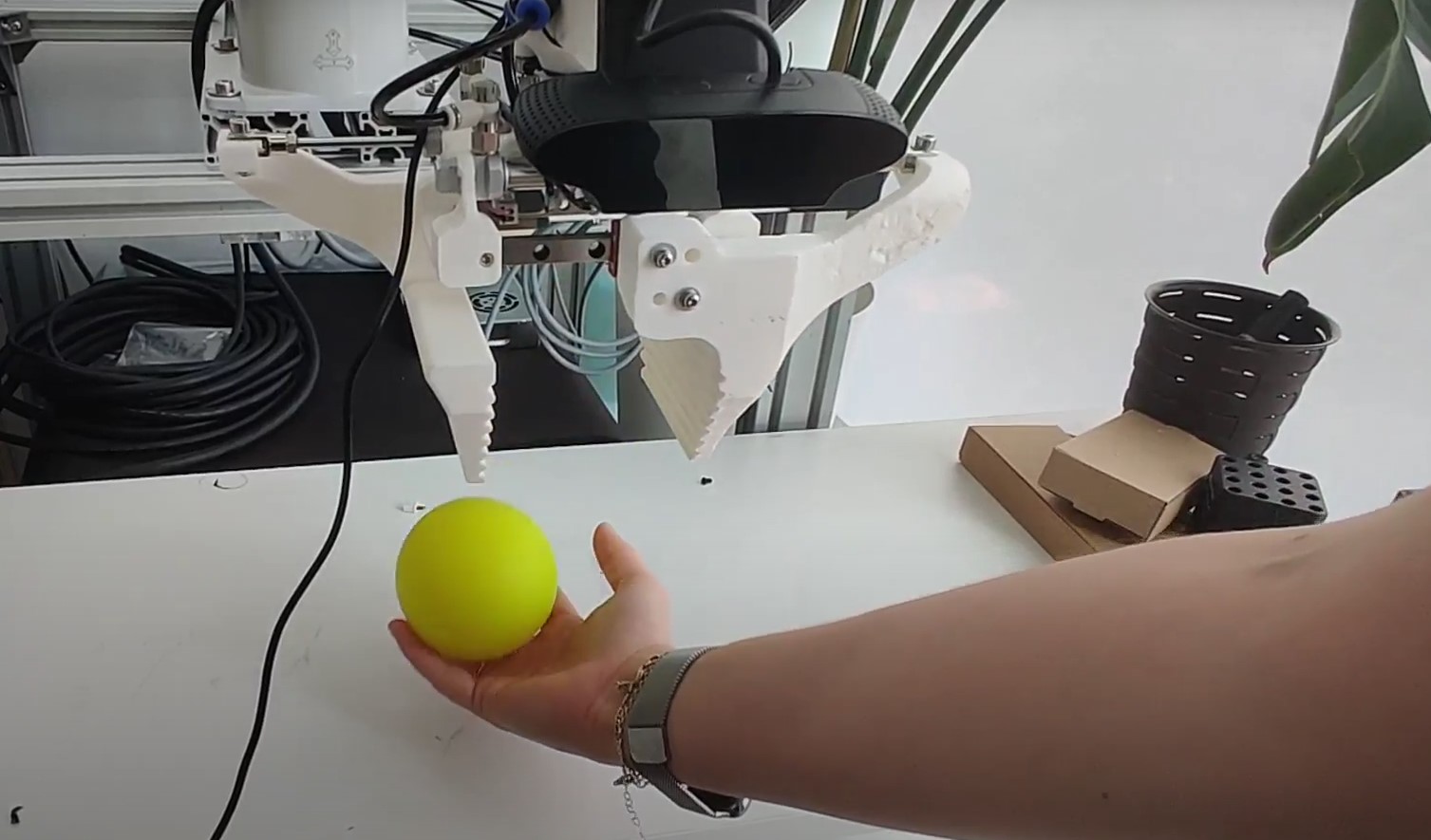}
        \caption{"Give it to me"}
    \end{subfigure}
    \caption{\textbf{Dynamic setup}. We depict three frames from three atomic actions when executing the following user instruction, "Go ahead, take the ball and then, give it back to my hand. The hand is moving and the robot must follow it". The images depict the DELTA robot.}
    \label{fig:dyn-exp1-delta}
\end{figure*}

\textbf{Dynamic experiments.} Figure \ref{fig:dyn-exp1-delta} show an example of a dynamic experiment for the SCARA robot. The results are shown in Table~\ref{tab:dyn-experiments}. 
Once again, InCoRo outshines CaP. The latter fails to resume any task successfully, clocking in at a 0\% success rate for all tasks. Overall, the SCARA Robot achieved a total success rate of 83.2\% and the DELTA reached the success rate of 65.6\%. 
It is important to note that InCoRo not only succeeds in completing the tasks but also in a time-efficient manner, indicating good balance between the execution speed and the reliability.
In the Appendix \ref{ap-sec:dynamic-experiments-fig} there are some qualitative figures and some videos of dynamic experiments. Despite high success rates in both static and dynamic, certain tasks, such as \emph{Stack all the blocks}, may take longer (43.8s), potentially due to the inherent challenges of estimating depth with 2D vision. In certain dynamic experiments, such as [10] and [12], the camera's field of view is limited, and not all objects are within its visual range. This necessitates the robot to engage in exploratory actions to locate the target or targets before executing the designated task. Such experimental setups can lead to increased completion times due to the time taken to locate objects. However, they significantly broaden the application scope in real-world environments, introducing a more versatile and adaptive robotic behavior.

\textbf{InCoRo vs CaP.} 
InCoRo's feedback mechanism allows for re-calculating trajectories based on feedback data, allowing high-precision tasks. When InCoRo makes a mistake, it can detect it and implement corrective measures. This may elongate the time required for task completion but significantly enhances reliability. This can be seen in Appendix \ref{ap-sec:dynamic-experiments-fig}. CaP, by contrast, lacks this dynamic adaptability, making it less effective in scenarios that demand high precision and reliability. In addition, while InCoRo might consume more time for task completion, this should not be interpreted as inefficiency but rather as a safeguard against mistakes and errors. The extra time is essential to ensure the task is executed correctly. We argue that system robustness is often more valuable than speed, especially in industrial applications where the cost of failure is significant.

\textbf{SCARA vs DELTA.} The operational range of DELTA robots varies with the z-axis due to their parallel kinematic structure, presenting path planning challenges as the workspace alters with height changes. The complexity of the DELTA robots' kinematic model, which requires synchronized joint movements, can lead to control errors and affect task accuracy. Initial experiments indicate that SCARA robots yield better results with a simpler 4-DOF system emphasizing planar motion than the more complex 3-DOF DELTA robots.
    

\textbf{Ablations.} Here, we ablate the importance of individual components in our control loop. We perform the ablations with the SCARA robot on dynamic setups. During the ablations, we alter (1) the perception of the system by either removing the segmentation -- \emph{SAM} -- or the tracking -- \emph{TAPIR} -- components, as well as (2) the in-context learning examples presented to the LLM controller. For the latter one, we remove examples of either the five basic robot movements -- \emph{LLM context ablation A} --, or the three picking up and moving object tasks -- \emph{LLM context ablation B} --, or the three examples of objects-environment interaction -- \emph{LLM context ablation C}. 
The ablation study in Table~\ref{tab:ablations} reveals distinct impacts of different components on the SCARA robot's performance. The \emph{SAM} component is identified as critically important, evidenced by a significant drop in success rate to 50.4\% from the baseline of 83.2\% when removed. This underlines SAM's crucial role in object recognition and scene understanding. In contrast, the \emph{TAPIR} tracking component, while contributing to the system's efficiency, shows a less critical impact with a slight decrease in success rate to 82.4\%. LLM context ablations demonstrate the high importance of specific task-related examples. Removal of basic movement examples (\emph{LLM context ablation A}) leads to a reduced success rate of 47.2\%, and this further decreases to 36.0\% when removing picking and moving object tasks (\emph{LLM context ablation B}). The most pronounced effect is observed in \emph{LLM context ablation C}, involving objects-environment interaction, where the success rate drastically drops to 27.2\%. These results highlight the LLM controller's reliance on varied and specific contextual examples for effective task execution in dynamic environments.




%% file: sec/IV_Related_work.tex
\section{Related work}

The interdisciplinary realm of robotics has witnessed substantial progress, especially at the convergence of robotics, language, and vision. This section delineates key contributions and methodologies on robotic control informed by language and vision. Our synthesis categorizes the literature into four predominant categories.

\textbf{Language-based Approaches:} Leveraging LLMs' recent advancements~\citep{wei2022emergent, dong2023survey, zhang2023language,verghese2023using,wu2023tidybot} facilitate the generation of executable instructions interpretable by robotic systems~\citep{saycan2022arxiv, chen2022leveraging,qtransformer}. Such methodologies capitalize on natural language, from task-specific instructions to more comprehensive policies~\citep{liang2023code, wu2023tidybot,10102606}. Integrating LLMs with robotics has fortified the translation of linguistic descriptors into robotic operations, heralding a paradigm where linguistic nuances inform robotic behavior~\citep{huang2023voxposer,chen2023sequential}.
    
\textbf{Vision-based Methodologies:} Exploiting visual information remains integral to robotic innovation~\citep{roboagent,caron2021emergingDINO,zhao2023fast,doersch2023tapir,abeyruwan2023agile,wang2023tracking,ke2023segment,kirillov2023segment,yang2023track,huang23avlmaps}. Techniques in this category predominantly focus on generating 3D spatial maps, harnessing the potential of multi-camera architectures, and devising intricate path-planning algorithms. Noteworthy among these is the methodology leveraging LLMs~\citep{tang2023saytap,saycan2022arxiv,zhao2023survey,wei2022emergent,zhang2023motiongpt,agia2023stap,zhou2023learning} to formulate 3D value maps based on linguistic directives~\citep{chen2022leveraging,huang2023voxposer,kulshrestha2023structural}, seamlessly melding advanced robotic operations with intuitive human-robot communication.
    
\textbf{Multi-modal Robotic Control Approaches:} A synergistic fusion of language and vision characterizes these methodologies, optimizing robotic responsiveness in dynamic environments. The vision-language-action~\citep{walke2023bridgedata,brown2020language,rt22023arxiv,chen2023rsprompter,roboagent,ha2023scaling,ze2023gnfactor} framework stands out, translating linguistic prompts into tangible robotic actions. Subsequent refinement through reinforcement learning algorithms ensures adept task execution, epitomizing linguistic and visual processing amalgamation~\citep{singh2022progprompt,cheng2023tracking,zhuang2023robot}.
    
\textbf{Imitation Learning Paradigms:} Grounded in observational learning, these approaches derive robotic protocols from demonstration data~\citep{walke2023bridgedata,bahl2023affordances,shi2023waypointbased,bousmalis2023robocat,caron2021emergingDINO,dalal2023imitating}. Such demonstrations, either originating from tangible robotic actions~\citep{hu2023causal} or synthesized via computational models like LLMs~\citep{lv2023parameter,shrivastava2023repofusion}, act as exemplars. Reinforcement learning strategies subsequently refine these robotic behaviors, ensuring fidelity in task replication and execution.

%% file: sec/5_Conclusions.tex
\section{Conclusions}
In this paper, we introduced a system that leverages in-context learning together with a feedback loop to guide the robot in executing complex task in real-world dynamic environments. Our system called, InCoRo, is built from a pre-processor unit, a control loop, and a robotic unit. The system's closed-loop feedback mechanism equipped with state-of-the-art scene perception allows dynamic responses to environmental changes, offering unparalleled adaptability and robustness in robotic systems. In our extended validation, we outperform prior art by a large margin and highlight InCoRo's effectiveness across both static and dynamic environments. Thus, our study marks a significant advancement in robotics, overcoming traditional limitations in robotic control, natural language processing, and visual tracking. 

Looking ahead, InCoRo's architecture opens up exciting avenues for future enhancements. A salient feature of InCoRo is its scalable action repertoire, allowing seamless integration of new dynamic actions without the prerequisite of expansive, multimodal datasets. This remarkable flexibility is attributed to the system's adept in-context learning capabilities and its versatile adaptive control loop. These features enable InCoRo to effectively interpret and respond to a diverse spectrum of scenarios and instructions, heralding a new era of flexibility and adaptability in robotic applications.

%% file: sec/X_suppl.tex
\newpage
\appendix

\setcounter{page}{1}

\etocsettocstyle{\section*{Appendix - InCoRo: In-Context Learning for Robotics Control with Feedback Loops}}{}

\localtableofcontents 

\section{Exemplary prompts to the LLM for both atomic action and object extractions.}
\addcontentsline{toc}{section}{Exemplary prompts to the LLM for both atomic action and object extractions}
\label{ap-sec:ex-prompts}

\begin{tcolorbox}[colback=green!5,colframe=green!40!black,title=Exemplary prompts]

Frame structure is: [commandType, , x, y, z,rotation, param1, param2]
\\
Specific robot limitations:
Max x: 400, Max y: 400, Max z: 150, Min x: 100, Min y: 0, Min z: 0 
\\
\textbf{Prompt1:} ``Move the grasp drawing a rectangle'' 
Sequence of Movements: [1, 80, 60, 110, 0, 0, 0], 
[1, 120, 60, 110, 0, 0, 0], [1, 120, 100, 110, 0, 0, 0], [1, 80, 100, 110, 0, 0, 0], [1, 80, 60, 110, 0, 0, 0],  
Execution Step:
[1, 86, 60, 110, 0, 0, 0]
\\
\textbf{Prompt2:} ``Move forward and turn right when you see the $<$object$>$.''
Sequence of Movements:
[1, 100, 100, 0, 110, 0, 0], [1, 150, 100, 0, 110, 0, 0], [380, 150, 100, 0, 110, 0, 0], , Input: $<$object$>$ at [(198, 365), (60, 778), (729, 536), (94, 256)], [1, 380, 220, 500, 0, 0, 0], [1, 380, 320, 0, 110, 0, 0],
Execution Step:
[1, 125, 100, 0, 110, 0, 0]
\\
\textbf{Prompt3:} ``Execute the most efficient figure-eight pattern, making a complete stop at the $<$object$>$.''
Sequence of Movements:
[1, 100, 100, 0, 0, 0, 0], [1, 200, 200, 0, 0, 0, 0], [1, 300, 100, 0, 0, 0, 0],
[1, 200, 200, 0, 0, 0, 0], [1, 100, 300, 0, 0, 0, 0], [1, 100, 100, 0, 0, 0, 0], [1, 200, 200, 0, 0, 0, 0], [1, 300, 100, 0, 0, 0, 0], [1, 200, 200, 0, 0, 0, 0], 
[1, 100, 300, 0, 0, 0, 0], Input: $<$object$>$ at ([(150, 150), (250, 250), (350, 350), (100, 200)], 
Execution Step:
[1, 150, 250, 0, 0, 0, 0], [1, 150, 250, 0, 0, 0, 0]
\\
\textbf{Prompt4:} ``Move in an L-shape and lower when  the $<$object$>$ appears at the end.''
Sequence of Movements:
[1, 110, 600, 600, 0, 0, 0], [1, 110, 600, 620, 0, 0, 0], [1, 110, 620, 620, 0, 0, 0], Input: $<$object$>$ at ([(150, 150), (250, 250), (350, 350), (100, 200)], [1, 480, 620, 620, 0, 0, 0], [1, 110, 620, 620, 40, 0, 0] Execution Step: [1, 110, 600, 610, 0, 0, 0],
\\
\textbf{Prompt5:} ``Trace the perimeter of an equilateral pentagon, pausing at the $<$object$>$.''
Sequence of Movements:
[1, 100, 100, 0, 0, 0, 0], [1, 300, 100, 0, 0, 0, 0], [1, 350, 275, 0, 0, 0, 0], [1, 200, 400, 0, 0, 0, 0], [1, 50, 275, 0, 0, 0, 0], [1, 100, 100, 0, 0, 0, 0], [1, 50, 275, 0, 0, 0, 0], [1, 100, 100, 0, 0, 0, 0], Input: $<$object$>$ at ([(150, 225), (275, 175), (325, 275), (250, 350), (125, 325)], Execution Step:
[1, 100, 175, 0, 0, 0, 0] 
\\
\textbf{Prompt6:} ``Move incrementally and pick-up the $<$object$>$'' 
Sequence of Movements:
[1, 80, 60, 0, 0, 0], [1, 90, 800, 0, 0, 0], [1, 100, 120, 0, 0, 0], Input: $<$object$>$ at ([(150, 150), (250, 250), (350, 350), (100, 200)], [1, 110, 140, 0, 0, 0], [1, 150, 160, 60, 0, 0], [1, 150, 160, 60, 1, 0],[1, 150, 160, 0, 0, 0] 
Execution Step: 
[1, 360, 86, 80, 0, 0, 0]
\\
\textbf{Prompt7:} ``Move, When you see the $<$object$>$ Stop and pick it''
Sequence of Movements:
[1, 100, 50, 0, 0, 0, 0], [1, 150, 50, 0, 0, 0, 0], Input: $<$object$>$ at ([(150, 150), (250, 250), (350, 350), (100, 200)], [1, 200, 50, 0, 0, 0, 0], [1, 200, 50, 60, 0, 0, 0], [1, 200, 50, 60, 1, 0, 0], [1, 200, 50, 0, 0, 0, 0]
Execution Step:
[1, 125, 50, 0, 0, 0, 0]
\\
\textbf{Prompt8:} ``Move incrementally in zigzag pattern; when you see the 5th $<$object5$>$, stop and pick it up''
Sequence of Movements:
[1, 80, 60, 0, 0, 0, 0], [1, 90, 80, 0, 0, 0, 0], Input: $<$object3$>$ at ([(150, 150), (250, 250), (350, 350), (100, 200)], [1, 100, 120, 0, 0, 0, 0], [1, 130, 120, 0, 0, 0, 0], [1, 150, 150, 0, 0, 0, 0], [1, 160, 160, 0, 0, 0, 0], [1, 170, 170, 0, 0, 0, 0], Input: $<$object5$>$ at [(110, 90), (130, 110), (150, 150), (160, 160), (170, 170)], [1, 170, 170, 60, 0, 0, 0], [1, 170, 170, 60, 1, 0, 0], [1, 170, 170, 0, 0, 0, 0] Execution Step:
[1, 86, 72, 0, 0, 0, 0]
\\
\textbf{Prompt9:} ``Pick up the $<$object1$>$ and put it inside $<$object2$>$'' 
Sequence of Movements:
[1, 90, 80, 110, 0, 0, 0], [1, 93, 110, 110, 0, 0, 0], [1, 160, 92, 110, 0, 0, 0], Input: $<$object3$>$ at ([(150, 150), (250, 250), (350, 350), (100, 200)], [1, 160, 92, 110, 150, 0, 0],[1, 160, 92, 110, 150, 0, 1],[1, 160, 92, 110, 0, 0, 0], Input:$<$object$>$at ([(150, 150), (250, 250), (350, 350), (100, 200)], [1, 170, 68, 0, 0, 0], [1, 180, 68, 0, 0, 0], [1, 180, 68, 0, 0, 1], 
Execution Step: 
[1, 92, 90, 110, 0, 0, 0]
\\
\textbf{Prompt10:} ``Pick-up and use the $<$object1$>$ to cut the $<$object2$>$ in two parts''
Sequence of Movements:
[1, 90, 80, 110, 0, 0, 0], [1, 93, 110, 110, 0, 0, 0], Input: $<$object1$>$ at [(150, 150), (250, 250), (350, 350), (100, 200)], [1, 160, 92, 110, 150, 0, 0], [1, 160, 92, 110, 150, 0, 1], [1, 160, 92, 110, 0, 0, 0], Input: $<$objec2t$>$ at [(150, 150), (250, 250), (350, 350), (100, 200)], [1, 180, 68, 150, 0, 0], [1, 172, 68, 0, 0, 1], [1, 165, 68, 0, 0, 1], [1, 160, 68, 0, 0, 1]
Execution Step: 
[1, 91, 100, 110, 0, 0, 0]
\\
\textbf{Prompt 11:} ``Pick-up and put the $<$object1$>$ next to $<$object2$>$'' Sequence of Movements:
[1, 190, 80, 110, 0, 0, 0], [1, 293, 110, 110, 0, 0, 0], Input: $<$object1$>$ at [(150, 150), (250, 250), (350, 350), (100, 200)], [1, 160, 92, 110, 0, 0, 0], [1, 160, 92, 110, 0, 0, 1], [1, 160, 92, 110, 0, 0, 0], Input: $<$object2$>$ at [(150, 150), (250, 250), (350, 350), (100, 200)], [1, 172, 68, 150, 0, 0, 0], [1, 175, 70, 150, 0, 0, 0], [1, 175, 70, 150, 0, 0, 1], [1, 175, 70, 150, 0, 0, 0]
Execution Step: 
[1, 91, 100, 110, 0, 0, 0],


\end{tcolorbox}

\section{Examples of low-level actions}
\label{ap-sec:low-level}

\begin{figure*}[h]
  \centering
  \includegraphics[width=0.9\linewidth]{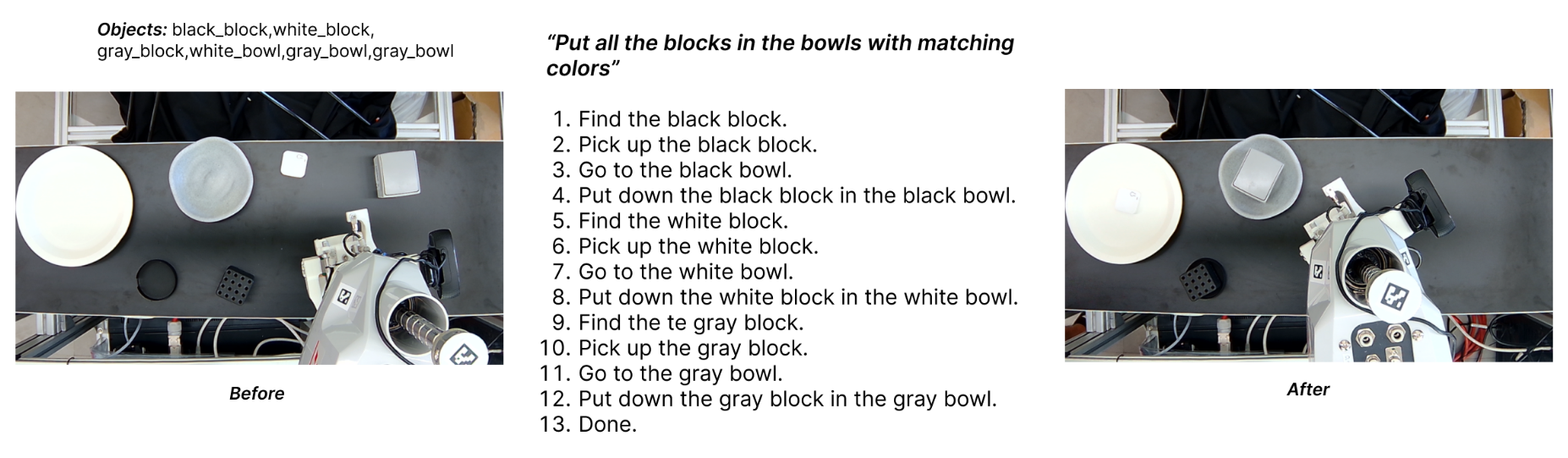}
  \caption{\textbf{Example of a subdivision into smaller and simpler low level actions and subtasks.} The LLM controller divides the overall task into smaller subtasks. The system will progressively execute each subtask until all of them have been accomplished.}
  \label{fig:tasks}
\end{figure*}

\begin{tcolorbox}[colback=green!5,colframe=green!40!black,title=Task: “Pick-up and put the $\langle object1\rangle$ next to $\langle object2\rangle$”]

You are the robot, I'm the User, learn what the task does and when I ask you, do the same, don't explain the subtask, just tell which is it. Sequence the actions I would do to accomplish the task:\\
\textbf{User:} How would you take the spoon from the drawer and put it in the dish rack? \\
\textbf{Robot:} 1. find the spoon, 2. pick up the spoon, 3. go to the dish rack, 4. put down the spoon, 5. done.\\ 
\textbf{User:} How would you take a bowl and a plate from the cabinet and place them on the dining table? \\
\textbf{Robot:} 1. find a bowl, 2. pick up the bowl, 3. go to the dining table, 4. put down the bowl, 5. find a plate, 6. pick up the plate, 7. go to the dining table, 8. put down the plate, 9. done. \\
\textbf{User:} How would you move the bread from the kitchen counter to the bread box? \\
\textbf{Robot:} 1. find the bread, 2. pick up the bread, 3. go to the bread box, 4. put down the bread, 5. done.\\ 
\textbf{User:} How would you bring me an apple and a bottle of water? \\
\textbf{Robot:} 1. find an apple, 2. pick up the apple, 3. find a bottle of water, 4. pick up the bottle of water, 5. bring them to you, 6. put down the apple, 7. put down the bottle of water, 8. done. \\
\textbf{User:} How would you pour a cup of coffee for me? \\
\textbf{Robot:} 1. find a coffee pot, 2. pick up the coffee pot, 3. find a cup, 4. pick up the cup, 5. pour coffee into the cup, 6. put down the coffee pot, 7. bring the cup to you, 8. put down the cup of coffee, 9. done. \\
\textbf{User:} How would you move the milk jug from the fridge to the kitchen counter? \\
\textbf{Robot:} 1. find the milk jug, 2. pick up the milk jug, 3. go to the kitchen counter, 4. put down the milk jug, 5. done. 
\textbf{User:} How would you dispose of a soda can?\\ 
\textbf{Robot:} 1. find a soda can, 2. pick up the soda can, 3. go to the trash can, 4. put down the soda can, 5. done.

\tcblower
Output: 

1.find $\langle object1\rangle$
2.pick up $\langle object1\rangle$
3.find $\langle object2\rangle$
4.go next to $\langle object2\rangle$
5.put down $\langle object1\rangle$
6.done

\end{tcolorbox}

Example of how the system divides the main task into different subtasks: the input is the task and the output is the subtask. The example context provided makes it possible to divide tasks into subtasks effectively.


\section{Tasks examples}
\label{ap-sec:examples-learning}

\begin{figure*}
  \centering
  \includegraphics[width=0.5\linewidth]{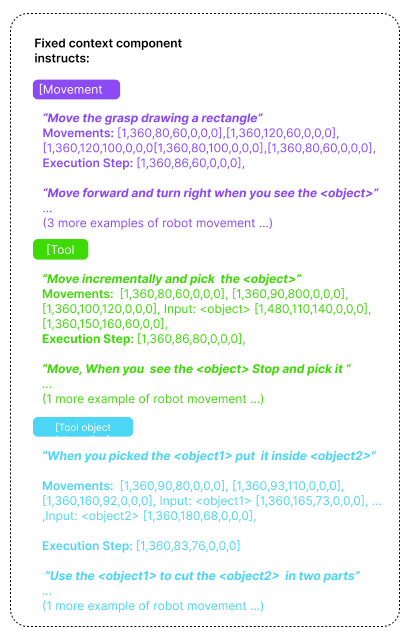}
  \caption{\textbf{Task examples of the LLM Controller.} The fixed context component is part of the LLM controller in our InCoRo system, serving as an instructional guide for in-context learning. It includes examples across three categories: five general robot movements, three tool-use techniques, and three object-to-object interactions. These examples collectively provide the system with a foundational understanding of various operations, enabling it to adapt and perform tasks autonomously in dynamic settings.}
  \label{fig:prompt_fixed}
\end{figure*}

Examples of our in-context learning examples can be seen in figure~\ref{fig:prompt_fixed}.

\newpage
\section{Pre-execution filter}
\label{ap-sec:pre-filter}

A raw example of one of the implemented system versions in the fixed context is shown on algorithm \ref{alg:filter}.

\begin{algorithm}
\caption{Pre-Execution Filter for Robotic Commands}
\label{alg:filter}
\begin{algorithmic}[b]
\Procedure{PreExecutionFilter}{Queue commandQueue, LLM lowLevelModule}
\State \textbf{Input:} Input Robot Frame \textbf{I}
\State Create empty list \( \text{flaggedCommands} \)
    \State \( \text{isValid} \leftarrow \text{True} \)
    \If{not \( \text{CheckStructure}( \text{\textbf{I}} ) \)}
        \State \( \text{isValid} \leftarrow \text{False} \)
    \EndIf
    \Comment{Check for Structural Inconsistencies, the frame should be in the form \textbf{([1, Z, X, Y, A1, A2, A3])}}
    \If{not \( \text{CheckHardwareConstraints}( \text{\textbf{I}} ) \)}
        \State \( \text{isValid} \leftarrow \text{False} \)
    \EndIf
    \Comment{Check that the frame does not exceed the limits of the robot and the working area.}
    \If{not \( \text{CheckTaskAppropriateness}( \text{\textbf{I}} ) \)}
        \State \( \text{isValid} \leftarrow \text{False} \)
    \EndIf
    \Comment{Check that the current frame is not too different from the previous one to avoid large movement differences.}
    \If{not \( \text{isValid} \)}
        \State Add \( \text{command} \) to \( \text{flaggedCommands} \)
        \State \( \text{RecycleToLLM}( \text{\textbf{I}}, \text{lowLevelModule} ) \)
    \EndIf
    
\EndProcedure
\end{algorithmic}
\end{algorithm}

\section{Robots specifications}
\label{ap-sec:robot}

\begin{figure}
  \centering
  \begin{subfigure}[b]{0.45\textwidth}
    \centering
    \includegraphics[width=0.71\textwidth]{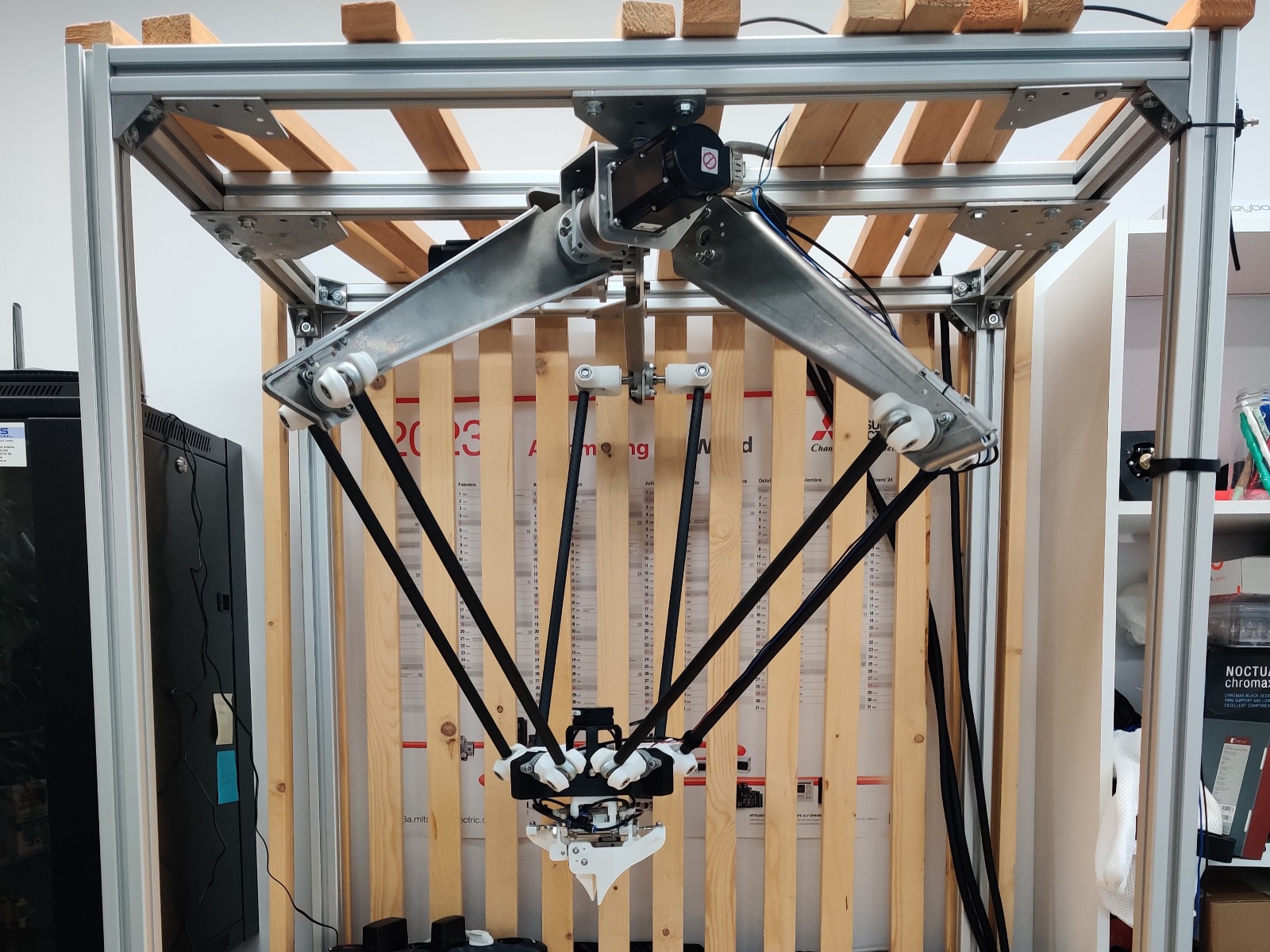}
    \caption{DELTA Robot Setup}
  \end{subfigure}
  \hfill 
  \begin{subfigure}[b]{0.45\textwidth}
    \centering
    \includegraphics[width=0.71\textwidth]{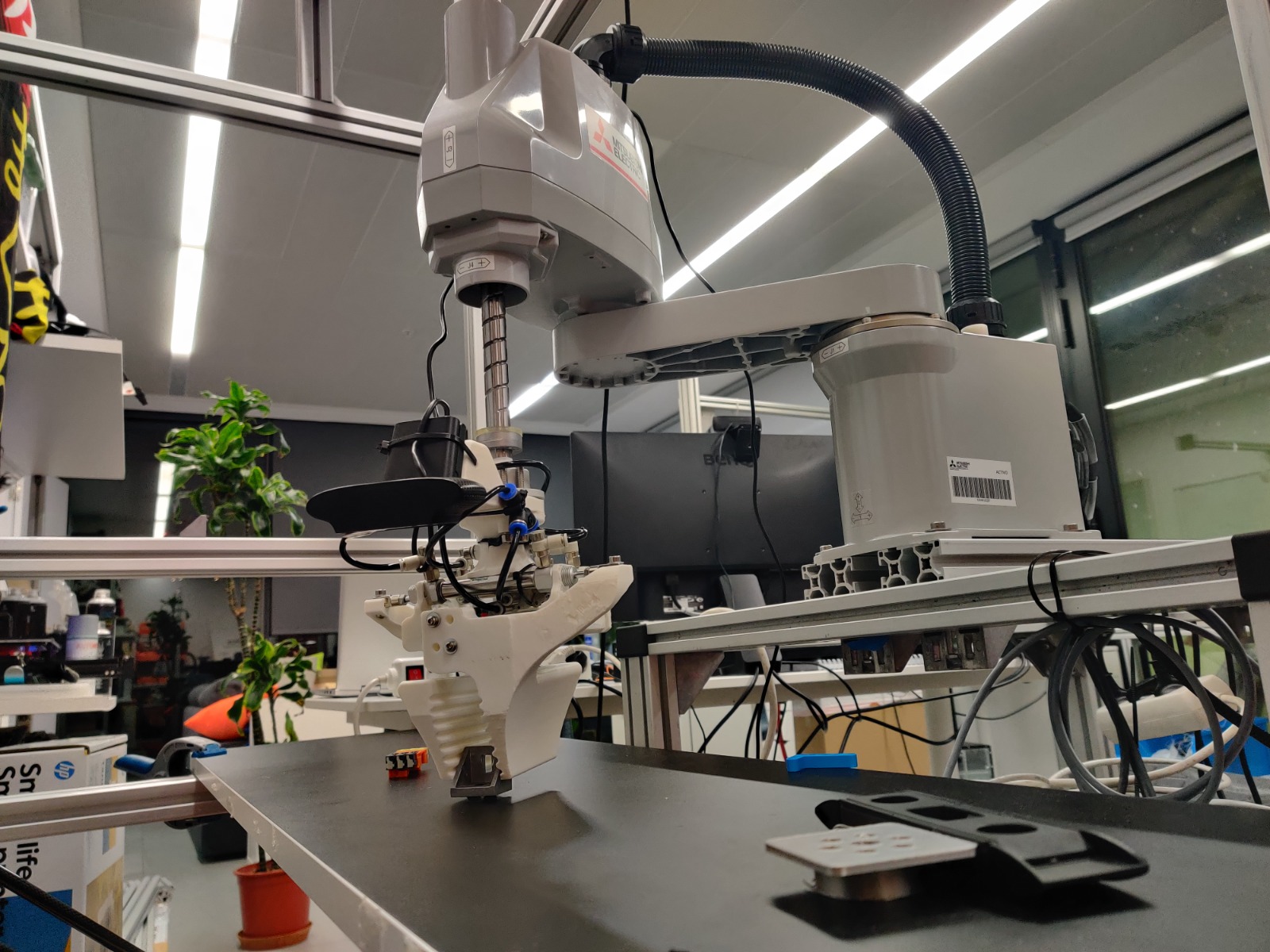}
    \caption{SCARA Robot Setup}
  \end{subfigure}
  \caption{\textbf{The images showcase the two distinct robotic setups.} In both configurations, there is a workspace with different objects with which the robot will interact.}
  \label{fig:delta-scara-setup}
\end{figure}

In figure \ref{fig:delta-scara-setup}, both robots can be seen.

\subsection{Robotic units and common specifications}
The DELTA robot is mounted on the roof of an aluminum extruded profile structure to allow it to reach its maximum operational range. The work table is located directly below. The robot workspace has a radius of 500 mm in the X and Y axes with a height range of 200 mm in the Z axis. It comes standard with three axes, and a fourth axis is added when the terminal end-effector is integrated into the system. The total payload is 5 kg. 
The SCARA robot in our setup originates from Mitsubishi Electric's RH-CRH series and is securely mounted on a base made of extruded aluminum profiles. This base is directly attached to the work table, allowing the robot to exploit its full operational range. The available operating surface for the robot has a radius of 400 mm in the X and Y axes and a vertical range of 180 mm in the Z axis. This geometry offers a sufficiently large workspace for the robot to conduct various tasks. The robot comes standard with four axes, and a fifth axis is added when the terminal end-effector is integrated into the system. With a payload capacity of 3 kg, the robot can handle a diverse array of objects and tools, aligning well with its intended application scenarios. 

SCARA robots, in contrast to DELTA robots, specialize in horizontal tasks like assembly and have simpler mechanics. They offer consistent repeatability and high rigidity in vertical operations, but they might not be as swift as DELTA robots and can occupy more space. 

We wanted to conduct experiments with both robots because of their differences and to demonstrate that InCoRo can generalize.

The same terminal end-effector, controller, and cameras have been used for both robots. 
The terminal end-effector comprises a basic gripper optimized to handle objects up to dimensions of 100 mm x 100 mm. It has been designed to provide sufficient gripping force while maintaining adaptability to various object shapes and sizes. A linear encoder is employed to measure the gripper jaws' position accurately. It transforms the linear position of an optical scale into a set of digital signals that represent the position. This technology lets us know whether the object has been properly caught. The low-level controller accepts commands through a syntax that specifies the target positions and orientations for each robotic coordinate and the gripper. The precision of the low-level controller is exceptionally high, specified as a double-precision real number with an accuracy up to 10$^{-9}$. The robot has two cameras, allowing it to continuously capture its environment and stream this visual data back into the system. Two different models of commercial webcams have been used to corroborate that the system doesn't need to use industrial or high-performance cameras to operate correctly. Depending on the experiment case, either webcams or only one webcam is used. The Logitech HD Pro C920 is placed on top of the robot and has a closer view of the scene. The Razer Kiyo Pro camera is located above the robot, 1 meter from the table, and has a working range of 100 cm. The Logitech HD Pro C920 has a resolution of 1080p at 30 fps, while the Razer Kiyo Pro has a resolution of 1080p at 60 fps.

\section{Calibration details}
\label{ap-sec:calibration}
\begin{figure}[h]
  \centering
  \hfill
  \begin{subfigure}[b]{0.45\textwidth}
    \centering
    \includegraphics[width=0.71\textwidth]{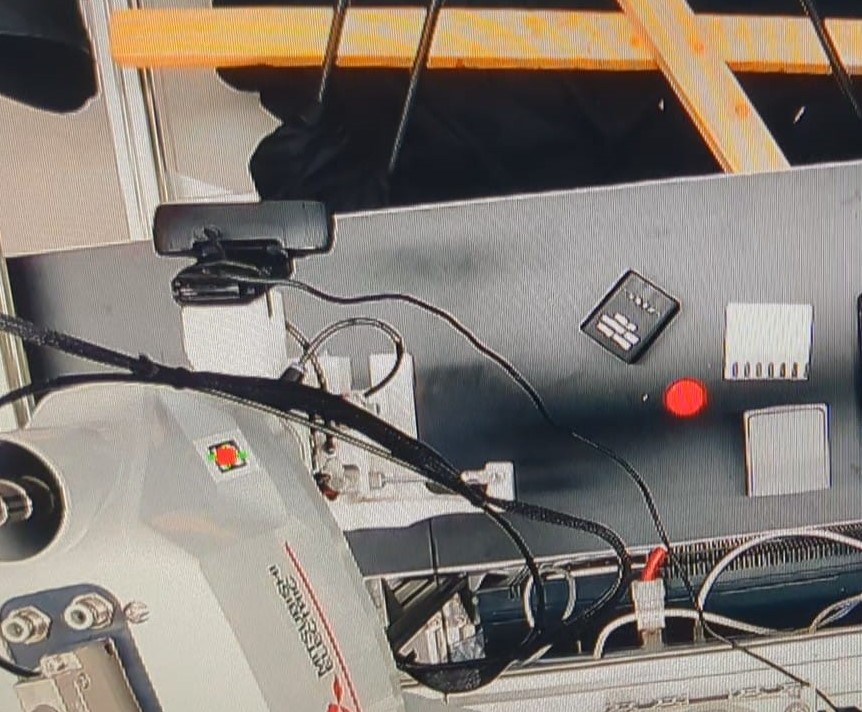}
    \caption{ArUco marker calibration method}
    \label{fig:image2}
  \end{subfigure}
  \hfill
  \begin{subfigure}[b]{0.45\textwidth}
    \centering
    \includegraphics[width=0.71\textwidth]{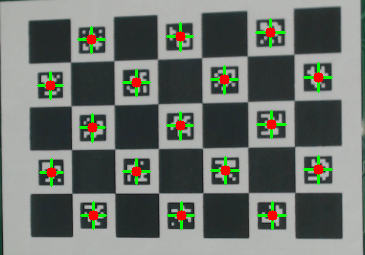} 
    \caption{ChArUco calibration method}
    \label{fig:image3}
  \end{subfigure}
  \caption{\textbf{Calibration process.} This triple approach capitalizes on the geometric rigor of Checkerboards for corner detection and the unique identification attributes of ArUco markers, offering a more robust and versatile framework  for accurate calibration of cameras to the environment and robot.}
  \label{fig:calibration}
\end{figure}

In figure \ref{fig:calibration}, the calibration can be seen.

\section{Tasks Description}
\label{ap-sec:task-description}

Tables \ref{tab:task-descriptions} and \ref{tab:dynamic-task-descriptions}, describe the different performed tasks.

All scenarios simulates real-world conditions where unpredictability is a given.

\begin{table}[h]
\centering
\caption{Descriptions of the Tasks Used in the Static Experiments}
\label{tab:task-descriptions}
\begin{tabular}{|c|p{12cm}|}
\hline
\textbf{Task ID} & \textbf{Description} \\
\hline
1 & Stack all the blocks. This task involves stacking a set of blocks, testing the robot's precision and stability in stacking. \\
\hline
2 & Put all the blocks on the specified corner/side. This task assesses the robot's ability to accurately place objects in a specific location on a corner or side of a workspace. \\
\hline
3 & Put the blocks in the specified receptacle-bowl. The robot must place blocks into a designated receptacle or bowl, testing precision in placing objects within confined spaces. \\
\hline
4 & Put all the blocks in the bowls with matching colors. A more complex task requiring the robot to identify the color of blocks and match them with bowls of the same color, evaluating color recognition and sorting capabilities. \\
\hline
5 & Pick up the block to the specified direction of the receptacle-bowl and place it on the specified corner/side. This task tests spatial awareness and precision in movement. \\
\hline
6 & Pick up the block at a specified distance to the receptacle-bowl and place it on the specified corner/side. Similar to Task 5 but focuses on the robot's ability to gauge distances accurately. \\
\hline
7 & Pick up the $n^{th}$ block from the specified direction and place it on the specified corner/side. This task tests both sequencing ability and precise placement. \\
\hline
\end{tabular}
\end{table}

\begin{table}
\centering
\caption{Descriptions of the Tasks Used in the Dynamic Experiments}
\label{tab:dynamic-task-descriptions}
\begin{tabular}{|c|p{12cm}|}
\hline
\textbf{Task ID} & \textbf{Description} \\
\hline
8 & Follow and pick-up the ball. This task involves tracking a moving ball and successfully picking it up, testing the robot's tracking and precision in a dynamic environment. \\
\hline
9 & Go ahead and take the ball and then give it back to me (I am the hand). The ball is changing its position, it's moving. This task requires the robot to take a ball from a moving hand and then return it, assessing its interaction and hand-off capabilities. \\
\hline
10 & Find and pick-up the $\langle round/any \rangle$ object and put it into the $\langle bucket/any location \rangle$. This task tests the robot's ability to identify, pick up a moving object, and place it in a designated location. \\
\hline
11 & Order everything in a logical way (Moving the objects during the experiment). The robot must organize objects logically while they are being moved, challenging its decision-making and adaptability in a dynamic setting. \\
\hline
12 & Give me the $\langle screwdriver/any object \rangle$ when you see my hand. This task requires the robot to identify an object and hand it over upon visual recognition of a hand, testing its reaction time and precision in dynamic object hand-off. \\
\hline
\end{tabular}
\end{table}

\newpage
\section{Hyperparameters}
\label{ap-sec:hyperparameters}

We have used the recommended hyperparameters in each of the models to demonstrate that the system is generalisable without having to be adapted.
\section{Qualitative figures of experiments}

In this Appendix, there are some qualitative figures of the diferent experiments.

\subsection{Static}
\label{ap-sec:static-experiments-fig}
The videos have been uploaded separately, directly on the OpenReview platform.

\begin{figure}[h]
    \centering
    \begin{subfigure}{0.32\textwidth}
        \includegraphics[width=\linewidth]{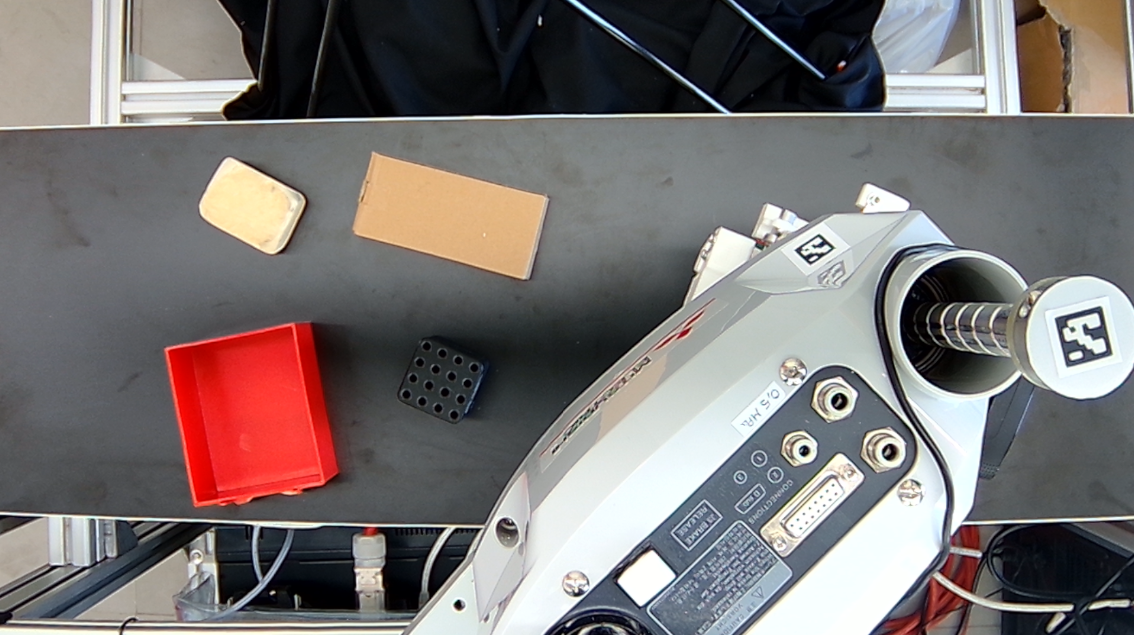}
        \caption{Stack all the blocks (initial state)}
    \end{subfigure}
    \hfill 
    \begin{subfigure}{0.32\textwidth}
        \includegraphics[width=\linewidth]{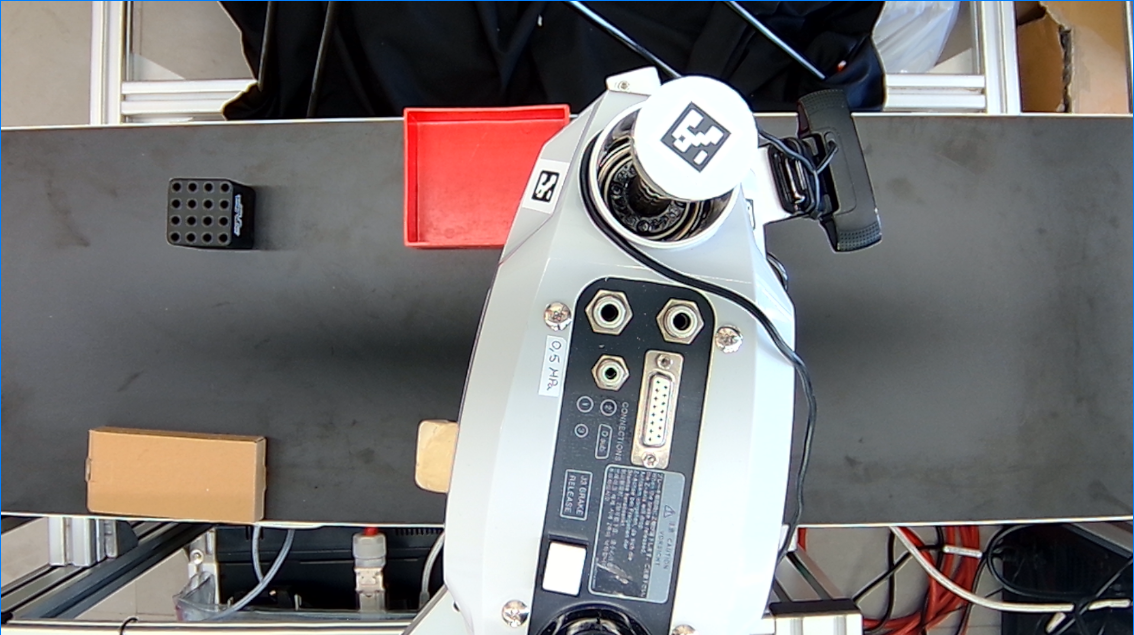}
        \caption{Put all the blocks on the corner (initial state)}
    \end{subfigure}
    \hfill
    \begin{subfigure}{0.32\textwidth}
        \includegraphics[width=\linewidth]{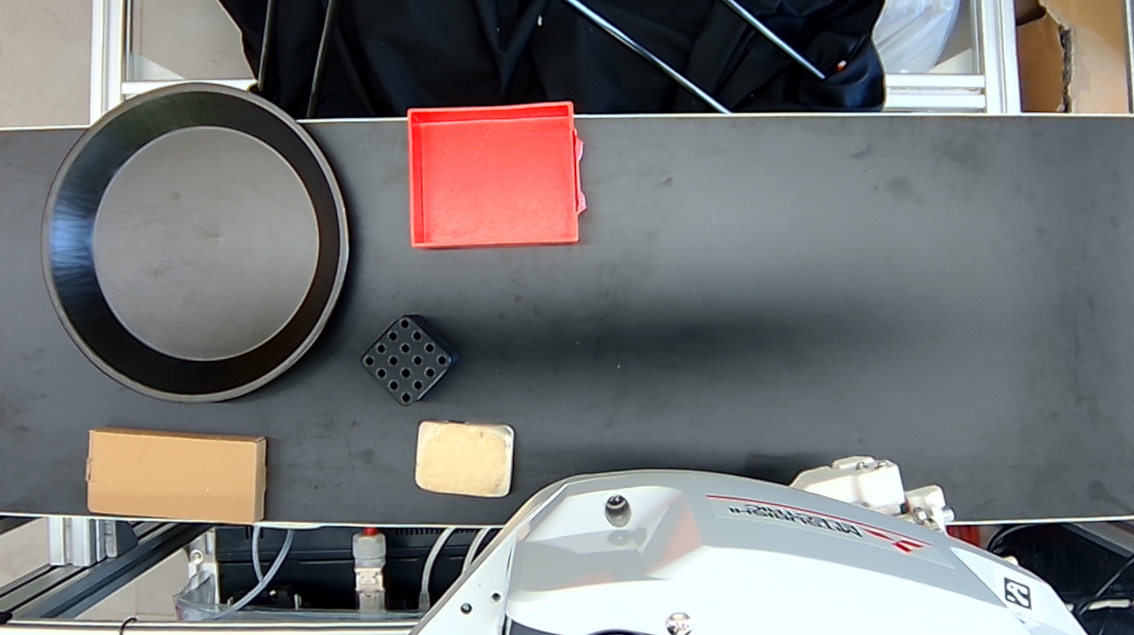}
        \caption{Put the blocks in the receptacle (initial state)}
    \end{subfigure}
    
    \vspace{1em} 
    \begin{subfigure}{0.32\textwidth}
        \includegraphics[width=\linewidth]{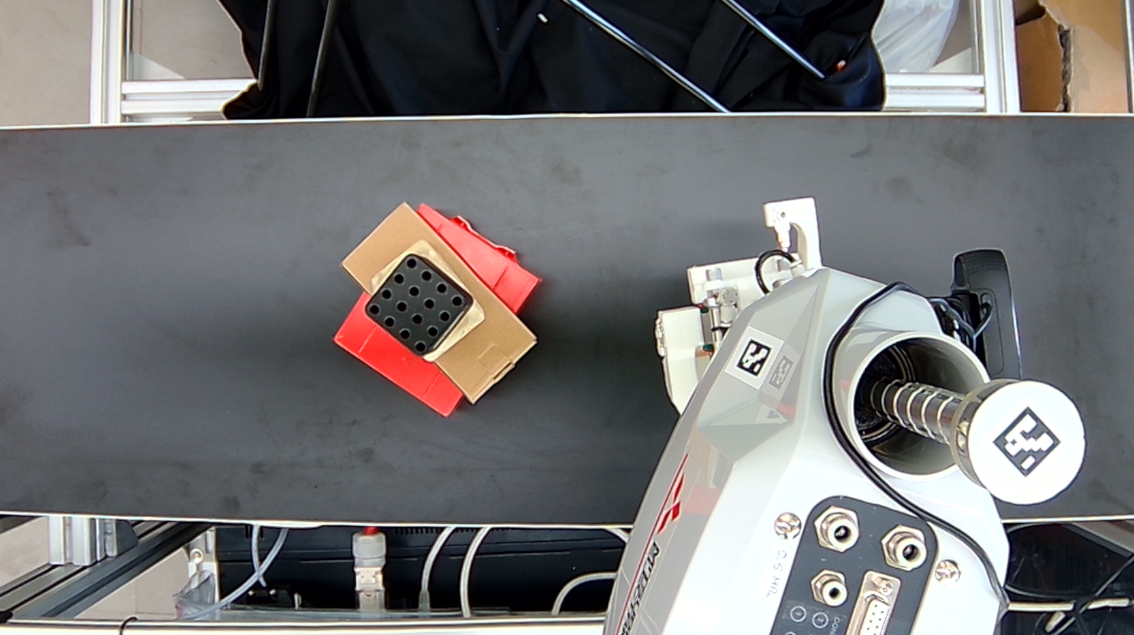}
        \caption{Stack all the blocks (final state)}
    \end{subfigure}
    \hfill
    \begin{subfigure}{0.32\textwidth}
        \includegraphics[width=\linewidth]{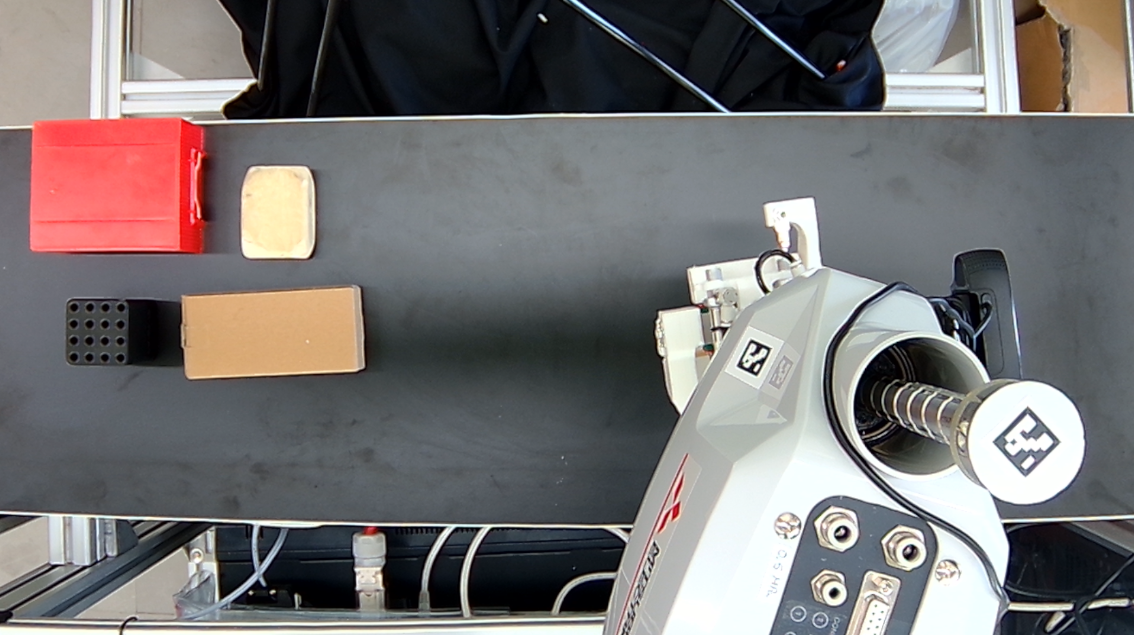}
        \caption{Put all the blocks on the corner (final state)}
    \end{subfigure}
    \hfill
    \begin{subfigure}{0.32\textwidth}
        \includegraphics[width=\linewidth]{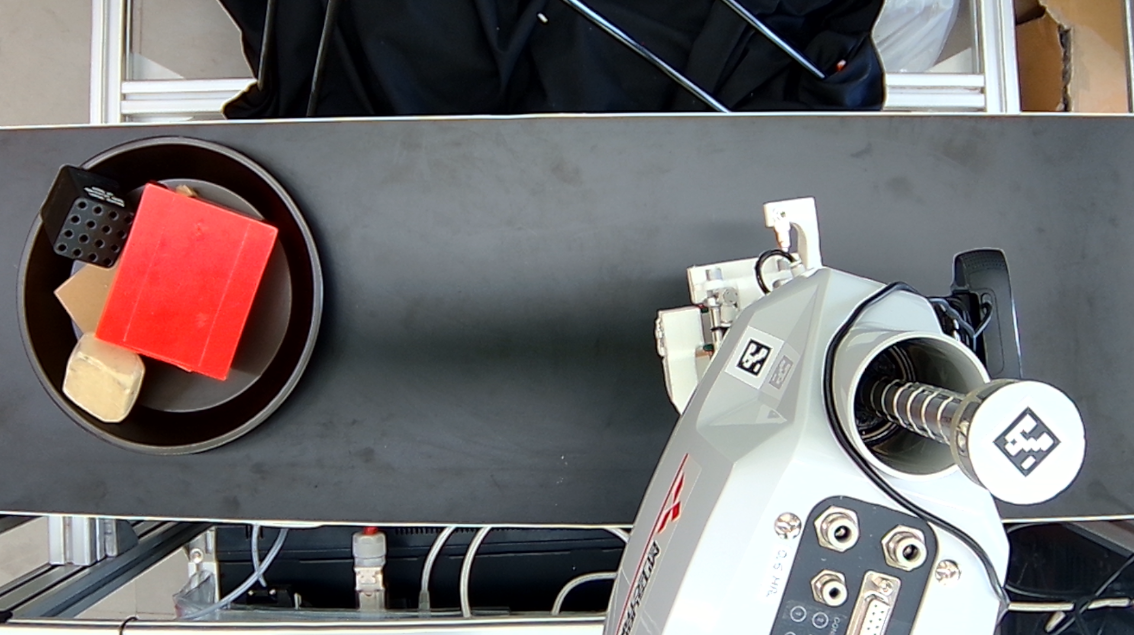}
        \caption{Put the blocks in the receptacle (final state)}
    \end{subfigure}

    \caption{Examples of static experiments on SCARA Robot - I}
    \label{fig:st-exp-delta-1}
\end{figure}

\begin{figure}[h]
    \centering
    \begin{subfigure}{0.32\textwidth}
        \includegraphics[width=\linewidth]{img/4-1.png}
        \caption{Put all the blocks in the bowls with
matching colors (initial state)}
    \end{subfigure}
    \hfill 
    \begin{subfigure}{0.32\textwidth}
        \includegraphics[width=\linewidth]{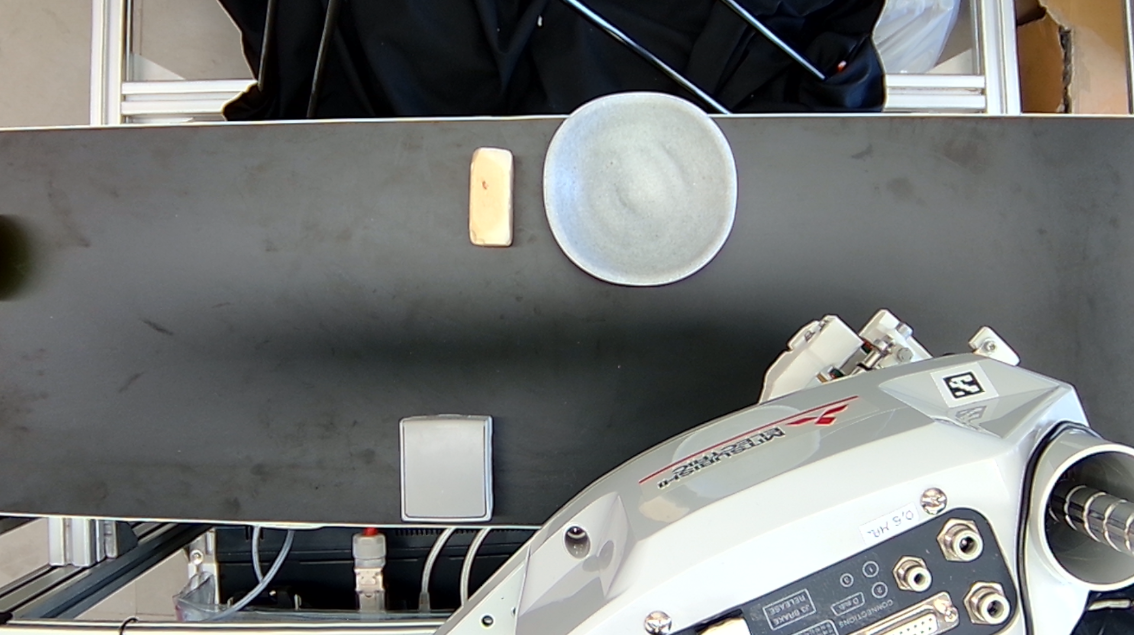}
        \caption{Pick up the block to the direction of the
receptacle and place it on the corner (initial state)}
    \end{subfigure}
    \hfill
    \begin{subfigure}{0.32\textwidth}
        \includegraphics[width=\linewidth]{img/6-1.png}
        \caption{Pick up the $3^{rd}$ block from the direction
and place it on the corner (initial state)}
    \end{subfigure}

    \vspace{1em} 
    \begin{subfigure}{0.32\textwidth}
        \includegraphics[width=\linewidth]{img/4-2.png}
        \caption{Put all the blocks in the bowls with
matching colors (final state)}
    \end{subfigure}
    \hfill
    \begin{subfigure}{0.32\textwidth}
        \includegraphics[width=\linewidth]{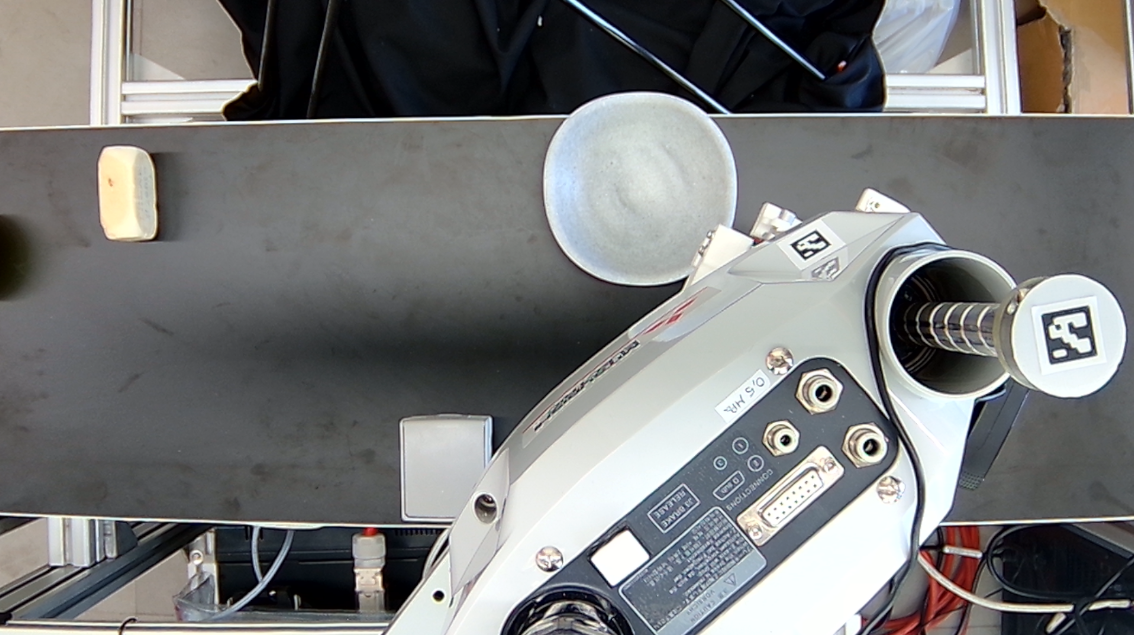}
        \caption{Pick up the block to the direction of the
receptacle and place it on the corner (final state)}
    \end{subfigure}
    \hfill
    \begin{subfigure}{0.32\textwidth}
        \includegraphics[width=\linewidth]{img/6-2.png}
        \caption{Pick up the $3^{rd}$ block from the direction
and place it on the corner (final state)}
    \end{subfigure}

    \caption{Examples of static experiments on SCARA Robot - II}
    \label{fig:st-exp-delta-2}
\end{figure}

\begin{figure}
    \centering
    \begin{subfigure}{0.32\textwidth}
        \includegraphics[width=\linewidth]{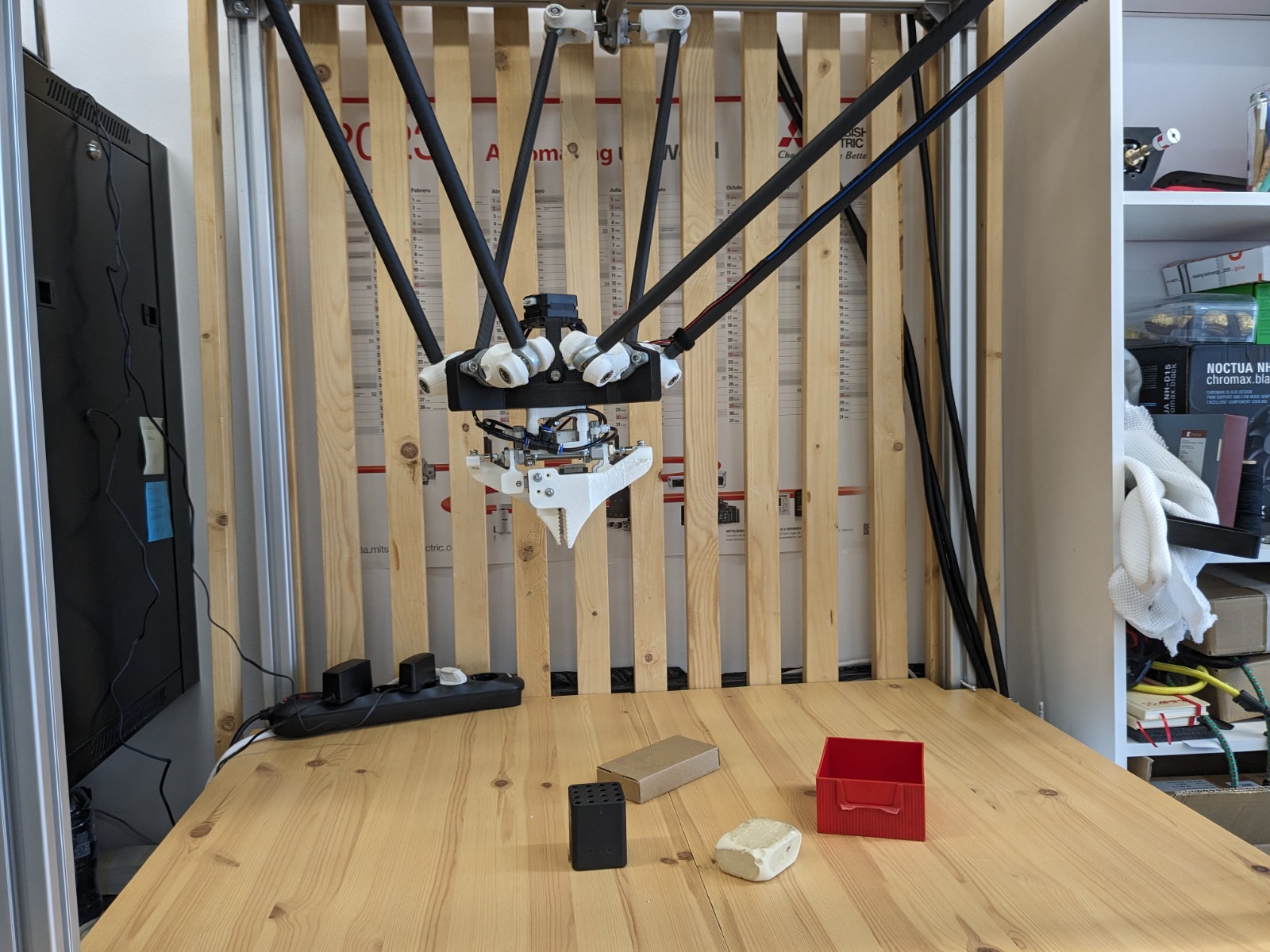}
        \caption{Stack all the blocks one on top of the other without falling off(initial state)}
    \end{subfigure}
    \hfill 
    \begin{subfigure}{0.32\textwidth}
        \includegraphics[width=\linewidth]{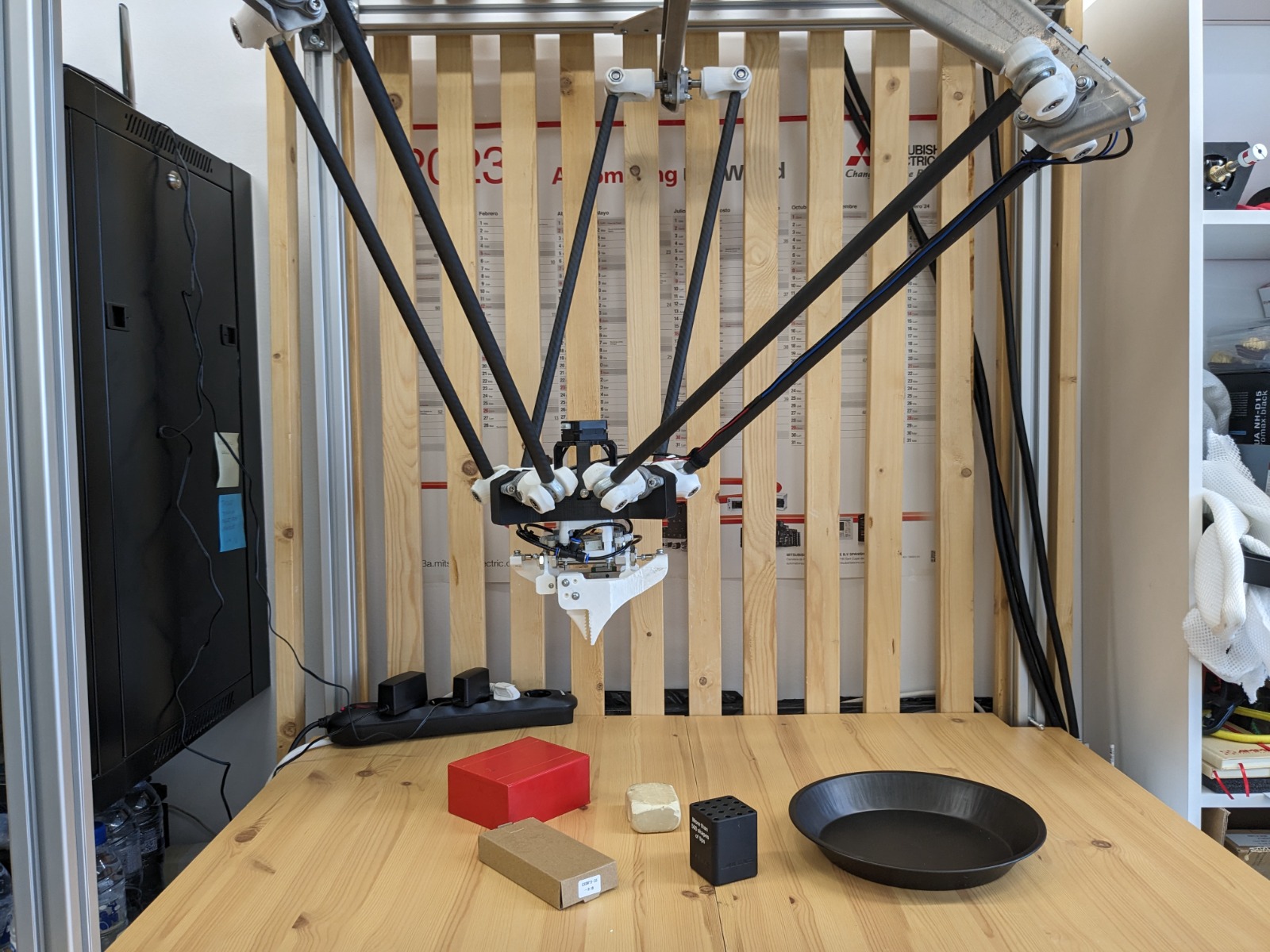}
        \caption{Put all the blocks in the receptacle, no one should be left out (initial state)}
    \end{subfigure}
    \hfill
    \begin{subfigure}{0.32\textwidth}
        \includegraphics[width=\linewidth]{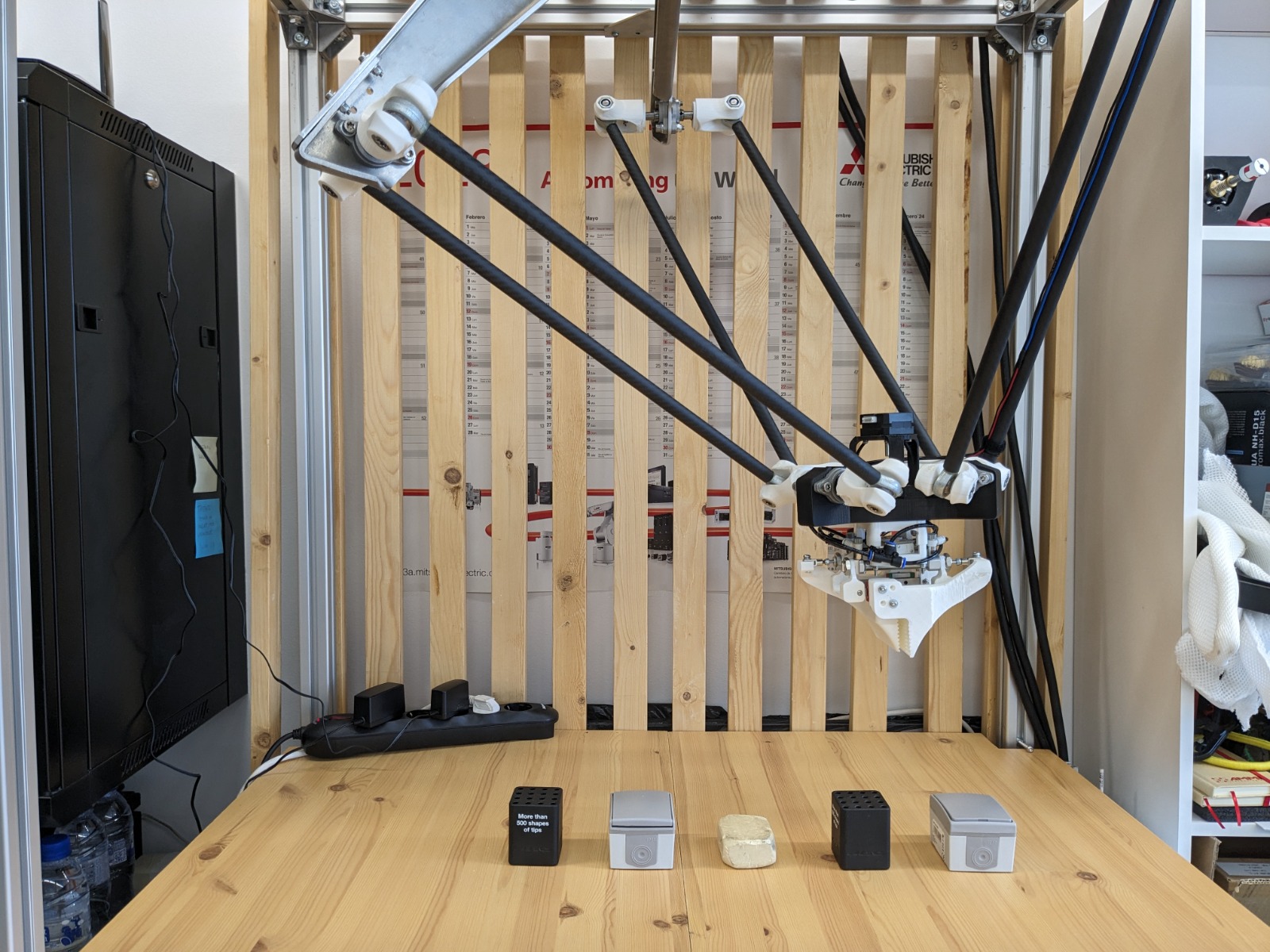}
        \caption{Pick up the $3^{rd}$ block from the direction and place it on the corner (initial state)}
    \end{subfigure}
    
    \vspace{1em} 
    \begin{subfigure}{0.32\textwidth}
        \includegraphics[width=\linewidth]{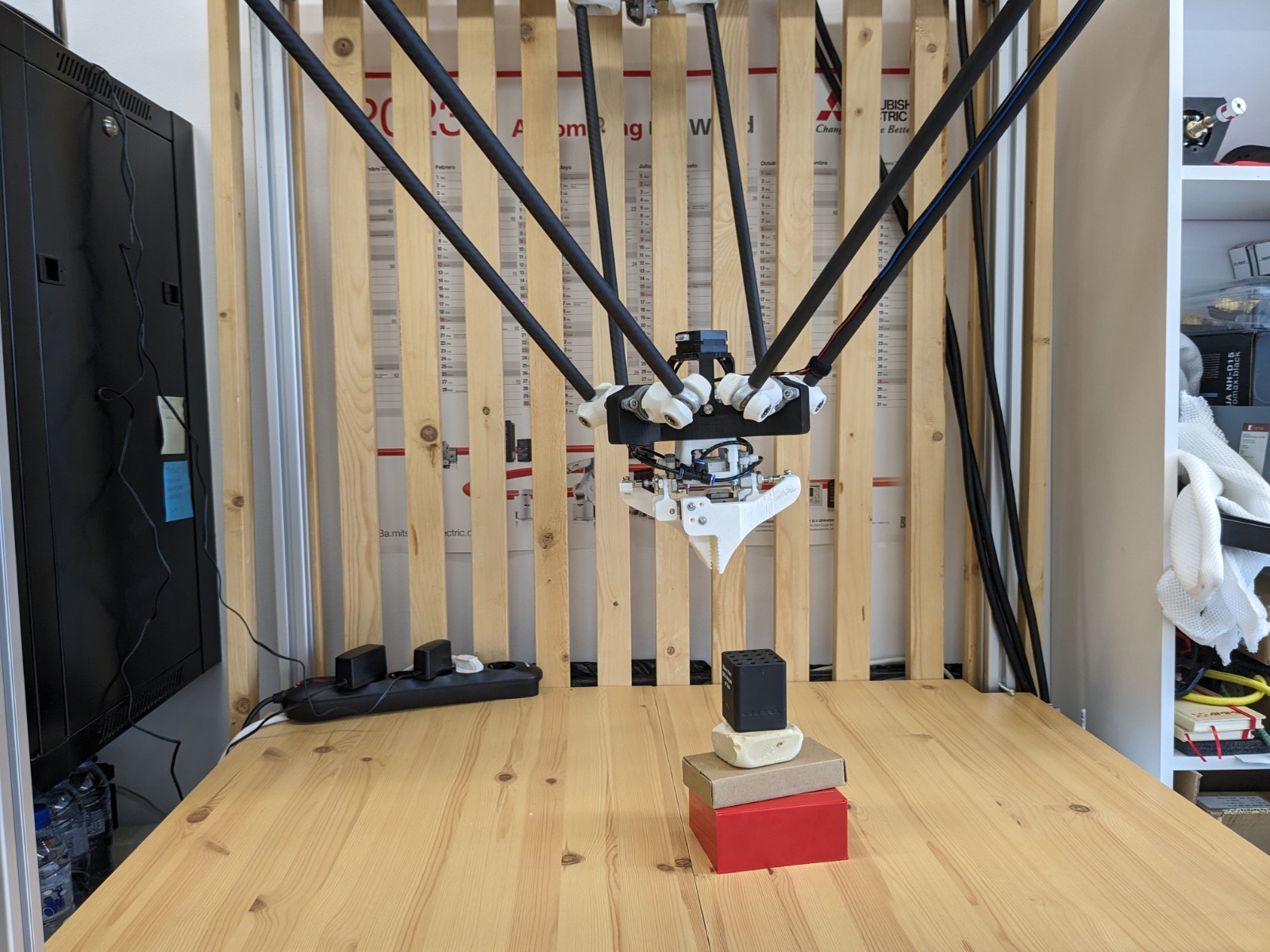}
        \caption{Stack all the blocks one on top of the other without falling off (final state)}
    \end{subfigure}
    \hfill
    \begin{subfigure}{0.32\textwidth}
        \includegraphics[width=\linewidth]{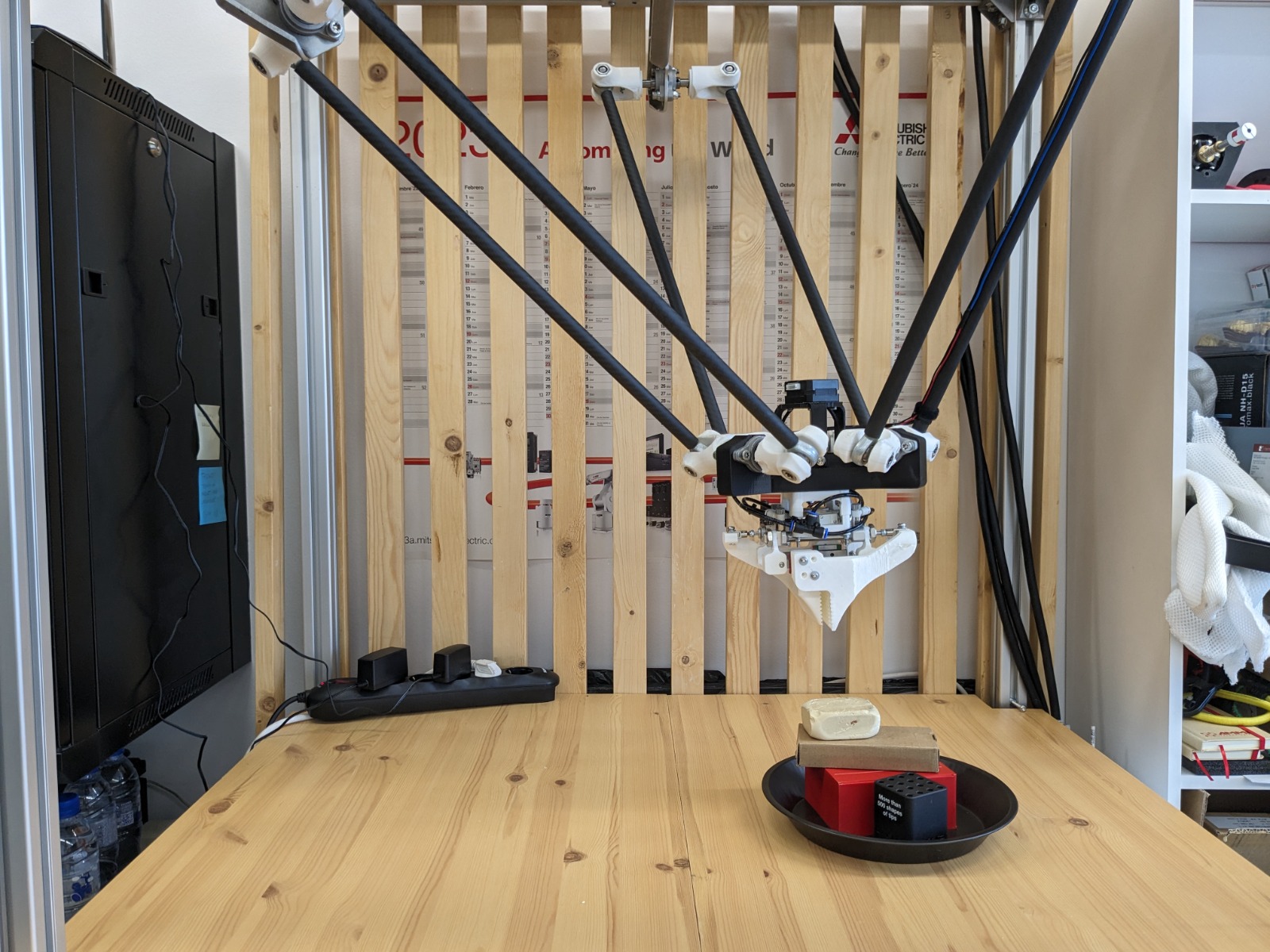}
        \caption{Put all the blocks in the receptacle, no one should be left out (final state)}
    \end{subfigure}
    \hfill
    \begin{subfigure}{0.32\textwidth}
        \includegraphics[width=\linewidth]{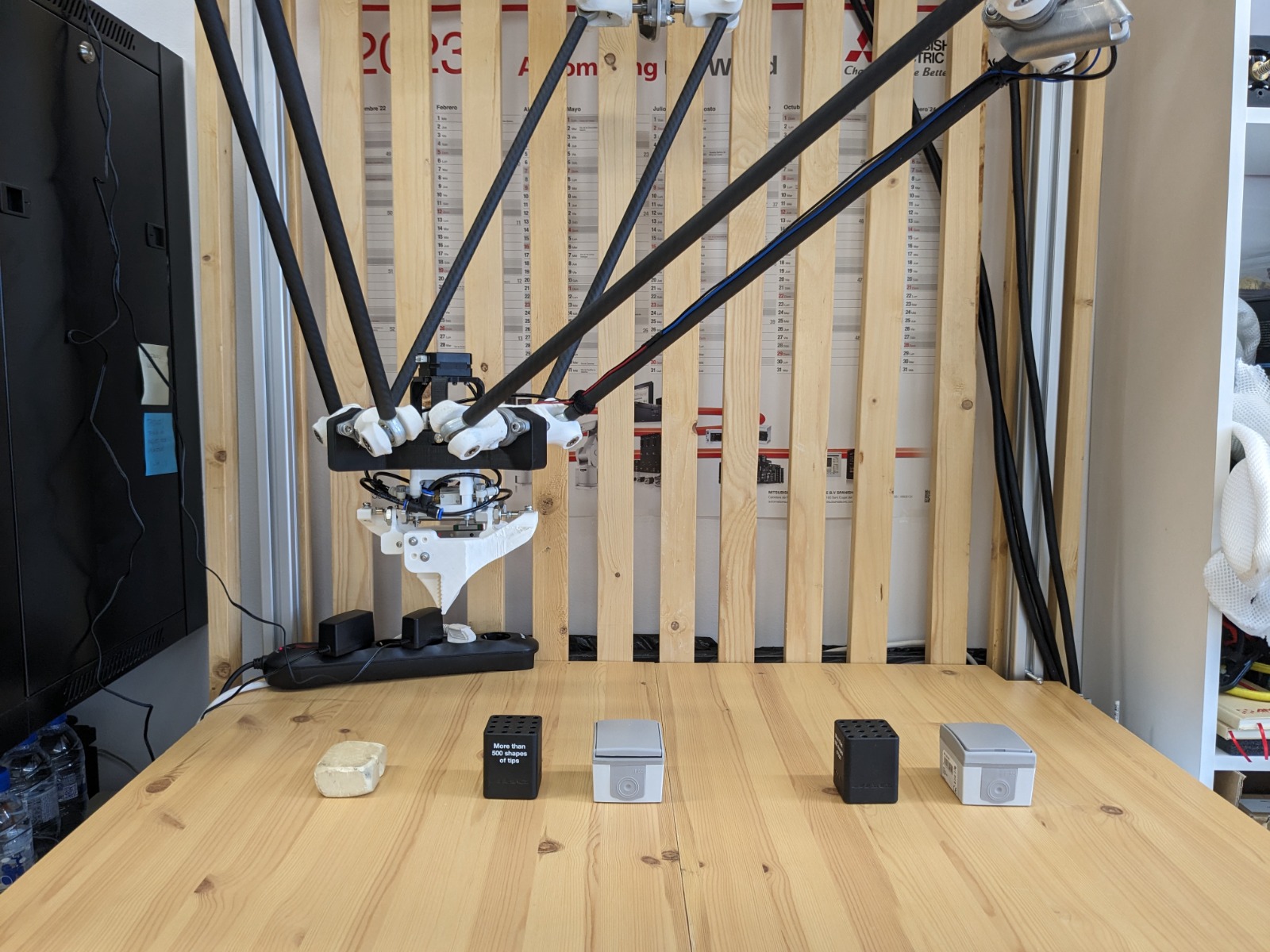}
        \caption{Pick up the $3^{rd}$ block from the direction and place it on the corner (final state)}
    \end{subfigure}

    \caption{Examples of static experiments on DELTA Robot}
    \label{fig:st-exp-delta-e}
\end{figure}

\newpage
\subsection{Dynamic}
\label{ap-sec:dynamic-experiments-fig}
The videos have been uploaded separately, directly on the OpenReview platform.

    \begin{figure}
    \centering
    
    \begin{subfigure}{0.32\textwidth}
        \includegraphics[width=1\textwidth, height=0.65\textwidth]{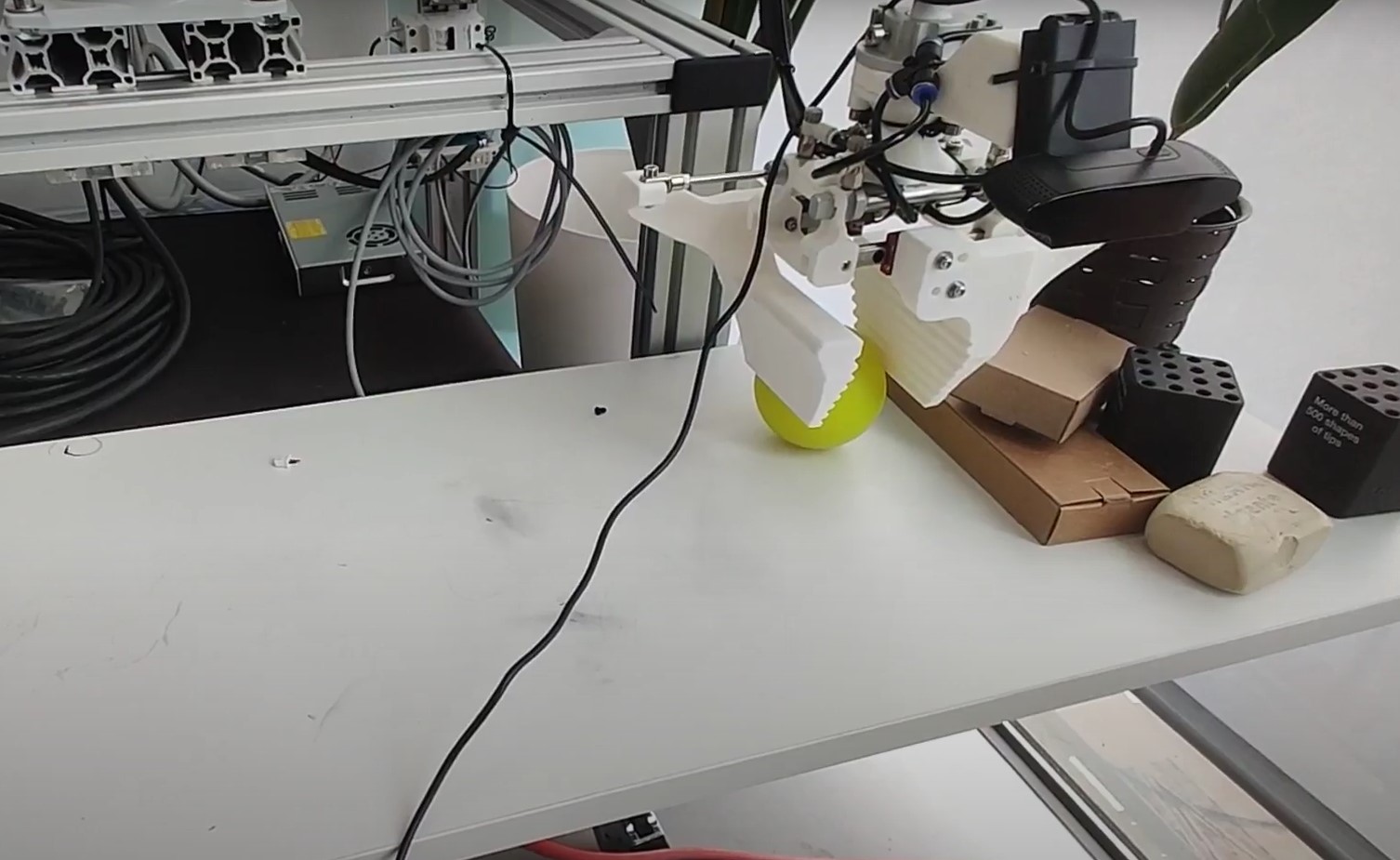}
        \caption{"Take the ball"}
    \end{subfigure}
    \hfill 
    \begin{subfigure}{0.32\textwidth}
        \includegraphics[width=1\textwidth, height=0.65\textwidth]{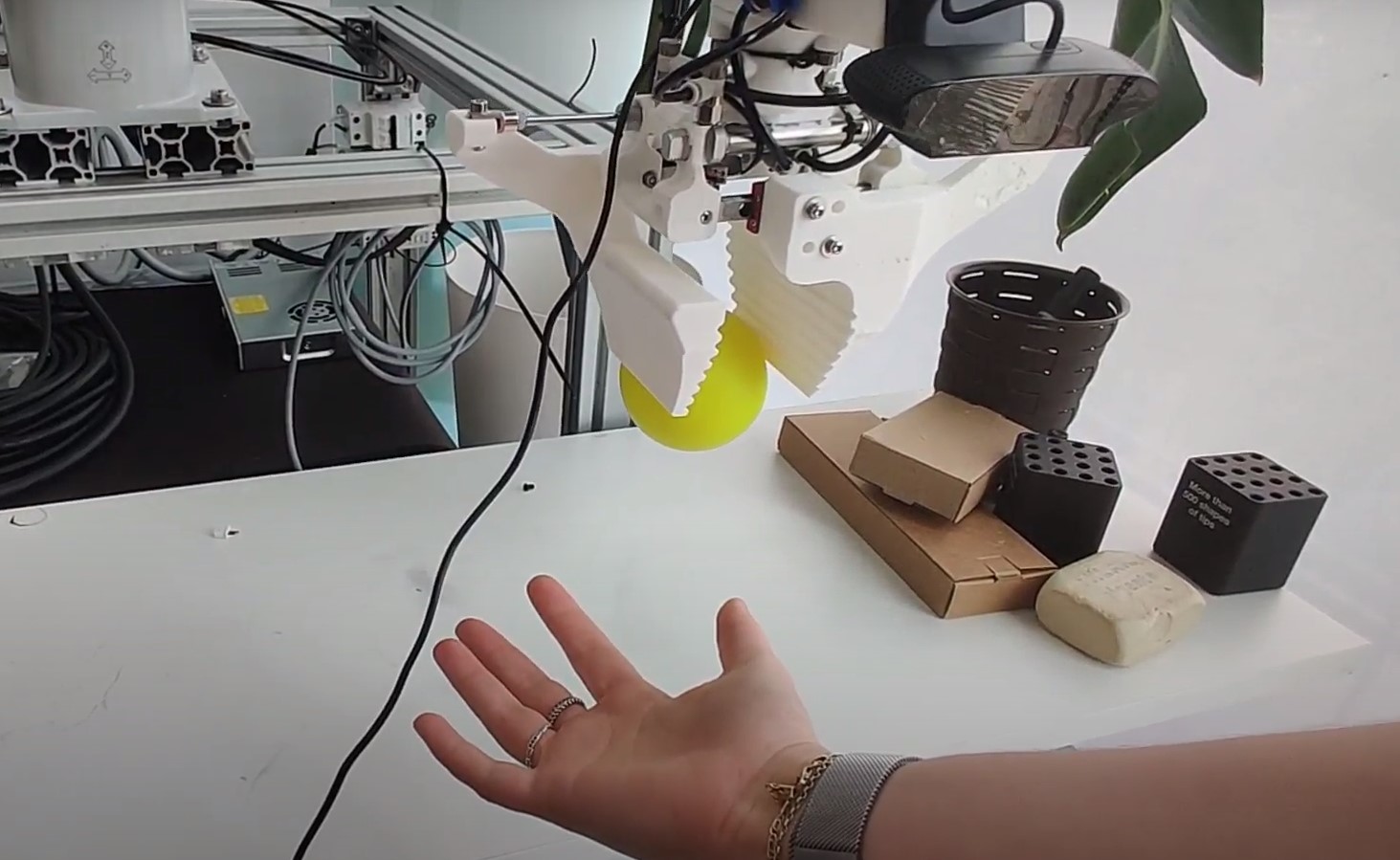}
        \caption{"Follow my hand"}
    \end{subfigure}
    \hfill
    \begin{subfigure}{0.32\textwidth}
        \includegraphics[width=1\textwidth, height=0.65\textwidth]{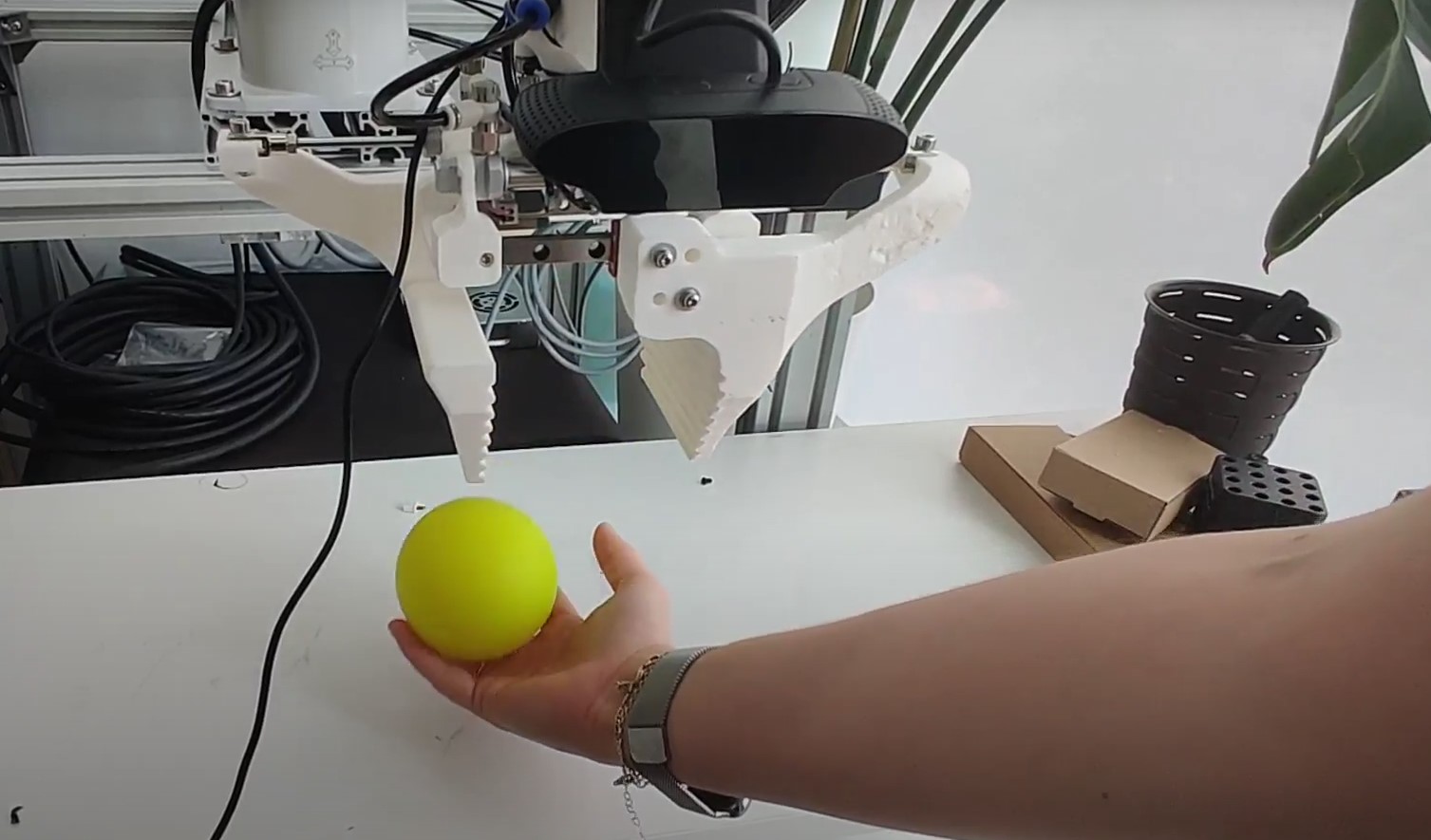}
        \caption{"Give it to me"}
    \end{subfigure}
    \caption{Task: "Go ahead, take the ball and then, give it back to my hand. The hand is moving and the robot must follow it" - Example of dynamic experiment on SCARA Robot}
    \label{fig:dyn-exp1-scara}
\end{figure}

\begin{figure}
    \centering
    \begin{subfigure}{0.32\textwidth}
        \includegraphics[width=1\textwidth, height=0.75\textwidth]{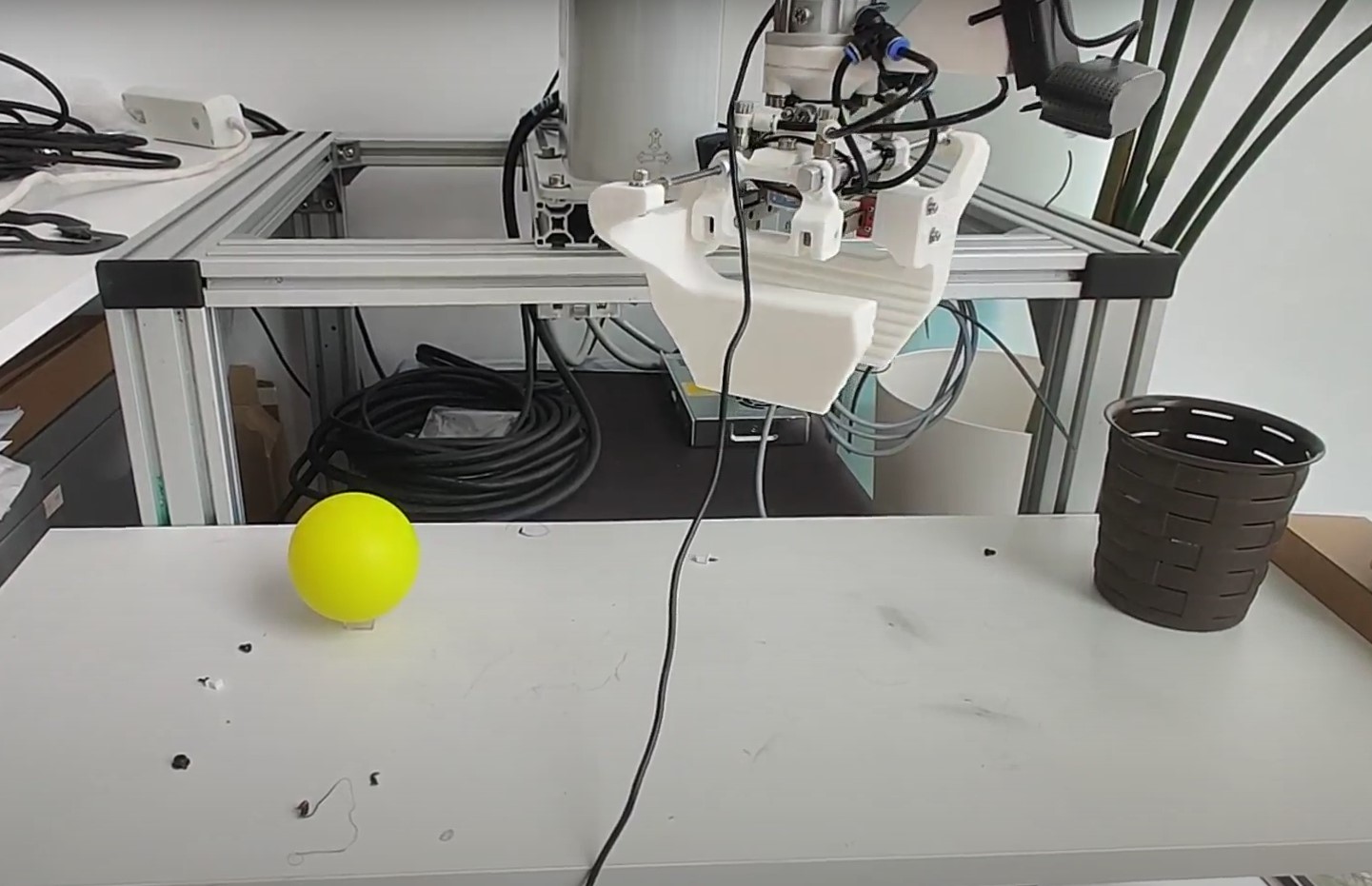}
        \caption{"Look for the ball in all the space"}
    \end{subfigure}
    \hfill 
    \begin{subfigure}{0.32\textwidth}
        \includegraphics[width=1\textwidth, height=0.75\textwidth]{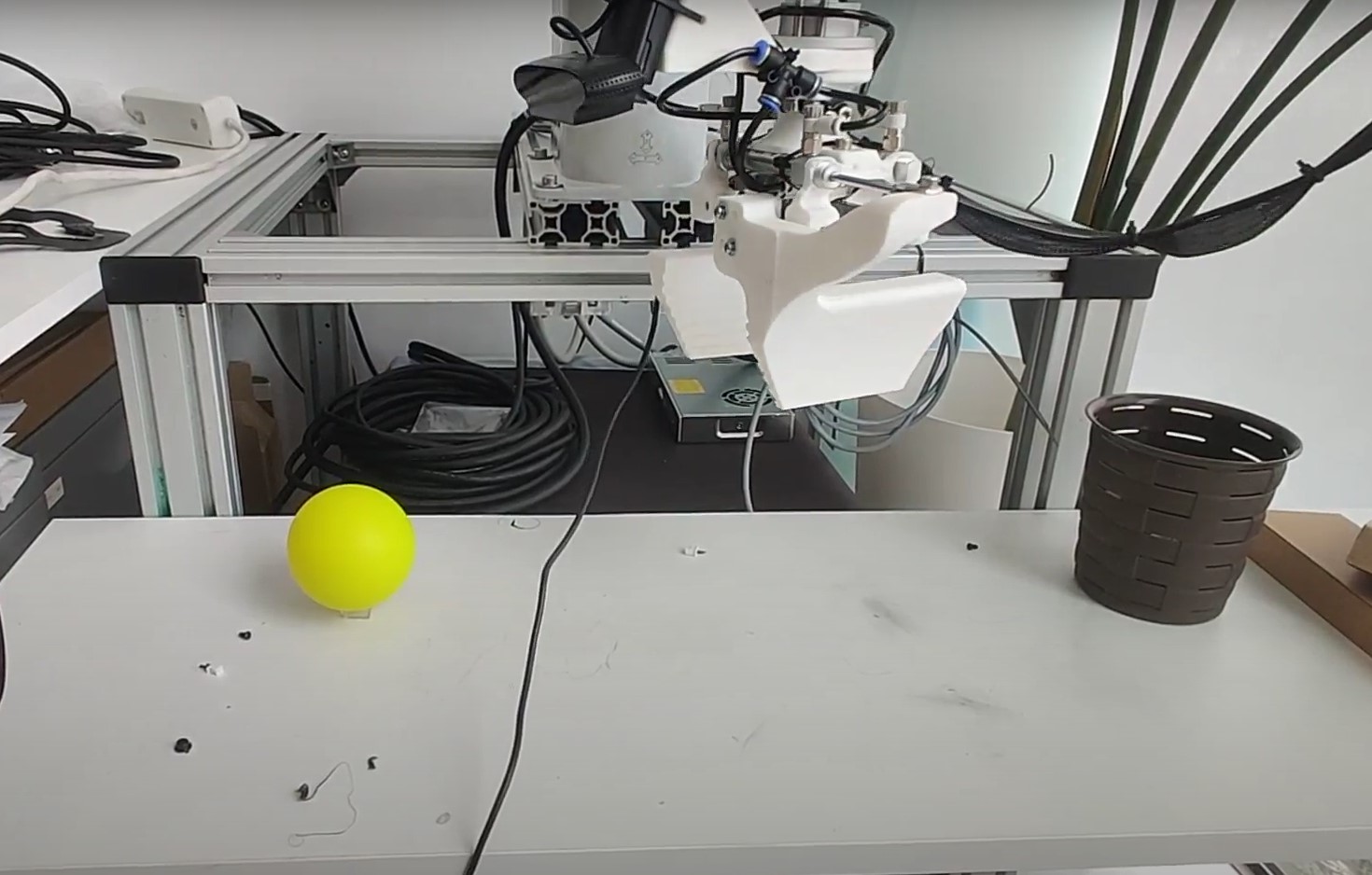}
        \caption{"Look for the ball in all the space"}
    \end{subfigure}
    \hfill
    \begin{subfigure}{0.32\textwidth}
        \includegraphics[width=1\textwidth, height=0.75\textwidth]{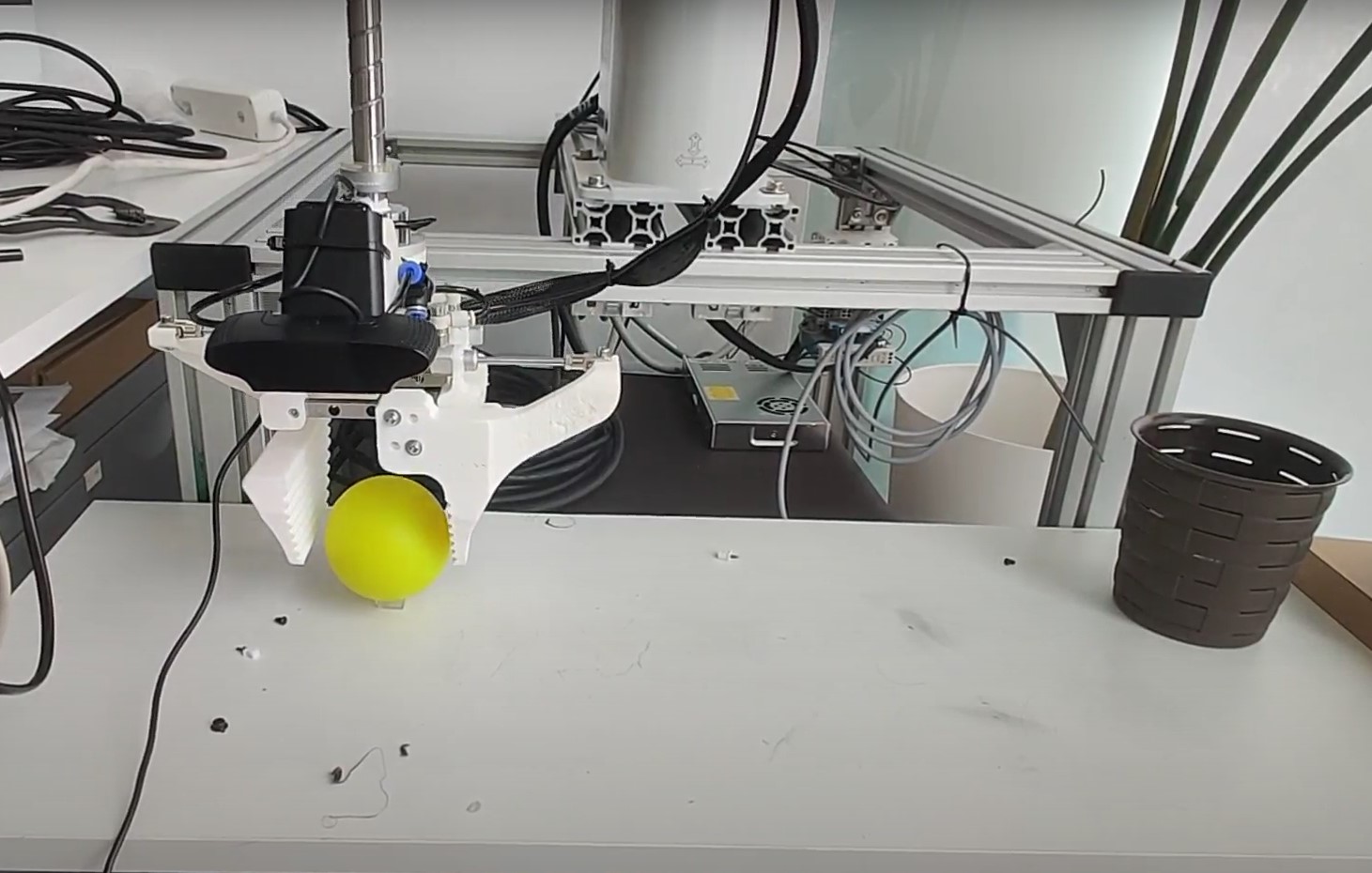}
        \caption{"Take the ball"}
    \end{subfigure}

    \vspace{1em} 
    \begin{subfigure}{0.32\textwidth}
        \includegraphics[width=1\textwidth, height=0.75\textwidth]{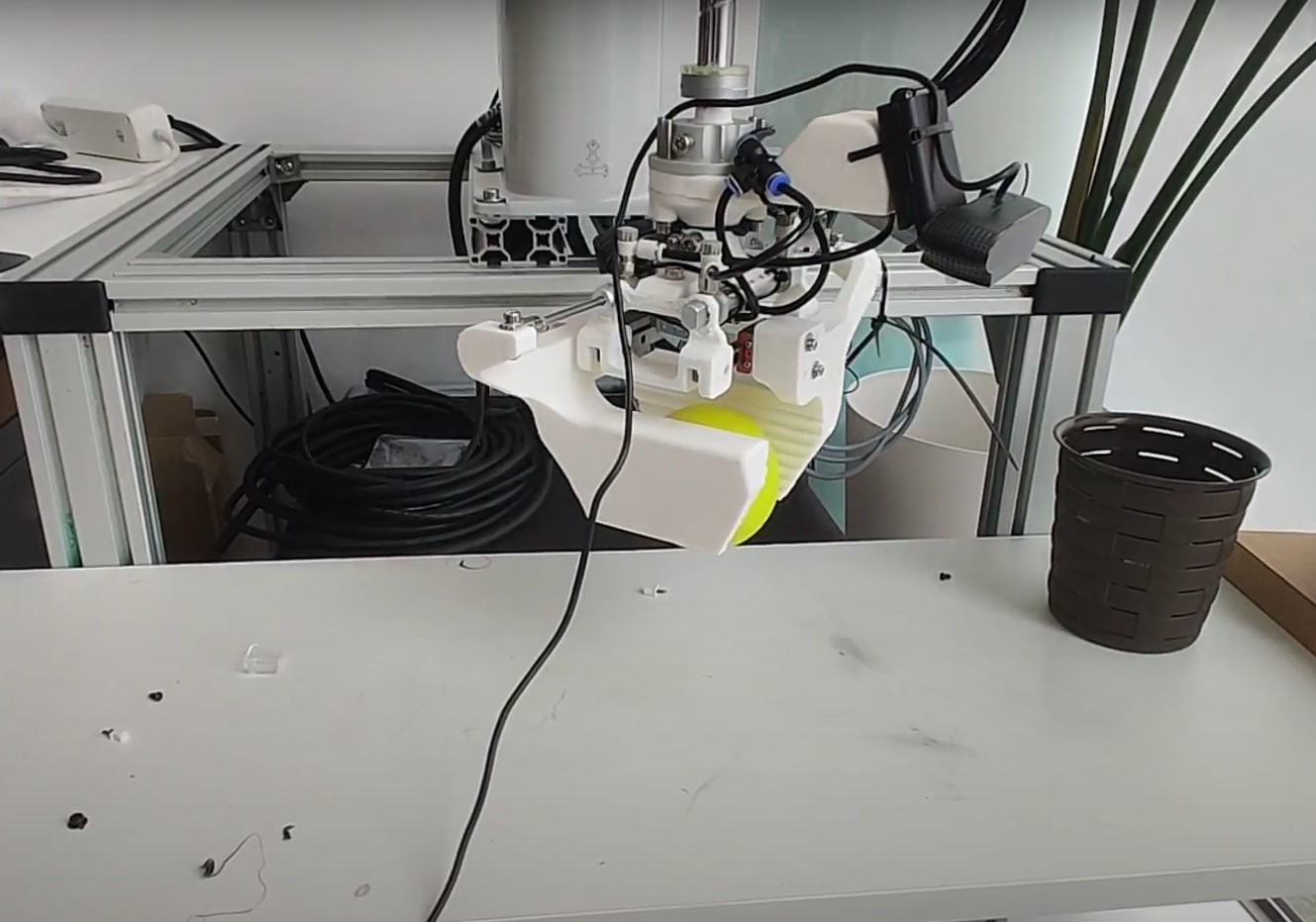}
        \caption{"Look for the bucket in all the space"}
    \end{subfigure}
    \hfill
    \begin{subfigure}{0.32\textwidth}
        \includegraphics[width=1\textwidth, height=0.75\textwidth]{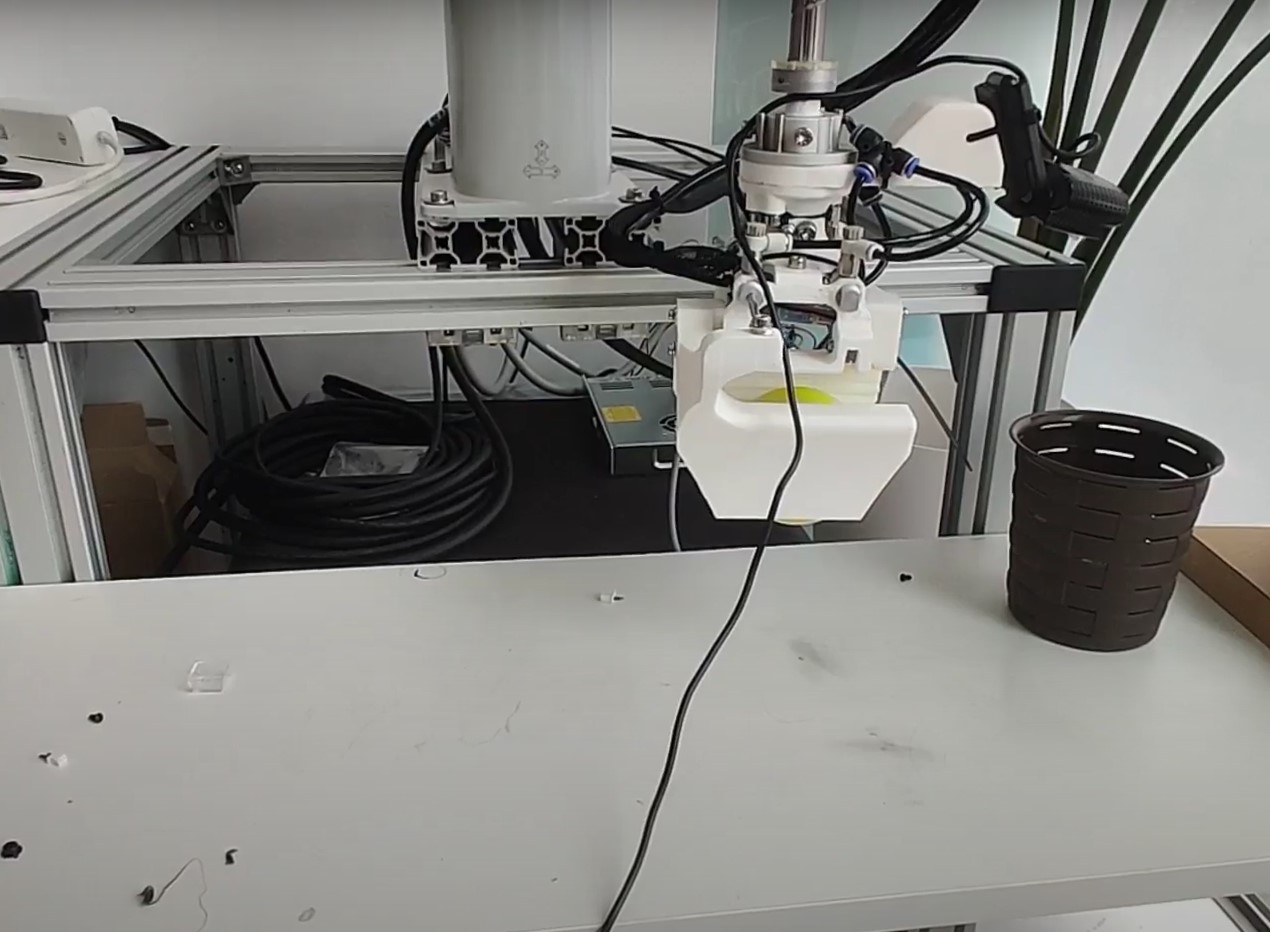}
        \caption{"Look for the bucket in all the space"}
    \end{subfigure}
    \hfill
    \begin{subfigure}{0.32\textwidth}
        \includegraphics[width=1\textwidth, height=0.75\textwidth]{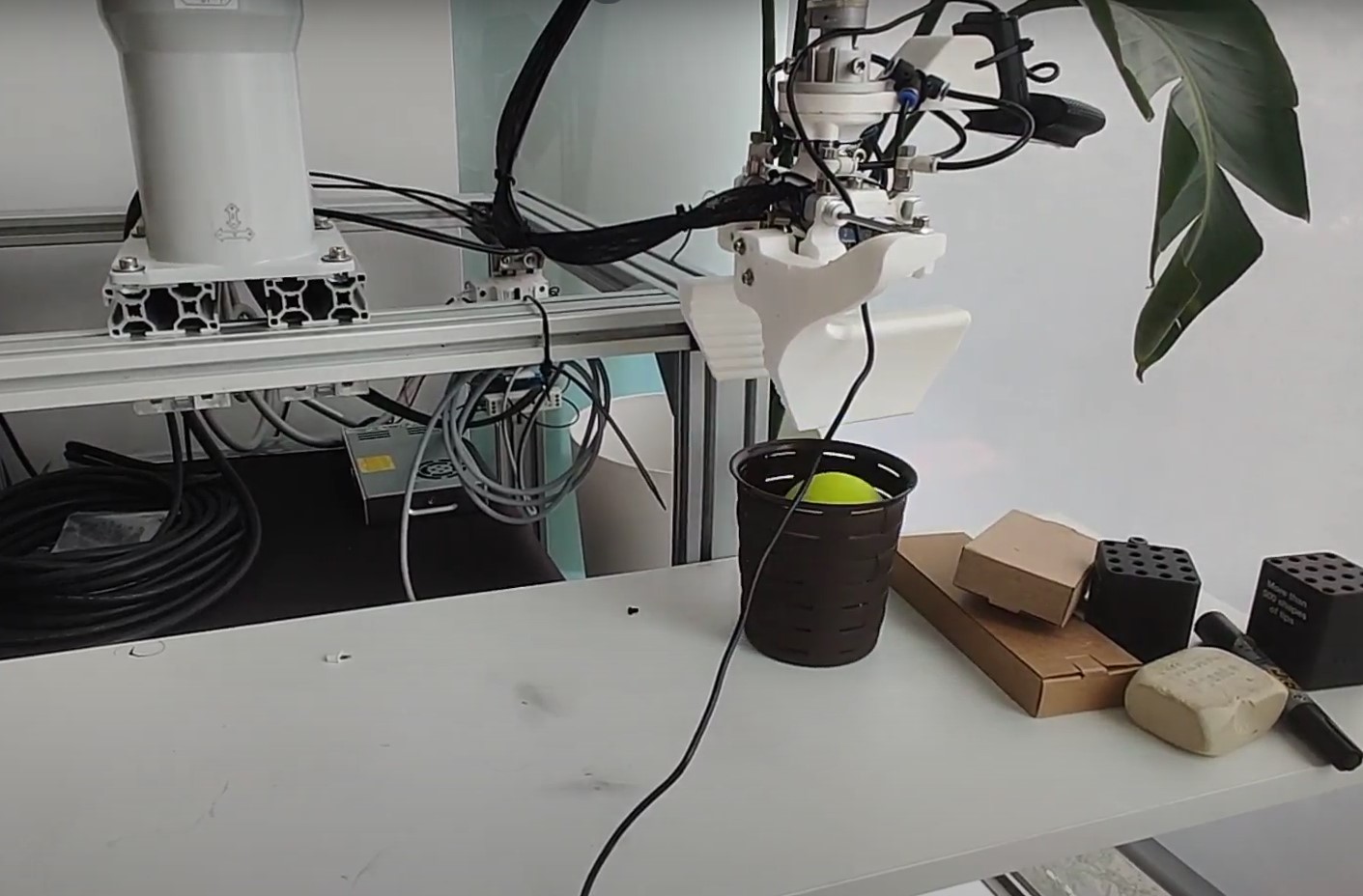}
        \caption{"Put the ball in the bucket"}
    \end{subfigure}
    
    \caption{Task: "Find and pick-up the round object and put it into the bucket" - Example of dynamic experiment on SCARA Robot}
    \label{fig:dyn-exp2-scara}
\end{figure}


\begin{figure}
    \centering
    \begin{subfigure}{0.32\textwidth}
        \includegraphics[width=\linewidth]{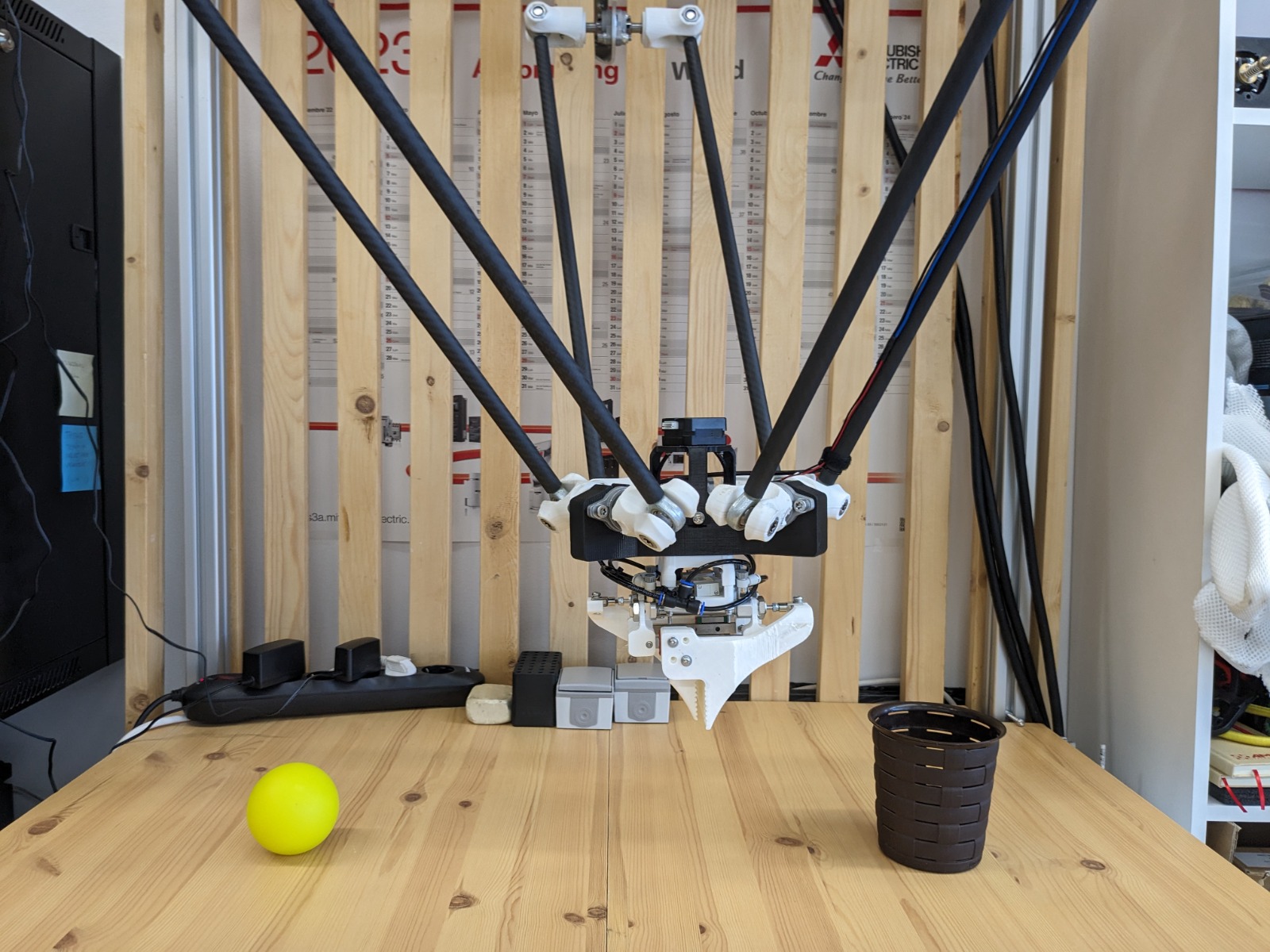}
        \caption{"Look for the ball in all the space"}
    \end{subfigure}
    \hfill 
    \begin{subfigure}{0.32\textwidth}
        \includegraphics[width=\linewidth]{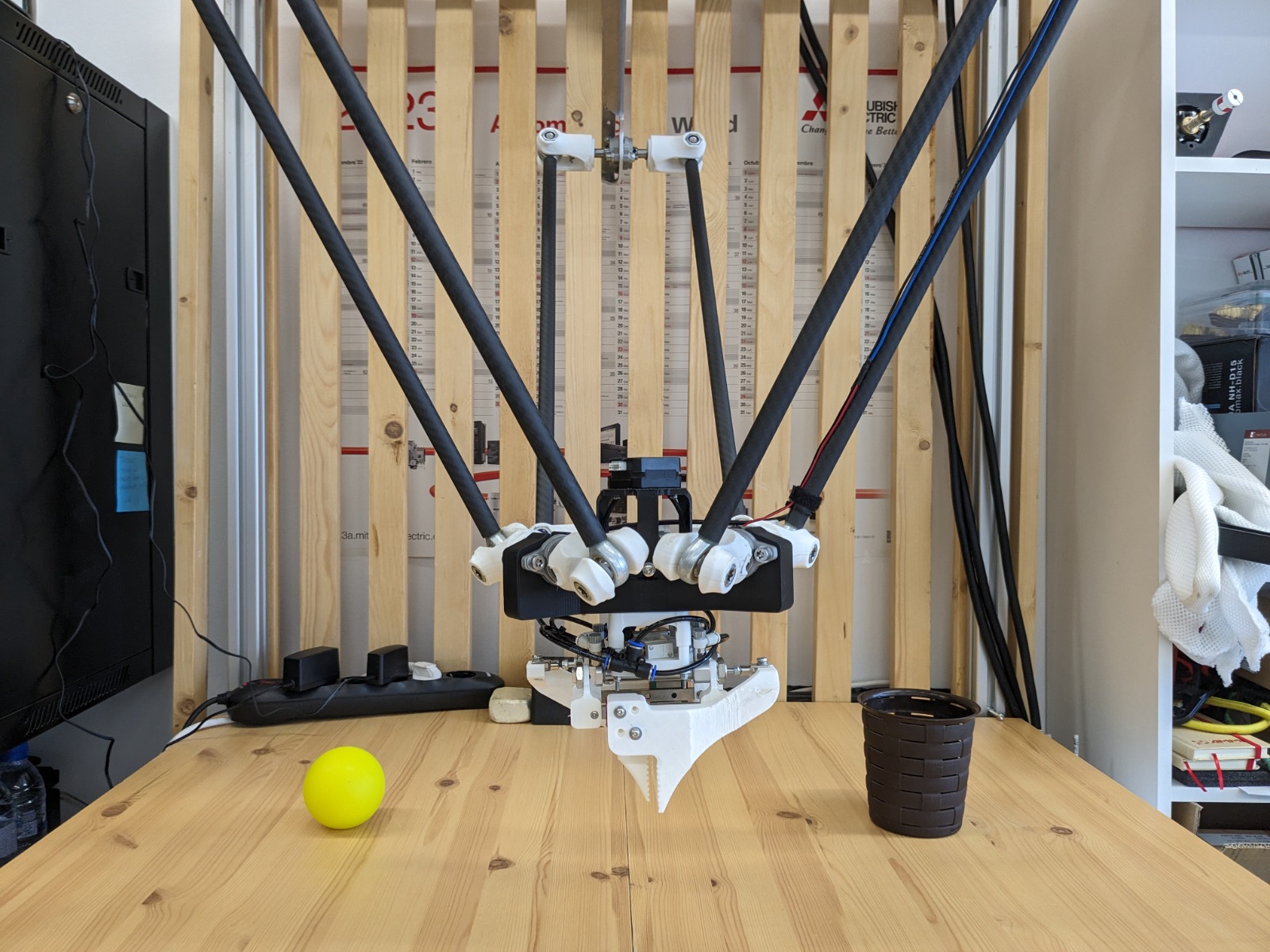}
        \caption{"Look for the ball in all the space"}
    \end{subfigure}
    \hfill
    \begin{subfigure}{0.32\textwidth}
        \includegraphics[width=\linewidth]{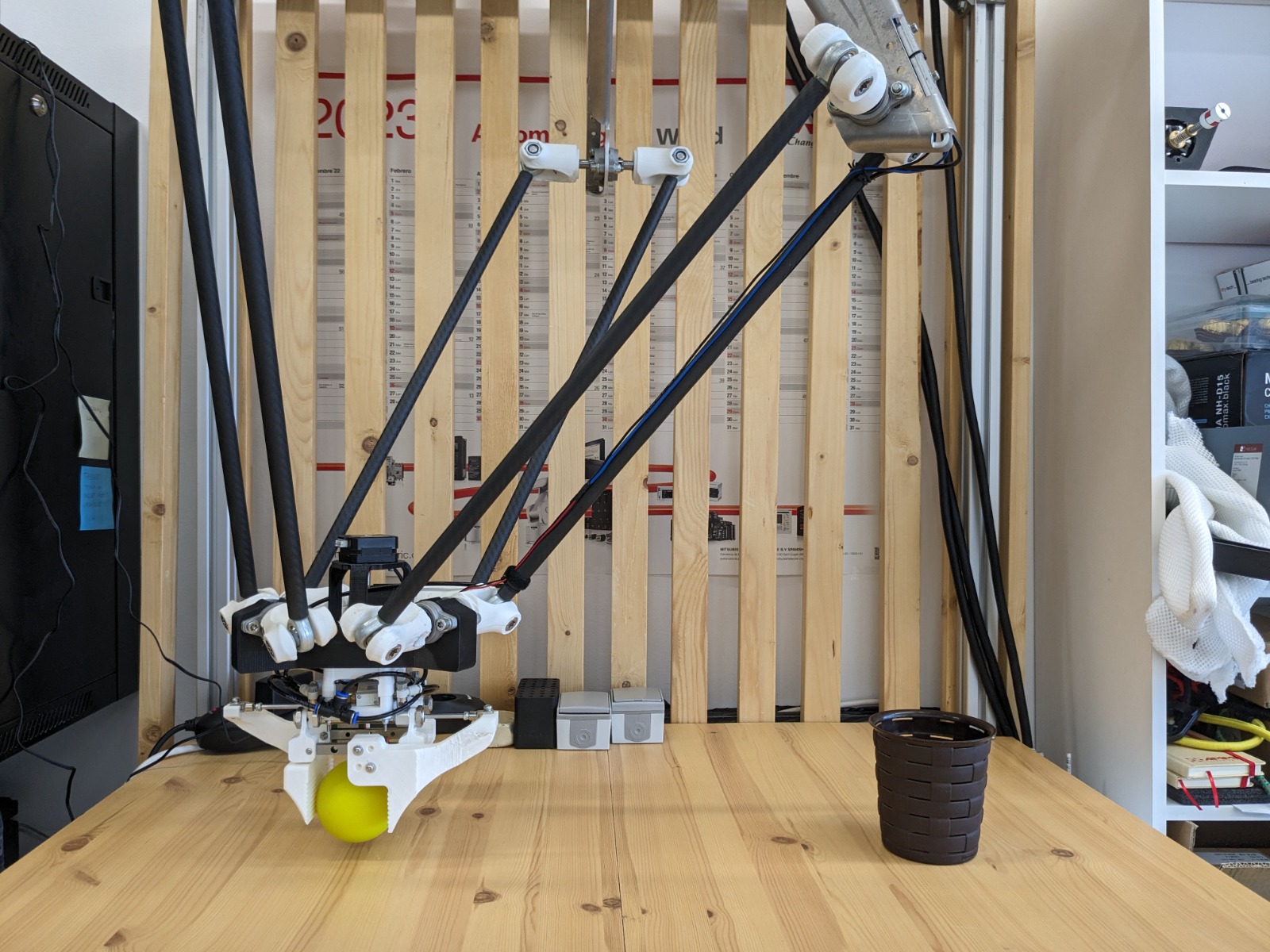}
        \caption{"Take the ball"}
    \end{subfigure}

    \vspace{1em} 
    \begin{subfigure}{0.32\textwidth}
        \includegraphics[width=\linewidth]{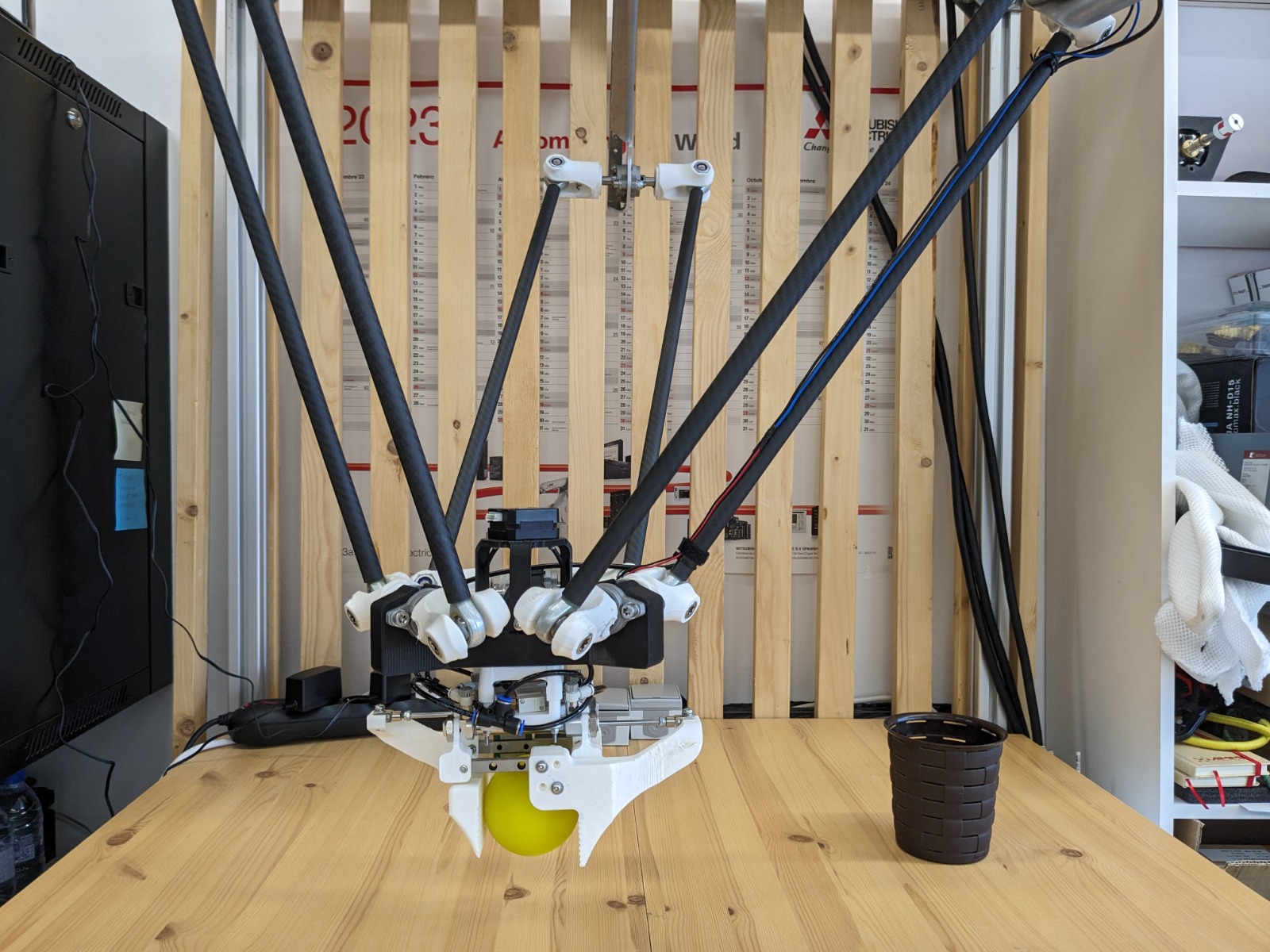}
        \caption{"Look for the bucket in all the space"}
    \end{subfigure}
    \hfill
    \begin{subfigure}{0.32\textwidth}
        \includegraphics[width=\linewidth]{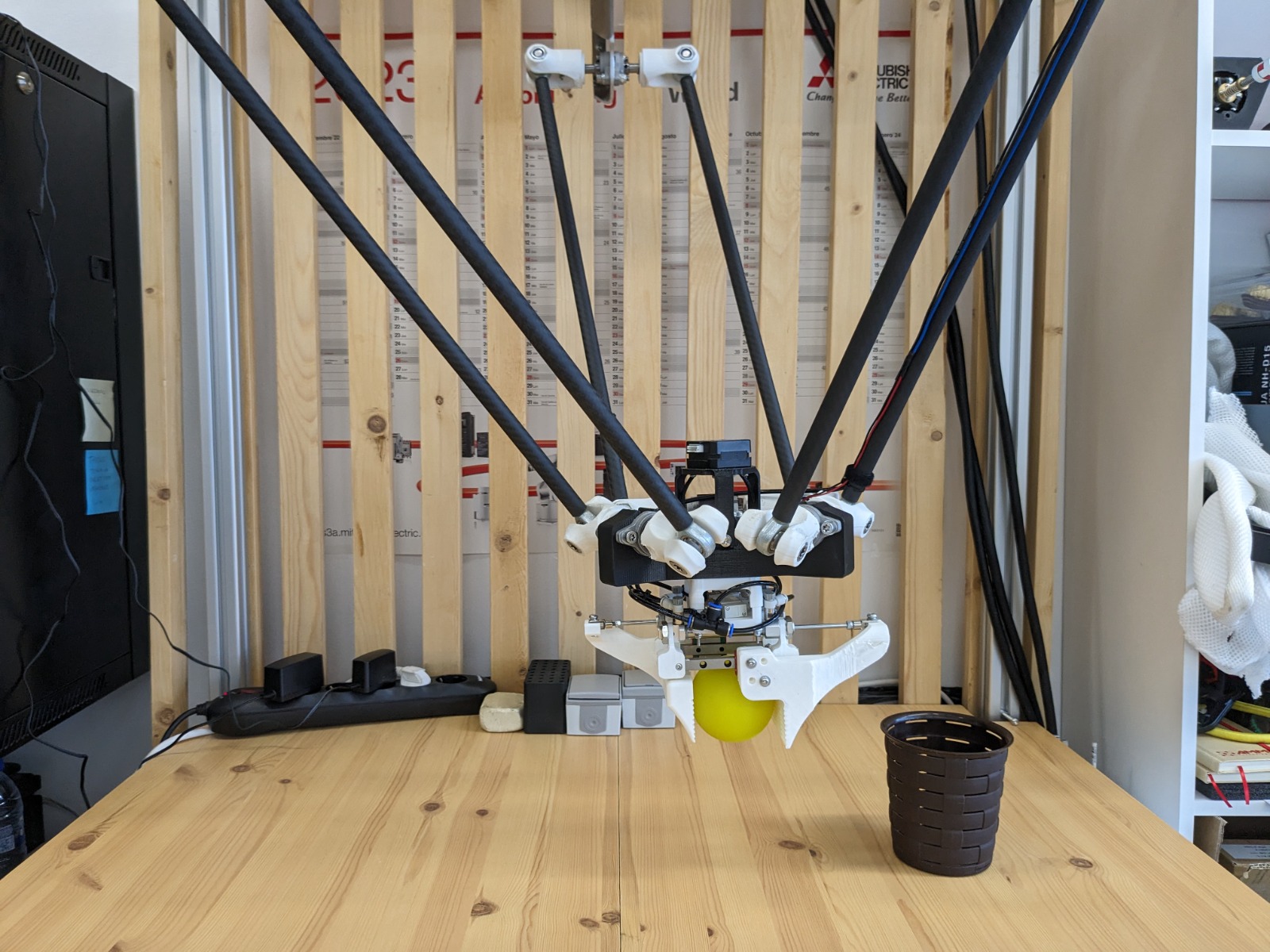}
        \caption{"Look for the bucket in all the space"}
    \end{subfigure}
    \hfill
    \begin{subfigure}{0.32\textwidth}
        \includegraphics[width=\linewidth]{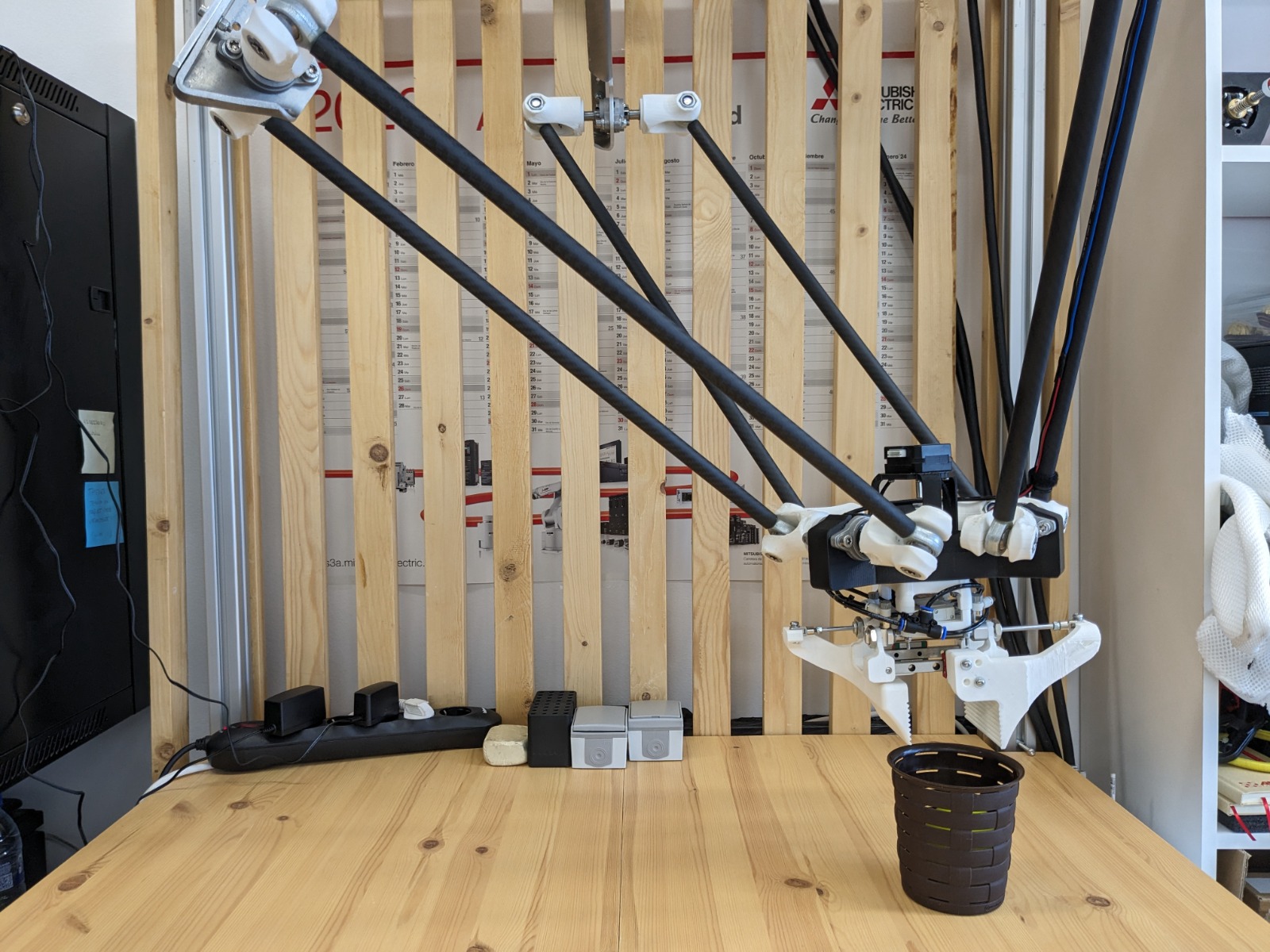}
        \caption{"Put the ball in the bucket"}
    \end{subfigure}
    
    \caption{Task: "Find and pick-up the round object and put it into the bucket" - Example of dynamic experiment on DELTA Robot}
    \label{fig:dyn-exp2-delta}
\end{figure}

    \begin{figure}
    \centering
    
    \begin{subfigure}{0.32\textwidth}
        \includegraphics[width=\linewidth]{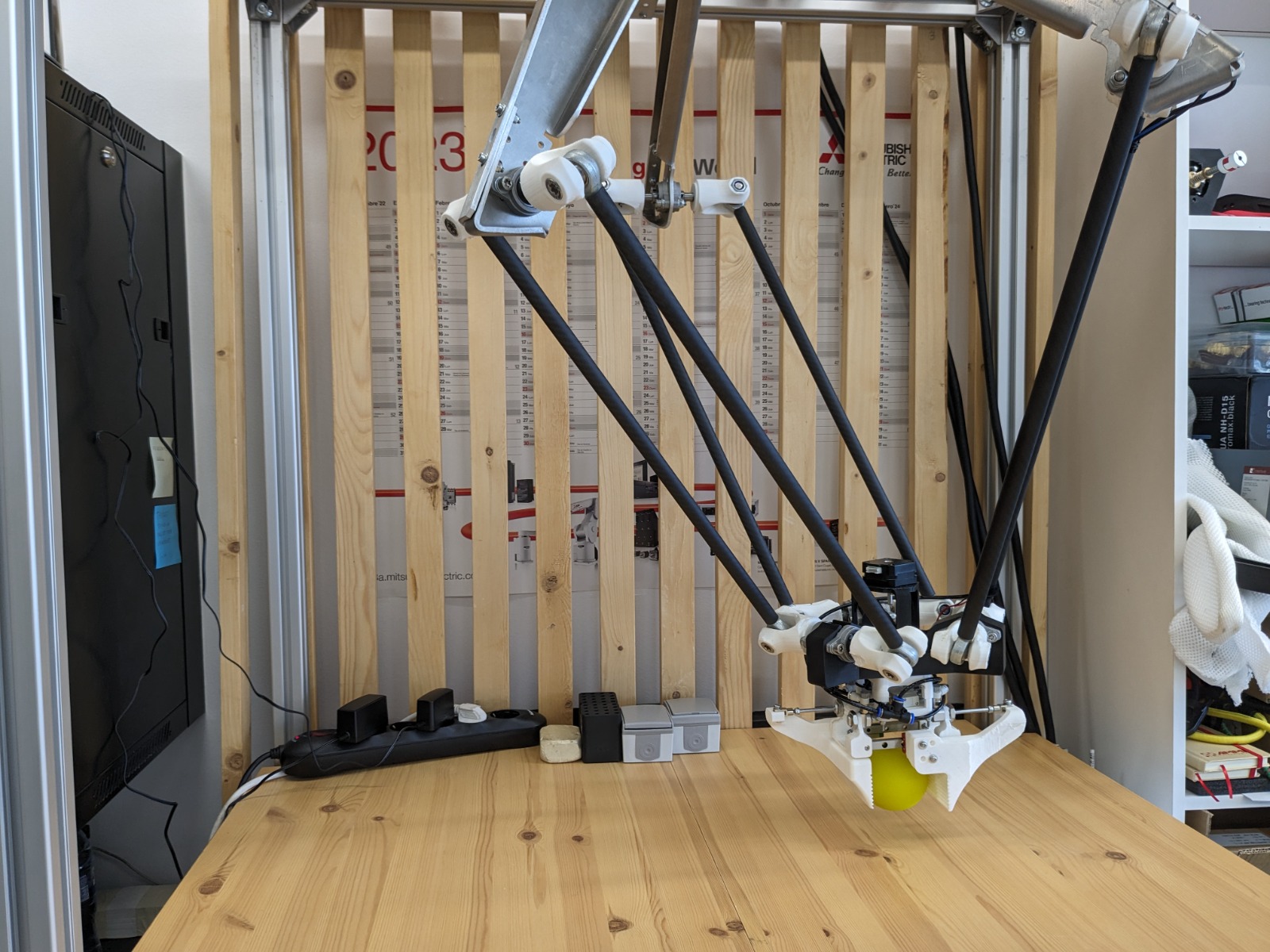}
        \caption{"Take the ball"}
    \end{subfigure}
    \hfill 
    \begin{subfigure}{0.32\textwidth}
        \includegraphics[width=\linewidth]{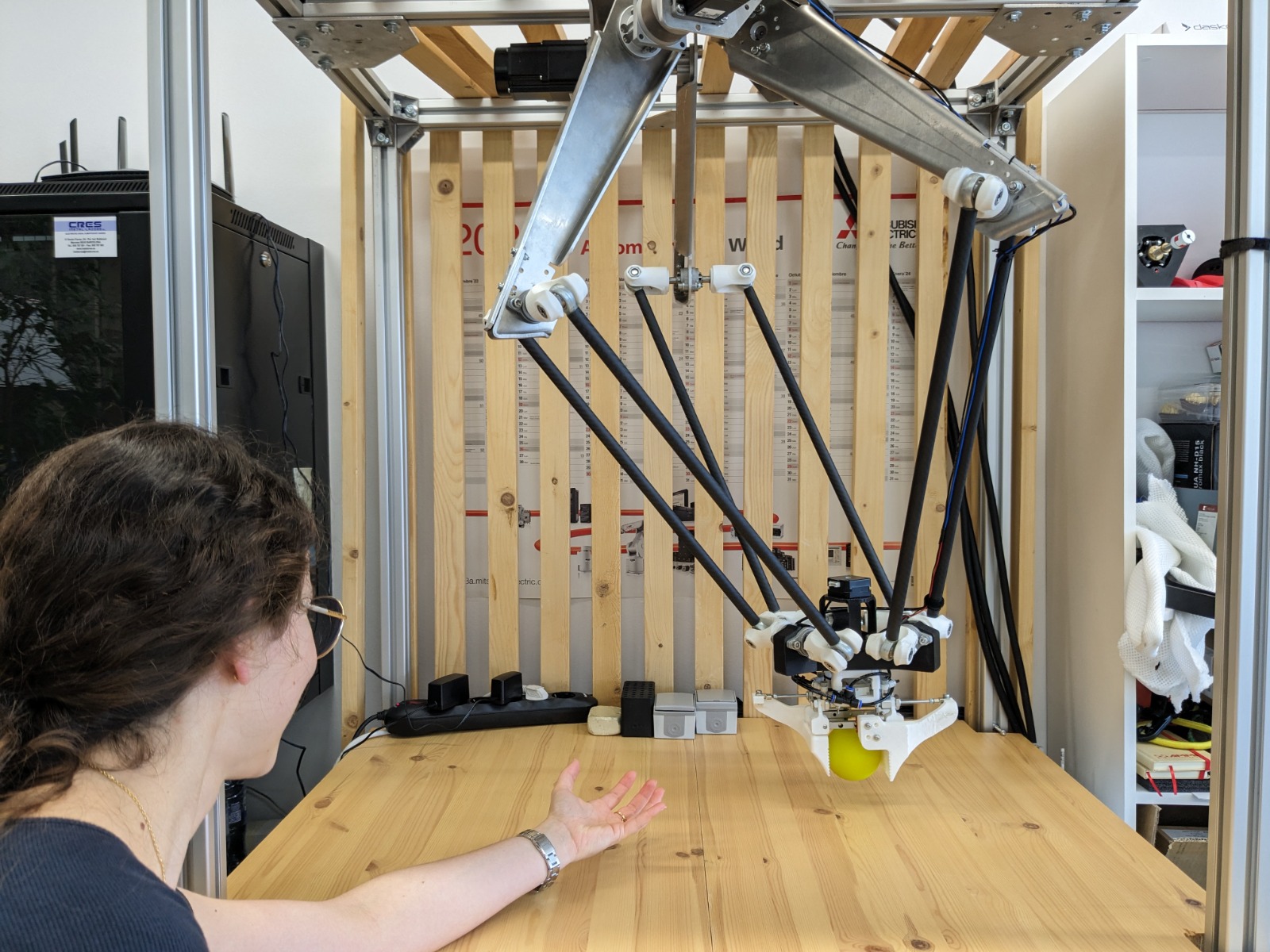}
        \caption{"Follow my hand"}
    \end{subfigure}
    \hfill
    \begin{subfigure}{0.32\textwidth}
        \includegraphics[width=\linewidth]{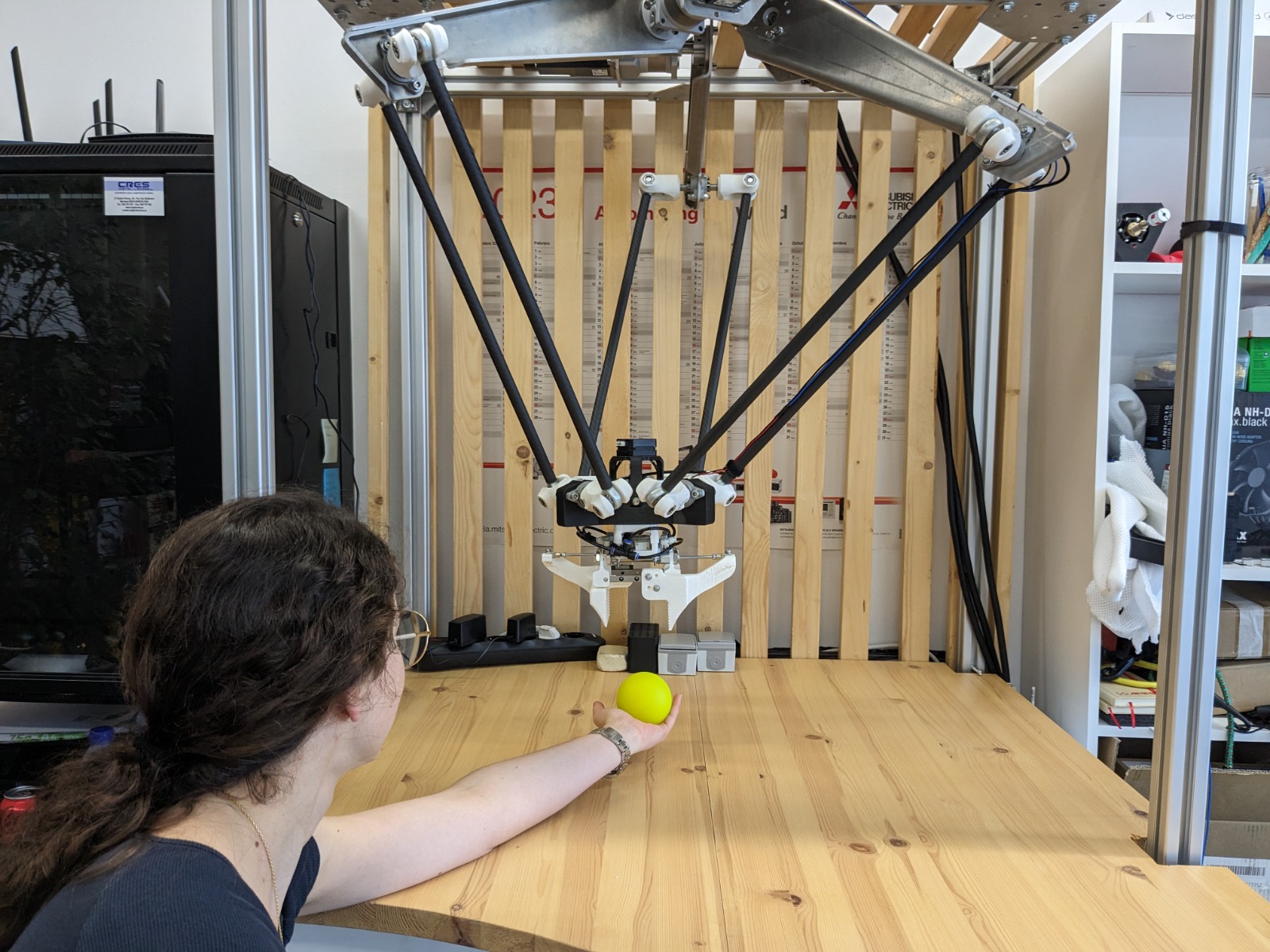}
        \caption{"Give it to me"}
    \end{subfigure}
    \caption{Task: "Go ahead, take the ball and then, give it back to my hand. The hand is moving and the robot must follow it" - Example of dynamic experiment on DELTA Robot}
    \label{fig:dyn-exp1-delta-e}
\end{figure}